\documentclass{article}

    \usepackage[final, nonatbib]{neurips_2022}

\usepackage[utf8]{inputenc} 
\usepackage[T1]{fontenc}    
\usepackage{hyperref}       
\usepackage{url}            
\usepackage{booktabs}       
\usepackage{amsfonts}       
\usepackage{nicefrac}       
\usepackage{microtype}      
\usepackage{amsmath}
\usepackage{graphicx}
\usepackage{colortbl}
\usepackage{xcolor}
\usepackage{wrapfig}
\usepackage{subcaption}
\usepackage{multirow}
\usepackage{makecell}
\usepackage{mathrsfs}
\usepackage{algorithm}
\usepackage{algpseudocode}

\title{CK4Gen: A Knowledge Distillation Framework\\
for Generating High-Utility\\
Synthetic Survival Datasets in Healthcare}

\author{Nicholas I-Hsien Kuo, Blanca Gallego, Louisa Jorm\\
Centre for Big Data Research in Health (CBDRH)\\
The University of New South Wales, Sydney, Australia\\
\footnotesize{\textcolor{white}{*}}\\
Corresponding author: Nicholas I-Hsien Kuo (\texttt{n.kuo@unsw.edu.au})
}

\begin{document}

\maketitle

\begin{abstract}
Access to real clinical data is heavily restricted by privacy regulations, hindering both healthcare research and education. These constraints slow progress in developing new treatments and data-driven healthcare solutions, while also limiting students' access to real-world datasets, leaving them without essential practical skills. High-utility synthetic datasets are therefore critical for advancing research and providing meaningful training material. However, current generative models -- such as Variational Autoencoders (VAEs) and Generative Adversarial Networks (GANs) -- produce surface-level realism at the expense of healthcare utility, blending distinct patient profiles and producing synthetic data of limited practical relevance. To overcome these limitations, we introduce \textbf{CK4Gen} (Cox Knowledge for Generation), a novel framework that leverages knowledge distillation from Cox Proportional Hazards (CoxPH) models to create synthetic survival datasets that preserve key clinical characteristics, including hazard ratios and survival curves. CK4Gen avoids the interpolation issues seen in VAEs and GANs by maintaining distinct patient risk profiles, ensuring realistic and reliable outputs for research and educational use. Validated across four benchmark datasets -- GBSG2, ACTG320, WHAS500, and FLChain -- CK4Gen outperforms competing techniques by better aligning real and synthetic data, enhancing survival model performance in both discrimination and calibration via data augmentation. As CK4Gen is scalable across clinical conditions, and with code to be made publicly available, future researchers can apply it to their own datasets to generate synthetic versions suitable for open sharing\footnote{\label{Ref:OpenSource}Our codes and generated synthetic datasets will be made publicly available after paper acceptance; and refer to reproducibility statement in Supplementary Section A.}.
\end{abstract}

\section{Background \& Summary}

Synthetic datasets are critical in machine learning (ML), especially in healthcare, where privacy regulations like HIPAA in the USA~\cite{nosowsky2006health} and the Privacy Act 1988 in Australia~\cite{okeefe2010privacy} limit access to real clinical data. Synthetic data enables the development of ML models without exposing sensitive information~\cite{kuo2022health}. However, generating high-utility synthetic datasets~\cite{nicholas2023generating} -- realism beyond single variable distributions to substitute real data for analysis -- remains challenging. This is especially difficult in survival analysis~\cite{clark2003survival}, a method that focuses on time-to-event outcomes and censored data.

While privacy is crucial, this paper focuses on \textbf{data utility}. Existing methods for generating synthetic survival data are typically tailored to specific illnesses, limiting their generalisability across different medical conditions~\cite{smith2022generating}. Additionally, many methods emphasise the validation of synthetic data realism or focus predominantly on quantitative metrics~\cite{ashhad2024conditioning, rollo2024syndsurv}, which, while often impressive in technical evaluations, may not fully address the practical needs of clinical settings. In practice, clinicians are more concerned with how synthetic data can be applied to inform patient care and guide decision-making. Thus, this study prioritises generating realistic, high-utility synthetic survival datasets, with disclosure control~\cite{el2020evaluating} left for future work.

Survival analysis poses greater complexity than binary or continuous outcomes due to censored data and the need to model time-to-event outcomes. Generative frameworks must ensure synthetic data retains structural and statistical fidelity to real-world data. 

We introduce \textbf{CK4Gen} (Cox Knowledge for Generation), a machine learning framework for generating high-utility synthetic datasets for survival analysis. Unlike traditional approaches, CK4Gen addresses the unique challenges of survival data, ensuring that synthetic datasets maintain their clinical relevance and statistical properties.

\underline{Addressing Bias in Synthetic Survival Data Generation}\\
Generative models such as variational autoencoders (VAEs)~\cite{kingma2014auto} and generative adversarial networks (GANs)~\cite{goodfellow2014generative} are commonly used for synthetic data generation. However, VAEs can suffer from posterior collapse~\cite{lucas2019understanding}, where distinct patient profiles are collapsed into similar latent representations, which is particularly problematic in survival analysis where unique time-to-event outcomes are crucial. GANs, meanwhile, can experience \textit{mode collapse}~\cite{goodfellow2016nips}, producing limited data diversity and failing to capture the broad spectrum of patient risk profiles inherent to survival data.

To overcome these issues, CK4Gen leverages Cox proportional hazards (CoxPH) models~\cite{cox1972regression} in survival analysis that is widely used in epidemiological and clinical research. CoxPH assumes that the effect of covariates on the hazard is constant over time -- the proportional hazards assumption -- and is expressed as:
\begin{equation}\label{Eq:CoxPH}
\mathscr{H}(t | \mathbf{X}_\text{[:]}) = \mathscr{H}_0(t) \exp(\beta_1 \mathbf{X}_\text{[1]} + \beta_2 \mathbf{X}_\text{[2]} + \dots + \beta_p \mathbf{X}_\text{[$p$]})
\end{equation}
where \( \mathscr{H}(t | \mathbf{X}) \) is the hazard at time \( t \), \( \mathscr{H}_0(t) \) is the baseline hazard, \( \mathbf{X} \) represents the covariates, and \( \beta_1, \beta_2, \dots, \beta_p \) are their respective coefficients. The hazard ratio \( \exp(\beta_k) \) indicates the change in risk associated with a one-unit increase in covariate \( \mathbf{X}_\text{[$k$]} \). In this work, we rely on CoxPH’s ability to handle censored data to regularise CK4Gen, informing the learnt heuristics in CK4Gen to be grounded in a trusted statistical model for generating high-utility synthetic survival data.

\underline{CK4Gen: Bridging Survival Analysis and High-Utility Synthetic Data}\\
CK4Gen distils relationships from the CoxPH model into a neural network, using principles akin to error-bounded lossy compression~\cite{jin2022improving, underwood2024understanding}, where controlled distortions preserve essential information. It consists of a \textbf{DCM Encoder} for encoding survival data into latent representations and a \textbf{SynthNet Decoder} for generating synthetic data. Like lossy compression techniques such as floating-point compression~\cite{lindstrom2006fast}, CK4Gen maintains critical survival structures ensuring key relationships remain intact while allowing approximations of non-essential variables.

A core insight of CK4Gen is that patients fall into distinct risk levels. Preserving this heterogeneity is vital for realistic synthetic data. CK4Gen leverages an internal clustering algorithm and knowledge distillation~\cite{hinton2015distilling} from CoxPH to ensure synthetic data reflects real-world survival patterns. Unlike VAEs and GANs, which may blend risk profiles, CK4Gen focuses on augmenting datasets or producing synthetic versions for public release, maintaining data utility without unrealistic profiles.

CK4Gen has been tested on four datasets: GBSG2~\cite{schumacher1994randomized}, ACTG320~\cite{hammer1997controlled}, WHAS500~\cite{goldberg1988incidence}, and FLChain~\cite{dispenzieri2012use}. These synthetic datasets and the algorithm will be made publicly available to support medical education and further research.

\section{Methods}\label{Sec:Methods}

In this work, we applied CK4Gen to generate synthetic data for four survival analysis datasets: GBSG2~\cite{schumacher1994randomized}, ACTG320~\cite{hammer1997controlled}, WHAS500~\cite{goldberg1988incidence}, and FLChain~\cite{dispenzieri2012use}. This section outlines the original datasets and the CK4Gen deep learning framework used to generate synthetic data, which includes all variables from the real datasets in identical formats. We begin with an overview of the original datasets and their clinical variables, followed by a description of the CK4Gen framework's architecture and training process, encompassing both the DCM encoder and SynthNet decoder.

\newpage
\subsection{The Ground Truth Datasets}

\begin{table}[h]
\footnotesize
\centering
\begin{tabular}{|l||l|l|l|l|p{3cm}|}
\hline
\textbf{Study} & 
\textbf{Country} & 
\textbf{Sample Size} & 
\textbf{Covariates} & 
\textbf{Median Follow-up} & 
\textbf{Event Definition} \\ 

\hline
\hline
\textbf{GBSG2}     
& Germany                       
& 686                  
& 8                   
& 5 years                   
& Recurrence,\newline new tumour, or death \\ 

\hline
\textbf{ACTG320} 
& USA 
& 1,151
& 6
& 38 weeks
& AIDS-defining\newline condition or death \\ 

\hline
\textbf{WHAS500}
& USA
& 500
& 11
& 1,627 days
& Death \\ 

\hline
\textbf{FLChain}
& USA
& 15,859
& 8
& 12.7 years
& Death \\ 

\hline
\end{tabular}
\caption{Summary of Four Ground Truth Datasets Used for Synthetic Data Generation.}
\label{tab:groundtruthdatasets}
\end{table}

A summary of the key characteristics of the datasets is presented in Table~\ref{tab:groundtruthdatasets}. Although these datasets are widely used across various research domains, they have often been processed differently depending on the study, with notable variations in how variables are defined and analytical approaches are applied. In Supplementary Section B, we provide a detailed account of the steps we followed to ensure a consistent definition of the ground truth across all datasets. Additionally, we confirmed that the original variables could replicate survival analysis results from prior studies, ensuring that the datasets serve as a robust baseline for evaluating the utility of the synthetic data generated.

Further details on the structure and descriptive statistics of the datasets are provided in Tables \ref{Tab:GBSG2} -- \ref{Tab:FLChain}. These tables provide a detailed overview of the synthetic datasets, designed to retain the structure of the original data. The statistics reflect the synthetic data, while the original dataset statistics can be found in the corresponding studies. Variables are shown as Numeric, Binary, or Multiple Binaries (with categories encoded as separate binary columns). Means and standard deviations are given for numeric variables, and proportions for binary variables.

\subsubsection{GBSG2}

The GBSG2 dataset from Schumacher \textit{et al.}~\cite{schumacher1994randomized} was designed to assess the effectiveness of breast cancer treatments, comparing chemotherapy with hormonal therapy. The study tracked 686 women over a median period of 5 years to observe recurrence-free and overall survival. The primary outcome was the time from surgery to cancer recurrence, the development of a new tumour, or death.

We format the dataset following Sauerbrei \textit{et al.}~\cite{sauerbrei1999modelling}. The baseline patient for CoxPH modelling is defined as a pre-menopausal woman aged 45 or younger, with a tumour size $\le$ 20 mm (WHO Grade I, low-grade), 1-3 positive lymph nodes, and progesterone and oestrogen receptor levels $\ge$ 20 fmol/mg. This patient did not receive tamoxifen hormonal therapy.

\subsubsection{ACTG320}

The ACTG320 dataset originates from Hammer \textit{et al.}~\cite{hammer1997controlled}, a study that evaluated the effectiveness of adding the protease inhibitor indinavir to a standard two-drug regimen for HIV-1 patients. The trial included 1,151 patients with prior zidovudine treatment and CD4 cell counts of $\le$ 200 cells/$\text{mm}^3$. Patients were followed for a median of 38 weeks, with the primary outcome being time to an AIDS-defining condition or death.

For CoxPH survival analysis, the baseline patient is characterised as a female with a CD4 cell count $\le$ 50 cells/$\text{mm}^3$, receiving the standard two-drug regimen, and showing no functional impairment, as measured by the Karnofsky performance scale~\cite{karnofsky1948use}.

\subsubsection{WHAS500}

The WHAS500 dataset is derived from the Worcester Heart Attack Study (WHAS), a longitudinal population-based study introduced by Goldberg \textit{et al.}~\cite{goldberg1988incidence} to investigate the incidence and survival rates following hospital admission for acute myocardial infarction (MI). The dataset includes 500 individuals, capturing factors associated with acute MI incidence and survival.

For CoxPH survival analysis, the baseline patient is characterised as a male with an average BMI, normal heart rate, and normal systolic and diastolic blood pressure. He has no history of cardiovascular disease, atrial fibrillation, or congestive heart failure, and is experiencing his first MI, specifically a non Q-wave type.

\newpage
\subsubsection{FLChain}

The FLChain dataset, from Dispenzieri \textit{et al.}~\cite{dispenzieri2012use}, was designed to assess the prognostic value of non-clonal serum immunoglobulin free light chains (FLCs) in predicting overall survival. FLCs are key blood proteins involved in immune responses, and their measurement is critical for diagnosing and monitoring conditions such as kidney disease and chronic inflammation. The study included 15,859 participants, with data collected between 1995 and 2003, and follow-up extending to 2009. In this analysis, an event is defined as death from any cause.

For CoxPH survival analysis, the baseline patient is defined as a female aged 50 or older, with serum immunoglobulin FLC levels not in the top decile, normal serum creatinine levels, and no diagnosis of MGUS (monoclonal gammopathy of undetermined significance).

\begin{table}[h]
\small
\centering
\begin{tabular}{|p{3cm}|p{2cm}|p{3cm}|p{4cm}|}
\hline
\textbf{Variable} & \textbf{Type} & \textbf{Category/Statistic} & \textbf{Value/Proportion} \\
\hline
\hline
\textbf{Age} & Multiple\newline Binaries & \(\leq 45\) years \newline 46-60 years \newline \(> 60\) years & Baseline \newline 53.06\% \newline 28.86\% \\
\hline
\hline
\textbf{Menopausal\newline State} & Binary & Pre- \newline Post- & Baseline \newline 53.79\% \\
\hline
\hline
\textbf{Tumour Size} & Multiple\newline Binaries & \(\leq 20\) mm \newline 21-30 mm \newline \(> 30\) mm & Baseline \newline 43.00\% \newline 31.20\% \\
\hline
\textbf{Tumour Grade} & Multiple\newline Binaries & I \newline II \newline III & Baseline \newline 67.2\% \newline 21.28\% \\
\hline
\hline
\textbf{Number of\newline Positive Nodes} & Multiple\newline Binaries & \(\leq 3\) \newline 4-9 \newline \(\geq 10\) & Baseline \newline 30.17\% \newline 14.72\% \\
\hline
\hline
\textbf{Progesterone\newline Receptor Level} & Binary & \(\geq 20\) fmol/mg \newline \(< 20\) fmol/mg & Baseline \newline 37.46\% \\
\hline
\textbf{Oestrogen\newline Receptor Level} & Binary & \(\geq 20\) fmol/mg \newline \(< 20\) fmol/mg & Baseline \newline 37.76\% \\
\hline
\hline
\textbf{Hormonal Therapy} & Binary & False \newline True & Baseline \newline 33.97\% \\
\hline
\hline
\textbf{Event} & Binary & Occurred & 43.59\% \\
\hline
\textbf{Duration} & Numeric & Mean ± Std Dev \newline Median (Q1 - Q3) & 1124.49 ± 642.79 \newline 1084.00 (567.75 - 1684.75) \\
\hline
\end{tabular}
\caption{\label{Tab:GBSG2}An overview of the structure and descriptive statistics for the GBSG2 synthetic dataset.}
\end{table}

\begin{table}[h]
\small
\centering
\begin{tabular}{|p{3cm}|p{2cm}|p{3cm}|p{4cm}|}
\hline
\textbf{Variable} & \textbf{Type} & \textbf{Category/Statistic} & \textbf{Value/Proportion} \\
\hline
\hline
\textbf{Age} & Numeric & Mean ± Std Dev \newline Median (Q1 - Q3) & 38.75 ± 9.16 \newline 38.00 (32.00 - 45.00) \\
\hline
\hline
\textbf{Sex} & Binary & Female \newline Male & Baseline \newline 85.66\% \\
\hline
\hline
\textbf{CD4 Cell Count} & Binary & \(\leq 50\) cells/mm³ \newline 51-200 cells/mm³ & Baseline \newline 52.56\% \\
\hline
\hline
\textbf{Treatment Indicator} & Binary & Control group \newline Treatment group & Baseline \newline 49.70\% \\
\hline
\hline
\textbf{Functional\newline Impairment} & Numeric & Mean ± Std Dev \newline Median (Q1 - Q3) & 0.86 ± 0.75 \newline 1.00 (0.00 - 1.00) \\
\hline
\hline
\textbf{Months of\newline Prior ZDV Use} & Numeric & Mean ± Std Dev \newline Median (Q1 - Q3) & 29.96 ± 28.49 \newline 20.00 (11.00 - 41.00) \\
\hline
\hline
\textbf{Event} & Binary & Occurred & 8.34\% \\
\hline
\textbf{Duration} & Numeric & Mean ± Std Dev \newline Median (Q1 - Q3) & 230.18 ± 89.88 \newline 257.00 (174.00 - 300.00) \\
\hline
\end{tabular}
\caption{\label{Tab:ACTG320}An overview of the structure and descriptive statistics for the ACTG320 synthetic dataset.}
\end{table}

\newpage
\begin{table}[h]
\small
\centering
\begin{tabular}{|p{4.25cm}|p{1.25cm}|p{3cm}|p{3.5cm}|}
\hline
\textbf{Variable} & \textbf{Type} & \textbf{Category/Statistic} & \textbf{Value/Proportion} \\
\hline
\hline
\textbf{Age} & Numeric & Mean ± Std Dev \newline Median (Q1 - Q3) & 69.84 ± 14.36 \newline 72.00 (59.00 - 82.00) \\
\hline
\hline
\textbf{BMI} & Numeric & Mean ± Std Dev \newline Median (Q1 - Q3) & 26.45 ± 4.15 \newline 25.67 (23.42 - 28.89) \\
\hline
\hline
\textbf{Sex} & Binary & Male \newline Female & Baseline \newline 38.8\% \\
\hline
\hline
\textbf{Heart Rate} & Numeric & Mean ± Std Dev \newline Median (Q1 - Q3) & 83.61 ± 19.06 \newline 80.51 (71.52 - 92.97) \\
\hline
\hline
\textbf{Systolic Blood Pressure\newline (SysBP)} & Numeric & Mean ± Std Dev \newline Median (Q1 - Q3) & 149.16 ± 35.85 \newline 146.91 (120.30 - 171.26) \\
\hline
\textbf{Diastolic Blood Pressure\newline (DiasBP)} & Numeric & Mean ± Std Dev \newline Median (Q1 - Q3) & 76.28 ± 30.94 \newline 78.86 (53.86 - 95.62) \\
\hline
\hline
\textbf{History of\newline Cardiovascular Disease (CVD)} & Binary & False \newline True & Baseline \newline 80.2\% \\
\hline
\textbf{History of\newline Atrial Fibrillation (AF)} & Binary & False \newline True & Baseline \newline 11.8\% \\
\hline
\textbf{History of\newline Congestive Heart Failure (CHF)} & Binary & False \newline True & Baseline \newline 29.6\% \\
\hline
\hline
\textbf{Myocardial Infarction Order\newline (MI Order)} & Binary & First \newline Recurrent & Baseline \newline 31.6\% \\
\hline
\textbf{Myocardial Infarction Type\newline (MI Type)} & Binary & non Q-wave \newline Q-wave & Baseline \newline 29.4\% \\
\hline
\hline
\textbf{Event} & Binary & Occurred & 43.0\% \\
\hline
\textbf{Duration} & Numeric & Mean ± Std Dev \newline Median (Q1 - Q3) & 882.44 ± 705.67 \newline 631.50 (296.50 - 1363.50) \\
\hline
\end{tabular}
\caption{\label{Tab:WHAS500}An overview of the structure and descriptive statistics for the WHAS500 synthetic dataset.}
\end{table}

\begin{table}[h]
\small
\centering
\begin{tabular}{|p{3.25cm}|p{2.25cm}|p{2.75cm}|p{3.75cm}|}
\hline
\textbf{Variable} & \textbf{Type} & \textbf{Category/Statistic} & \textbf{Value/Proportion} \\
\hline
\hline
\textbf{Age (10 year brackets)} & Numeric & Mean ± Std Dev \newline Median (Q1 - Q3) & 5.96 ± 0.94 \newline 6.00 (5.00 - 7.00) \\
\hline
\hline
\textbf{Sex} & Binary & Female \newline Male & Baseline \newline 44.6\% \\
\hline
\hline
\textbf{Creatinine} & Numeric & Mean ± Std Dev \newline Median (Q1 - Q3) & 1.05 ± 0.19 \newline 1.02 (0.93 - 1.16) \\
\hline
\hline
\textbf{Top FLC Decile} & Binary & Not in Top Decile \newline In Top Decile & Baseline \newline 8.61\% \\
\hline
\textbf{Decile of Total FLC} & Numeric & Mean ± Std Dev \newline Median (Q1 - Q3) & 5.32 ± 2.70 \newline 5.00 (3.00 - 7.00) \\
\hline
\textbf{Serum FLC Kappa} & Numeric & Mean ± Std Dev \newline Median (Q1 - Q3) & 1.36 ± 0.66 \newline 1.23 (0.94 - 1.66) \\
\hline
\textbf{Serum FLC Lambda} & Numeric & Mean ± Std Dev \newline Median (Q1 - Q3) & 1.61 ± 0.65 \newline 1.53 (1.21 - 1.92) \\
\hline
\hline
\textbf{Diagnosed with MGUS} & Binary & False \newline True & Baseline \newline 1.26\% \\
\hline
\hline
\textbf{Event} & Binary & Occurred & 27.55\%\\
\hline
\textbf{Duration} & Numeric & Mean ± Std Dev \newline Median (Q1 - Q3) & 3661.04 ± 1432.68 \newline 4302.00 (2852.00 - 4773.00) \\
\hline
\end{tabular}
\caption{\label{Tab:FLChain}An overview of the structure and descriptive statistics for the FLChain synthetic dataset.}
\end{table}

\newpage

\begin{figure}[h]
    \centering
    \includegraphics[width=\textwidth]{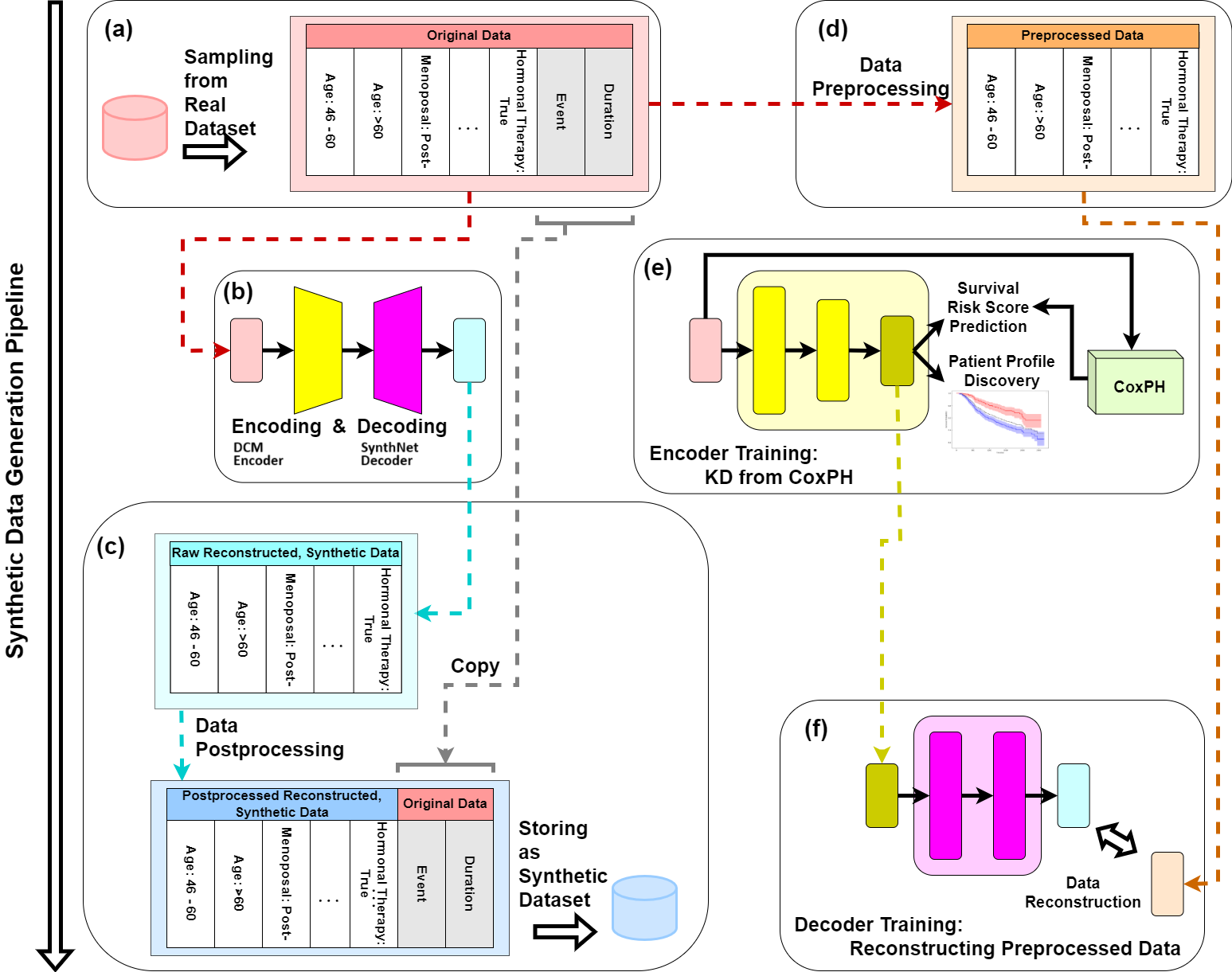}
    \caption{Overview of the CK4Gen framework for generating synthetic datasets.\newline
    Panels (a), (b), and (c) illustrate the main pipeline.\newline
    \hspace*{10mm}Panel (a) shows the process starting with sampling from the real dataset.\newline
    \hspace*{10mm}Panel (b) depicts the CK4Gen framework, where the DCM encoder and SynthNet decoder\newline
    \hspace*{15mm}reconstruct the real data into synthetic data.\newline
    \hspace*{10mm}Panel (c) highlights the postprocessing of reconstructed data to restore the original scale and\newline
    \hspace*{15mm}format, with Event and Duration directly copied from the original dataset and combined\newline
    \hspace*{15mm}with the synthetic data to form the final output.\newline
    Panels (d), (e), and (f) provide supplementary details.\newline
    \hspace*{10mm}Panel (d) illustrates the preprocessing of raw patient data.\newline
    \hspace*{10mm}Panel (e) shows the training of the DCM encoder using knowledge distillation from a\newline
    \hspace*{15mm}pre-trained CoxPH model, with raw data as input.\newline
    \hspace*{10mm}Panel (f) depicts the SynthNet decoder, which receives latent representations from the DCM\newline
    \hspace*{15mm}encoder and is trained to reconstruct the preprocessed synthetic data.\newline
    \textcolor{white}{.}\newline
    This figure is best viewed in colours.\newline
    \hspace*{10mm}Red: \hspace*{4.5mm}Appears in panel (a). Represents the raw, original data from the real dataset.\newline
    \hspace*{10mm}Grey: \hspace*{3.375mm}Appears in panels (a) and (c). Represents Event and Duration from the original data.\newline
    \hspace*{10mm}Yellow: \hspace*{0.25mm}Appears in panels (b) and (e). Represents the DCM encoder and its representations.\newline
    \hspace*{10mm}Purple: \hspace*{1.125mm}Appears in panels (b) and (f). Represents the SynthNet decoder.\newline
    \hspace*{10mm}Cyan: \hspace*{3mm}Appears in panel (c). Represents the raw reconstructed data generated by SynthNet.\newline
    \hspace*{10mm}Blue: \hspace*{3.75mm}Appears in panel (c). Represents the postprocessed reconstructed data.\newline
    \hspace*{10mm}Orange: Appears in panel (d). Represents the preprocessed ground truth data.}
    \label{fig:CK4GenOvervoew}
\end{figure}

\newpage
\subsection{The CK4Gen Framework}

Figure \ref{fig:CK4GenOvervoew} illustrates the CK4Gen autoencoder framework, comprising the DCM encoder and SynthNet decoder. These components work together to extract clinical features from real patient data, learning latent representations and exploring potential phenotypes to generate synthetic patient data.

A patient’s survival outcome is shaped by a combination of factors that define their phenotype, such as comorbidities~\cite{nagamine2020multiscale} and pathology characteristics~\cite{barbieri2022predicting}. The DCM encoder captures these survival-related features by clustering patients into distinct, data-driven subgroups. This clustering ensures that CK4Gen accounts for population variability and preserves distinctions in the synthetic data. By maintaining the integrity of these subgroups, the SynthNet decoder generates synthetic patients without mixing data across different subgroups, ensuring that the generated patients exhibit realistic and clinically meaningful risk profiles.

To maintain the clinical integrity, CK4Gen utilises a CoxPH model to regularise the heuristics learned by the DCM encoder. Additionally, to preserve real-world distributions for time-to-event outcomes, CK4Gen directly copies the \texttt{Event} and \texttt{Duration} variables from the original data.

Below, we use uppercase \(\mathbf{X}\) to represent the population's clinical features (see Equation \eqref{Eq:CoxPH}) and lowercase \(\mathbf{x}_i\) refers to individual patient data. Neural network layers are denoted as \(\{j\}\).

\subsubsection{The Encoder Network of CK4Gen}

\underline{Components of the DCM Encoder}\\
The encoder network is based on the Deep Cox Mixture (DCM) model~\cite{nagpal2021deep}, a deep learning architecture designed to capture nuanced differences within survival data. The DCM encoder processes real patient features, excluding the copied \texttt{Event} and \texttt{Duration} columns, and extracts latent representations that are key to predicting survival outcomes.

The architecture consists of three fully connected hidden layers, containing 64, 32, and 16 neurons, respectively. Each layer captures progressively more complex patterns from the input data. The transformation at each layer follows:
\begin{equation}
\mathbf{h}_{\{j\}} = \text{BatchNorm}\left(\text{ReLU}(\mathbf{W}_{\{j\}} \mathbf{h}_{\{j-1\}} + \mathbf{b}_{\{j\}})\right),
\end{equation}
where \(\mathbf{W}_{\{j\}}\) and \(\mathbf{b}_{\{j\}}\) are the weights and biases of the layer, and ReLU activations~\cite{nair2010rectified} combined with Batch Normalisation~\cite{ioffe2015batch} help to regularise the network. The input to the first layer, \(\mathbf{h}_{\{0\}} = \mathbf{x}\), consists of all patient features, mirroring the input structure of a traditional CoxPH model.

\underline{Risk Scores and Patient Profiles}\\
The output from the final hidden layer, \(\mathbf{h}_{\{3\}}\), serves two purposes: it generates logits (\(\mathbf{y}_{\text{logits}}\)) for survival outcome prediction, following the logic of CoxPH, and produces mixture weights (\(\mathbf{\omega}_{\text{mix}}\)), which capture heterogeneity in the population represented by probability distributions.

The logits are computed as:
\begin{equation}
\mathbf{y}_{\text{logits}} = \mathbf{W}_{\{\text{output}\}} \mathbf{h}_{\{3\}} + \mathbf{b}_{\{\text{output}\}},
\end{equation}
while the mixture weights are calculated using a softmax function:
\begin{equation}\label{Eq:Clustering}
\mathbf{\omega}_{\text{mix}} = \text{softmax}(\mathbf{W}_{\{\text{mix}\}} \mathbf{h}_{\{3\}} + \mathbf{b}_{\{\text{mix}\}}).
\end{equation}
This allows the model to assign each individual with distinct survival characteristics to a combination of latent phenotypic subgroups.

\underline{Training the DCM Encoder Using Knowledge Distillation}\\
To train the DCM encoder, we employ Knowledge Distillation (KD)~\cite{hinton2015distilling}, where a student model learns to replicate the predictions of a teacher model. The CoxPH teacher model (\(g^\text{teacher}\)) is first trained on the survival data using the \texttt{lifelines} library~\cite{Davidson-Pilon2019}. This teacher model estimates log HRs and produces risk scores, which serve as soft targets for training the student DCM model. The risk score for each patient is given by:
\begin{equation}
\mathbf{y}_{\text{CoxPH}, i} = g^\text{teacher}(\mathbf{x}_i) = \exp(\mathbf{x}_i \mathbf{\beta}),
\end{equation}
where \(\mathbf{\beta}\) represents the coefficients learned by the CoxPH model.

The student DCM encoder, parameterised by \(\theta\) (\textit{i.e.,} weights \(\mathbf{W}_{\{j\}}\) and biases \(\mathbf{b}_{\{j\}}\)), is trained by minimising the mean squared error (MSE) between the predicted risk scores and the soft targets from the CoxPH teacher model. The objective function for training is:
\begin{equation}
\theta^* = \underset{\theta}{\arg\min} \, \frac{1}{N} \sum_{i=1}^{N} \mathscr{L}_{\text{KD}}(f_\theta^\text{student}(\mathbf{x}_i), g^\text{teacher}(\mathbf{x}_i)),
\end{equation}
where \(\mathscr{L}_{\text{KD}}\) is the loss function for knowledge distillation. The Adam optimiser~\cite{kingma2014adam} is used to update the model’s parameters during training.

For complete details on the training process, refer to Supplementary Section C.

\subsubsection{The Decoder Network of the CK4Gen}

We refer to the decoder of the CK4Gen framework as SynthNet. SynthNet reconstructs patient data from the final latent representations generated by the DCM encoder, \(\mathbf{h}_{\{3\}}\), while preserving the risk profiles identified through the encoder. 

Unlike the DCM decoder, which directly maps input data as it was trained using knowledge distillation from a CoxPH teacher model (and thus requires no additional preprocessing or postprocessing). The SynthNet decoder requires specific data handling, the input data (\(x\)) is preprocessed into a standardised format (\(\tau\)). Numerical variables are transformed using Box-Cox transformations~\cite{box1964analysis} and rescaled to the range [0,1], while binary variables are mapped to \{0,1\}. This ensures consistency in the range of all variables, regardless of their original type. After creating novel patient data (\(\hat{\tau}_\text{synth}\)) via a reconstruction process, we apply postprocessing to map the data back to its original scale and format (\(\hat{x}_\text{synth}\)), ensuring the output remains consistent with the real variable distributions.

\underline{Components in the SyntheNet Decoder}\\
The SynthNet decoder has a similar but simplified architecture of the DCM encoder. It consists of two intermediate layers followed by an output layer. Each layer involves a linear transformation followed by a non-linear activation function, with hidden dimensions defaulted to 128 neurons per layer.

At each intermediate layer, the transformation can be expressed as:
\begin{equation}
\mathbf{z}_{\{j\}} = \text{ReLU}(\mathbf{U}_{\{j\}} \mathbf{z}_{\{j-1\}} + \mathbf{v}_{\{j\}})
\end{equation}
and the output layer for data reconstruction is:
\begin{equation}
\mathbf{\hat{\tau}}_{\text{synth}} = \text{Sigmoid}(\mathbf{U}_{\{3\}} \mathbf{z}_{\{2\}} + \mathbf{v}_{\{3\}}).
\end{equation}
For the SynthNet decoder, the weights are $\mathbf{U}_{\{j\}}$, the biases are $\mathbf{v}_{\{j\}}$, and $\mathbf{\hat{\tau}}_{\text{synth}}$ refers to the reconstructed, synthetic data before post-processing is applied. Note that $\mathbf{v}_{\{0\}} = \mathbf{h}_{\{3\}}$, where the SynthNet decoder input is the DCM encoder latent state. The sigmoid function at the output ensures that the values of $\mathbf{\hat{\tau}}_{\text{synth}}$ are constrained within the range [0, 1], making them appropriate for both binary and continuous data reconstruction.

\underline{Updating and Fine-Tuning the SynthNet Decoder}\\
During training, SynthNet minimises a loss function to ensure that the reconstructed data closely resembles the original data. We employ MSE as the decoder loss function, which measures the difference between $\mathbf{\hat{\tau}}_{\text{synth}_i} $ and the original rescaled patient data $\mathbf{\tau}_i$:
\begin{equation}
\mathcal{L}_{\text{MSE}} = \frac{1}{N} \sum_{i=1}^{N} \left( \mathbf{\tau}_i - \mathbf{\hat{\tau}}_{\text{synth}, i} \right)^2,    
\end{equation}
and the model parameters are updated using the Adam optimiser.

Refer to more training details for the SynthNet decoder in Supplementary Section D.

\underline{Postprocessing SynthNet Outputs}\\
Binary features are thresholded at 0.5 to convert the continuous outputs from the sigmoid function into discrete binary values, ensuring accurate representation of variables like \texttt{Hormonal Therapy} (for GBSG2, see Table \ref{Tab:GBSG2}) in the synthetic dataset. For continuous features, such as \texttt{Age} (for ACTG320, see Table \ref{Tab:ACTG320}), the data is rescaled and inverse Box-Cox transformations are also applied, to restore the original distribution and match the original feature ranges in the ground truth datasets. 

Refer to more details on preprocessing and postprocessing in Supplementary Section E.

\subsubsection{CK4Gen: To VAE or Not to VAE?}

This section discusses why CK4Gen is designed as an autoencoder instead of a variational autoencoder (VAE)~\cite{kingma2014auto}. Readers primarily interested in the dataset can proceed directly to the Results section.

CK4Gen is designed as an autoencoder, intentionally avoiding models like VAEs or GANs, which may generate synthetic patients that fail to reflect true clinical risk profiles. These models, which rely on sampling from latent spaces or random vectors, often produce data that appears realistic at first glance but diverges from authentic clinical patterns.

\underline{An Information Bottleneck in the VAE}\\
To explain why VAEs are unsuitable for this purpose, it is important to understand their internal mechanism. VAEs encode input data \( \mathbf{x} \) into a latent space \( \mathbf{z} \), sampled from a Gaussian distribution: 
\begin{equation}
q(\mathbf{z}|\mathbf{x}) = \mathcal{N}(\mu(\mathbf{x}), \sigma^2(\mathbf{x}))
\end{equation}
The decoder then reconstructs the data as \( \mathbf{\hat{x}} = f_{\text{decoder}}(\mathbf{z}) \). However, the fact that \( \mathbf{z} \) is sampled from a single Gaussian distribution implies that any inherent differences in patient risk profiles are coerced into this single distribution.

This presents a critical flaw. Regardless of the diversity of risk profiles in the real data, they are forced to fit into the confines of a single Gaussian distribution. As a result, when sampling the latent space, VAEs are likely to interpolate between distinct patient profiles, potentially generating synthetic patients with inconsistent or blended characteristics. This problem is compounded by the fact that patient profiles often depend on complex relationships between multiple variables. As we demonstrate in the Results section, VAEs (as well as GANs) generate synthetic datasets that can significantly degrade the performance of downstream models.

\underline{Lossy Compression}\\
CK4Gen avoids the issues inherent in VAEs by employing the DCM encoder with an internal clustering algorithm (see Equation \eqref{Eq:Clustering}) that discovers distinct clinical profiles. Additionally, by directly copying the \texttt{Event} and \texttt{Duration} from the real dataset to the synthetic one, CK4Gen preserves the critical temporal structures necessary for accurate survival analysis. This ensures that the synthetic datasets retain clinically valid survival outcomes, avoiding the risk of generating synthetic patients with blended or non-sensical risk profiles.

CK4Gen’s design can be likened to error-bounded lossy compression~\cite{jin2022improving, underwood2024understanding}, where critical data structures are preserved while controlled approximations are allowed in less essential areas. This concept is similar to floating-point compression techniques~\cite{lindstrom2006fast}, where approximations are introduced without compromising the key relationships within the data.

\underline{Potential Limitations}\\
While CK4Gen’s focus on accurate reconstruction ensures realistic synthetic data, it may limit the novelty and diversity of the generated data compared to VAE-based approaches. The direct copying of event and duration data could restrict the generation of novel survival outcomes, a potential limitation in specific applications. We further explore these limitations in the Discussion section and in Supplementary Section F, where we introduce alternative approaches to increase synthetic data variability in CK4Gen.

\newpage
\section{Data Record}\label{Sec:DataRecord}

All synthetic datasets are stored as comma-separated value (CSV) files. The synthetic datasets mirror the format of their real counterparts, as detailed in the Methods section under ``The Ground Truth Datasets'' (see Tables \ref{Tab:GBSG2} - \ref{Tab:FLChain}). This section outlines the specific properties of the synthetic datasets, with quality assurance tests provided in the Technical Validation section. 

\subsection{Synthetic GBSG2 Dataset}\label{Sec:DR_GBSG2}

The synthetic GBSG2 dataset is 23.1 KB, containing 686 patients each represented by a single row in a traditional survival analysis format. This dataset comprises 14 columns, with the first 12 dedicated to clinical variables and the remaining two to survival outcomes, namely Event and Duration. 

The clinical variables are listed in Table \ref{Tab:GBSG2} and their visualisations are shown in Figure \ref{fig:Comp_GBSG2}. The descriptive statistics for the synthetic dataset are provided in the accompanying table, including quartiles (25th percentile, median, and 75th percentile) for numeric variables, as well as mean values and standard deviations. For binary and categorical variables, the table shows the proportion of each unique class. Some variables, such as Age 46-60 years, exhibit a more balanced distribution, while others, for instance the Number of Positive Nodes $\ge$10, show significant class imbalance.

All variables for this dataset are presented in a binary format. This approach is employed to reflect the non-linear impact that these variables can have on patient outcomes. For instance, the number of positive lymph nodes is known to significantly influence prognosis at different thresholds~\cite{schumacher1994randomized}, with higher counts associated with increased HRs. To capture this complexity, the dataset uses multiple binary columns to encode these variables, ensuring that the dataset accurately mirrors the clinical reality observed in breast cancer studies. Consider the Number of Positive Nodes, which is divided into three categories: $\leq$3, 4-9, and $\geq$10, with $\leq$3 serving as the baseline category. Two binary columns are used to represent the non-baseline categories: one column indicates whether a patient has 4-9 positive nodes, and another indicates whether the count is $\geq$10. For any given patient, if the entry for 4-9 nodes is 1, the entry for $\geq$10 nodes will be 0, and vice versa. Both columns will have an entry of 0 if the patient has $\leq$3 positive nodes. This method is also applied on Age, Tumour Size, and Tumour Grade to capture the non-linear effects on tumour recurrence and mortality.

\subsection{Synthetic ACTG320 Dataset}\label{Sec:DR_ACTG320}
The synthetic ACTG320 dataset is 40.1 KB in size and contains data for 1,151 patients across 8 columns -- with 6 clinical variables followed by survival Event and Duration. The clinical variables, detailed in Table \ref{Tab:ACTG320}, comprise 3 numeric variables and 3 binary variables. The variable distributions are illustrated in Figure \ref{fig:Comp_ACTG320}. Variable Sex has a noticeable class imbalance with a higher proportion of males, and variable Months of Prior ZDV Use exhibits a long-tailed distribution.

\subsection{Synthetic WHAS500 Dataset}\label{Sec:DR_WHAS500}
The synthetic WHAS500 dataset is 34.0 KB in size and contains data for 500 patients across 13 variables -- 11 clinical features followed by the survival Event and Duration. The descriptive statistics for these variables are detailed in Table \ref{Tab:WHAS500}. This dataset includes 5 numeric variables and 6 binary variables, with variable distributions illustrated in Figure \ref{fig:Comp_WHAS500}. The numeric variables do not have pronounced long tails; however, most binary variables display significant class imbalance.

\subsection{Synthetic FLChain Dataset}\label{Sec:DR_FLChain}
The synthetic FLChain dataset is 459.0 KB in size and comprises 7,874 patients with 10 variables -- 8 clinical features followed by the survival Event and Duration. The descriptive statistics for these variables are provided in Table \ref{Tab:FLChain}. This dataset includes 5 numeric variables and 2 binary variables, with variable distributions  illustrated in Figure \ref{fig:Comp_FLChain}. We noted that creatinine and FLC-related measurements exhibit long right tails; in addition, the variable indicating a history of MGUS is particularly imbalanced.

\newpage
\section{Technical Validation}\label{Sec:TechValid}
This section presents a rigorous, three-tiered validation framework designed to evaluate the performance and utility of CK4Gen's synthetic data generation for survival analysis. The process encompasses Realisticness Validation, Data Augmentation Assessment, and Utility Verification.

In the first phase, we validate the fidelity of CK4Gen by comparing the distributions and correlations of its synthetic datasets against the original data. The second phase quantitatively examines whether integrating synthetic and real data enhances the predictive accuracy of a CoxPH model, benchmarking CK4Gen against leading techniques addressing data imbalance and generative ML models such as VAEs and GANs. The final and most critical phase qualitatively tests the clinical utility of the synthetic data by assessing its impact on clinically meaningful key metrics, including survival curves and HRs, relative to the original datasets.

Survival analysis with synthetic data is still in its early stages, and our validation framework not only builds upon current best practices but also expands them. Many existing methods lack scalability, as they are often designed for specific clinical conditions and cannot generalise across diverse illness profiles~\cite{smith2022generating}. Moreover, machine learning approaches tend to over-emphasise quantitative performance metrics without sufficiently considering the clinical value of the synthetic data they produce~\cite{ashhad2024conditioning, rollo2024syndsurv}. This includes a failure to evaluate the data's healthcare utility or to make the synthetic datasets publicly available for scrutiny. Building on the work of Norcliffe \textit{et al.}~\cite{norcliffe2023survivalgan}, our approach improves validation practices by thoroughly assessing the alignment of HRs between synthetic and real datasets. By training survival models on both, and demonstrating that HRs remain consistent, CK4Gen strengthens confidence in the reliability and clinical relevance of its synthetic datasets.

\subsection{Realism Validation}
This phase evaluates the static properties of the synthetic data by comparing the distributions and statistical moments (\textit{i.e.,} mean and variance) of the real and synthetic variables. We begin by overlaying the probability density functions of the synthetic and real numeric variables using kernel density estimations (KDEs)~\cite{davis2011remarks} for visualisation. For binary (and multiple binary) variables, we employ side-by-side histogram plots. Next, we assess the correlations between variables by calculating and comparing these coefficients for each variable pair in both the synthetic and real datasets, ensuring inter-variable relationships are faithfully captured.

\begin{figure}[h]
    \centering
    \includegraphics[width=0.9\textwidth]{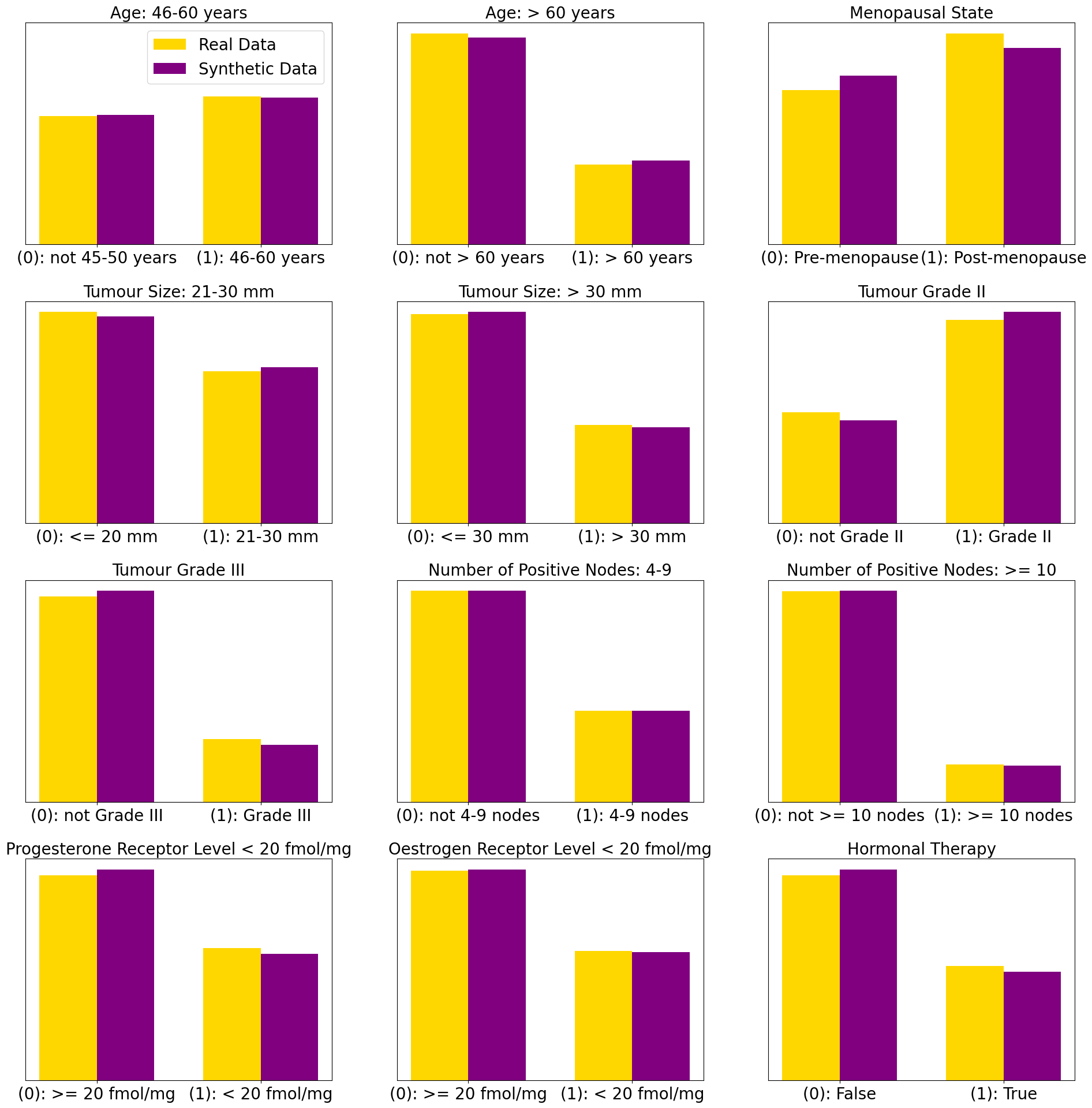}
    \caption{A side-by-side histograms comparing the distributions of binary clinical variables between the real (gold) and synthetic (purple) datasets for the GBSG2 study. The binary variables are compared using histograms. The figure reveals that CK4Gen is capable of synthesising datasets with both balanced and heavily imbalanced variable distributions.}
    \label{fig:Comp_GBSG2}
\end{figure}

First, we compare the distributions for the GBSG2 dataset in Figure \ref{fig:Comp_GBSG2}. Gold represents the ground truth data, while purple denotes the synthetic dataset. No significant visual discrepancies are observed between the distributions of the real and synthetic datasets.

\begin{figure}[h]
    \centering
    \includegraphics[width=0.6\textwidth]{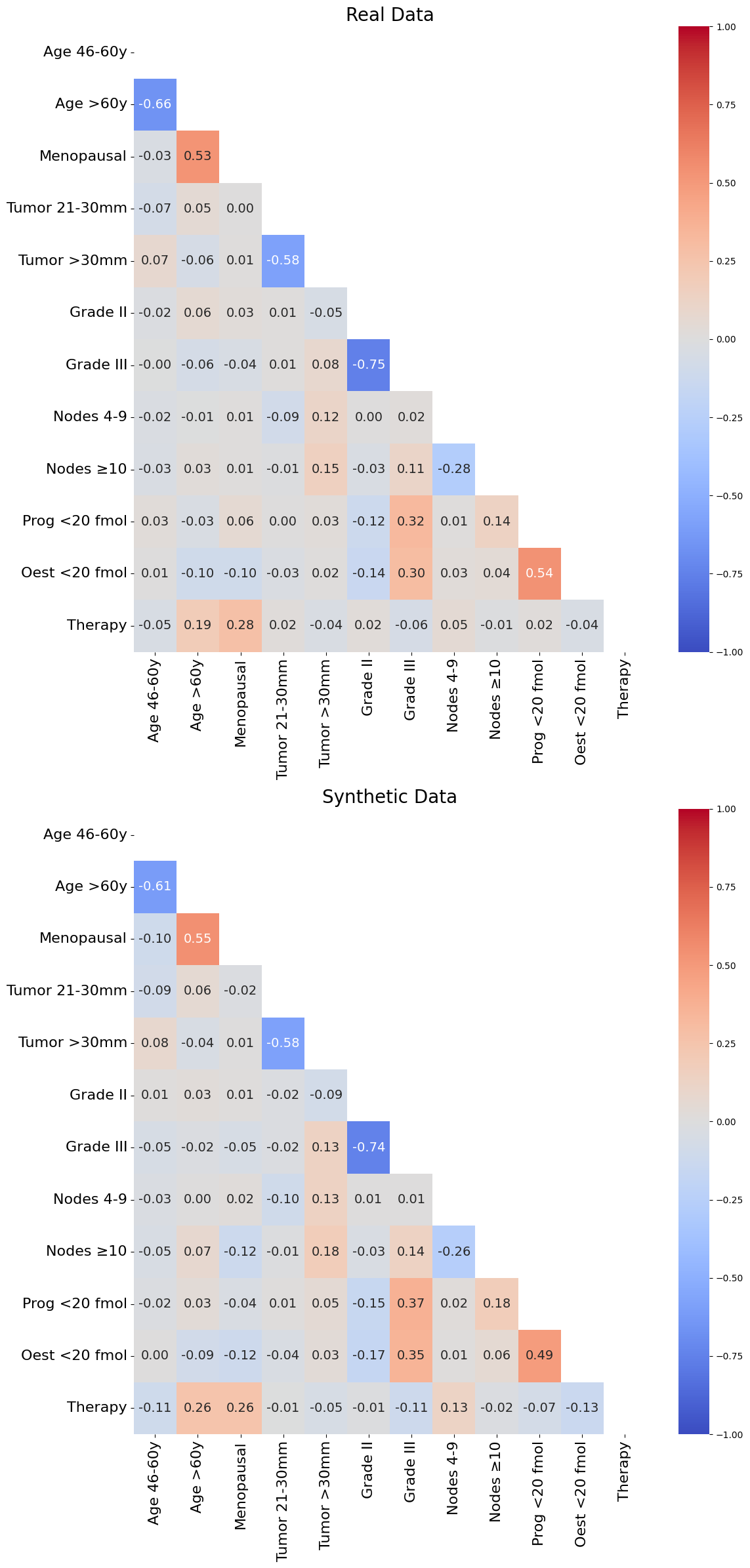}
    \caption{A side-by-side comparison of correlation matrices for the GBSG2 dataset, with real data on the top and synthetic data on the bottom. Blue indicates negative correlations, while red indicates positive correlations. Although some individual correlations show slight variations, CK4Gen generates data that closely mirror the inter-variable relationships of the real data.}
    \label{fig:Corr_GBSG2}
\end{figure}

Next, we assess the correlations shown in Figure \ref{fig:Corr_GBSG2}. While the overall alignment is strong, some variable pairs in the synthetic dataset exhibit slight increases in correlation magnitudes. For instance, the correlation between therapy status and oestrogen level, where the synthetic data shows a magnitude increase from -0.04 to -0.13. Such minor elevations in correlation strength have been observed in other synthetic data studies~\cite{kuo2022health} and likely result from the tendency of generative ML models to exploit data correlations rather than causal relationships. These small differences do not significantly detract from the overall fidelity of the correlations.

\begin{figure}[h]
    \centering
    \includegraphics[width=0.9\textwidth]{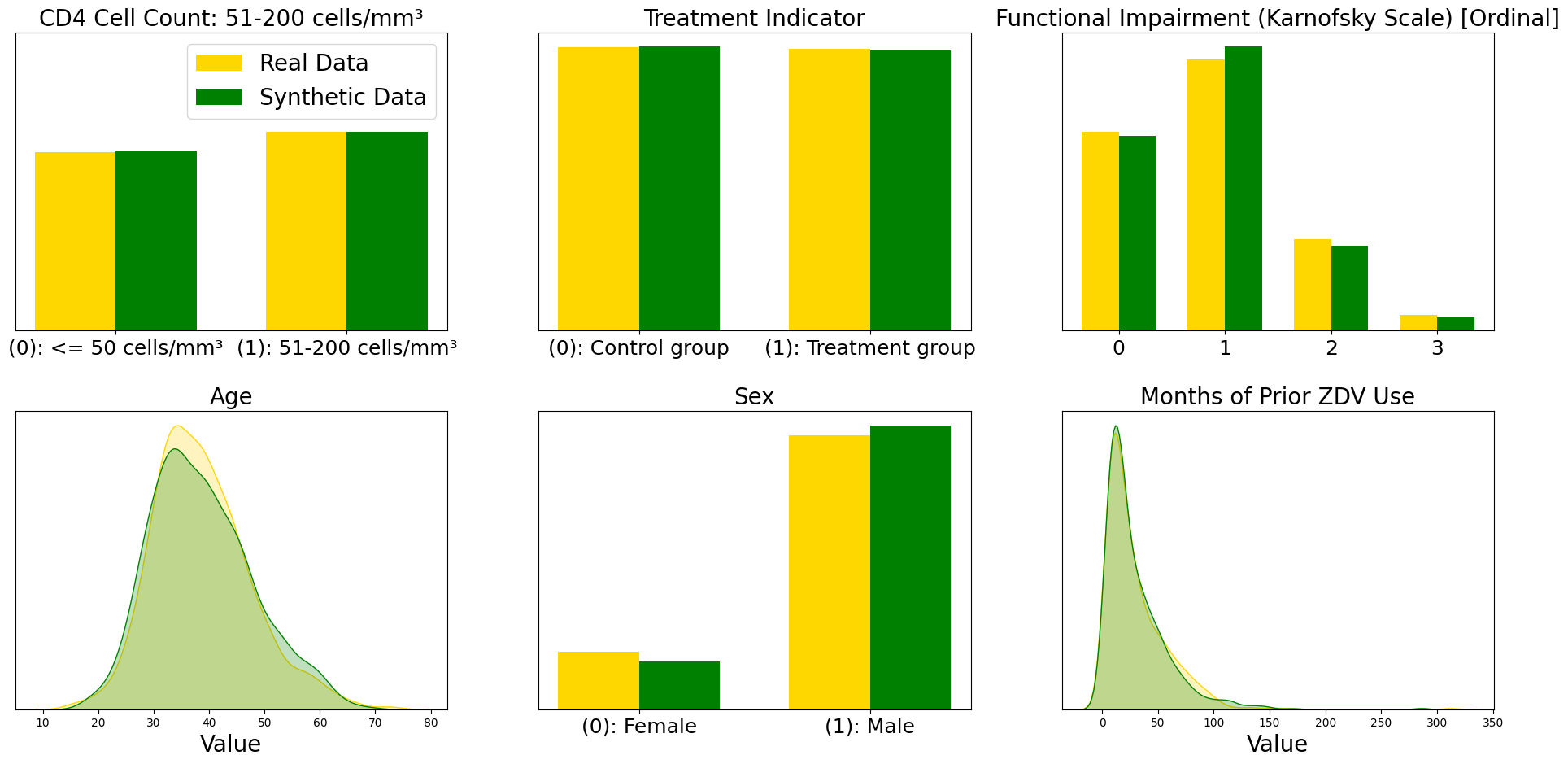}
    \caption{A side-by-side histograms and KDEs comparing the distributions of clinical variables between the real (gold) and synthetic (green) datasets for the ACTG320 study. The binary variables are compared using histograms, while numeric variables are overlaid using KDEs to assess the similarity between distributions. The figure highlights that CK4Gen can generate both balanced and imbalanced binary variables, as well as numeric variables with long-tailed distributions.}
    \label{fig:Comp_ACTG320}
\end{figure}

\begin{figure}[h]
    \centering
    \includegraphics[width=0.9\textwidth]{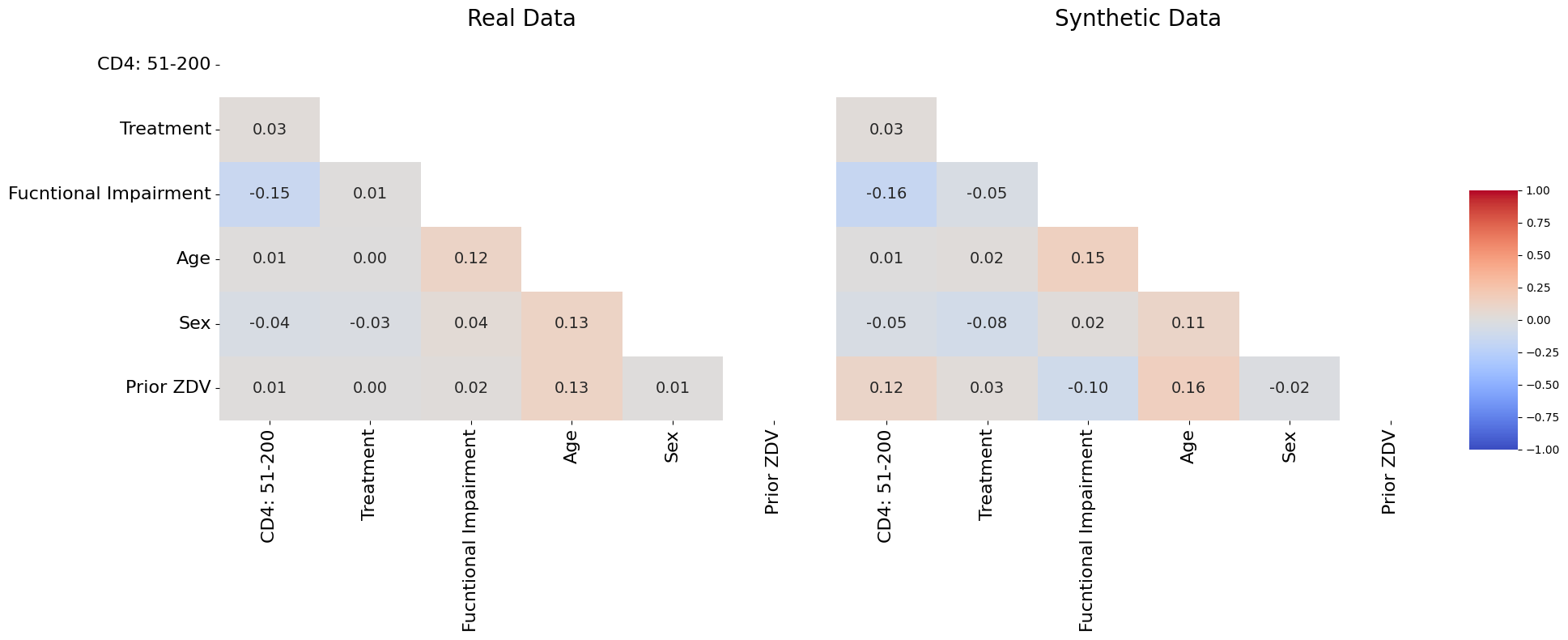}
    \caption{A side-by-side comparison of correlation matrices for the ACTG320 dataset, with real data on the left and synthetic data on the right. CK4Gen demonstrates the ability to generate datasets with highly realistic correlations among numeric variables, binary variables, and between numeric and binary variables.}
    \label{fig:Corr_ACTG320}
\end{figure}

\begin{figure}[h]
    \centering
    \includegraphics[width=0.9\textwidth]{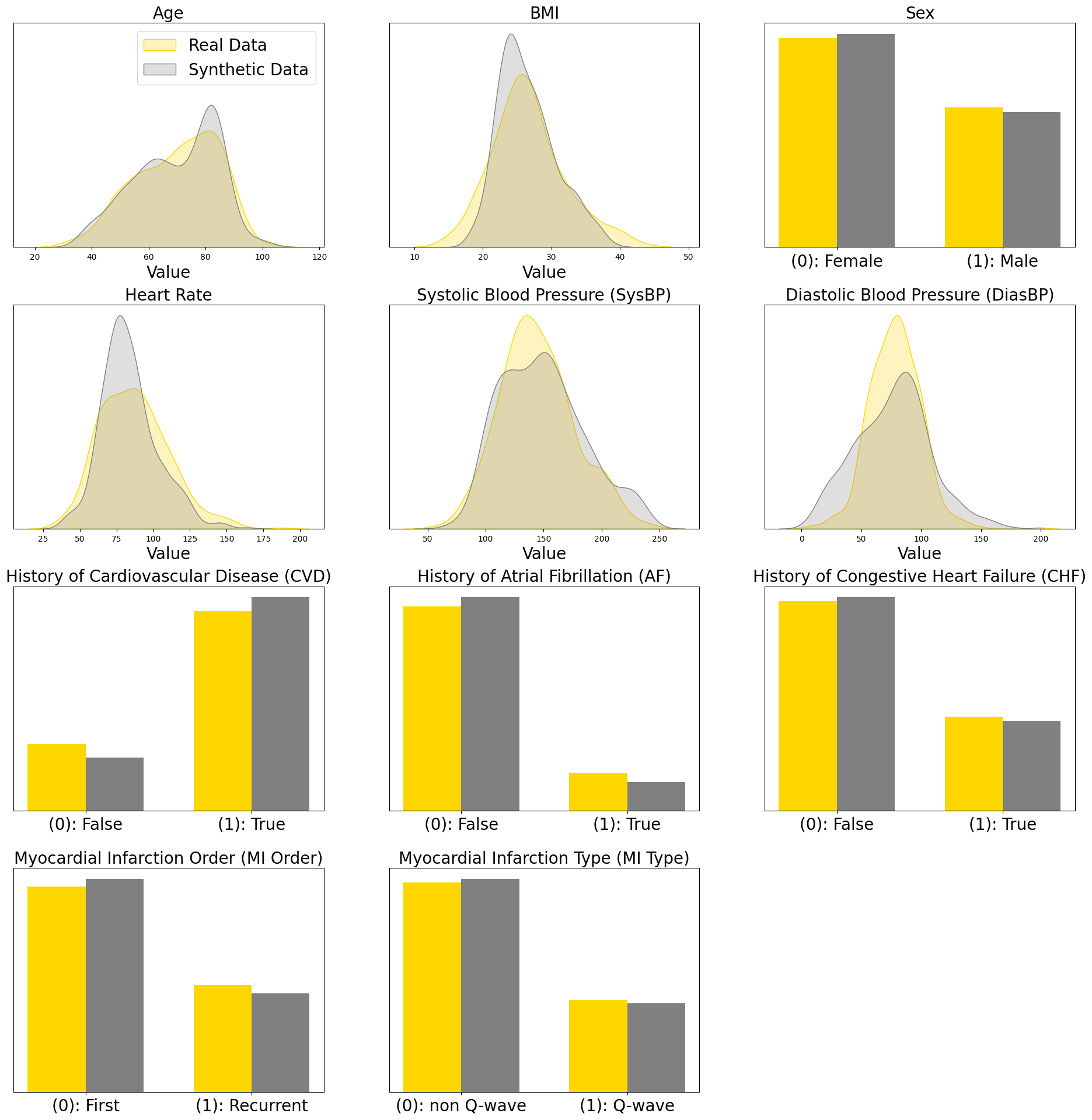}
    \caption{Comparing WHAS500 dataset variables, with real data in gold and synthetic data in grey.}
    \label{fig:Comp_WHAS500}
\end{figure}

\begin{figure}[h]
    \centering
    \includegraphics[width=0.9\textwidth]{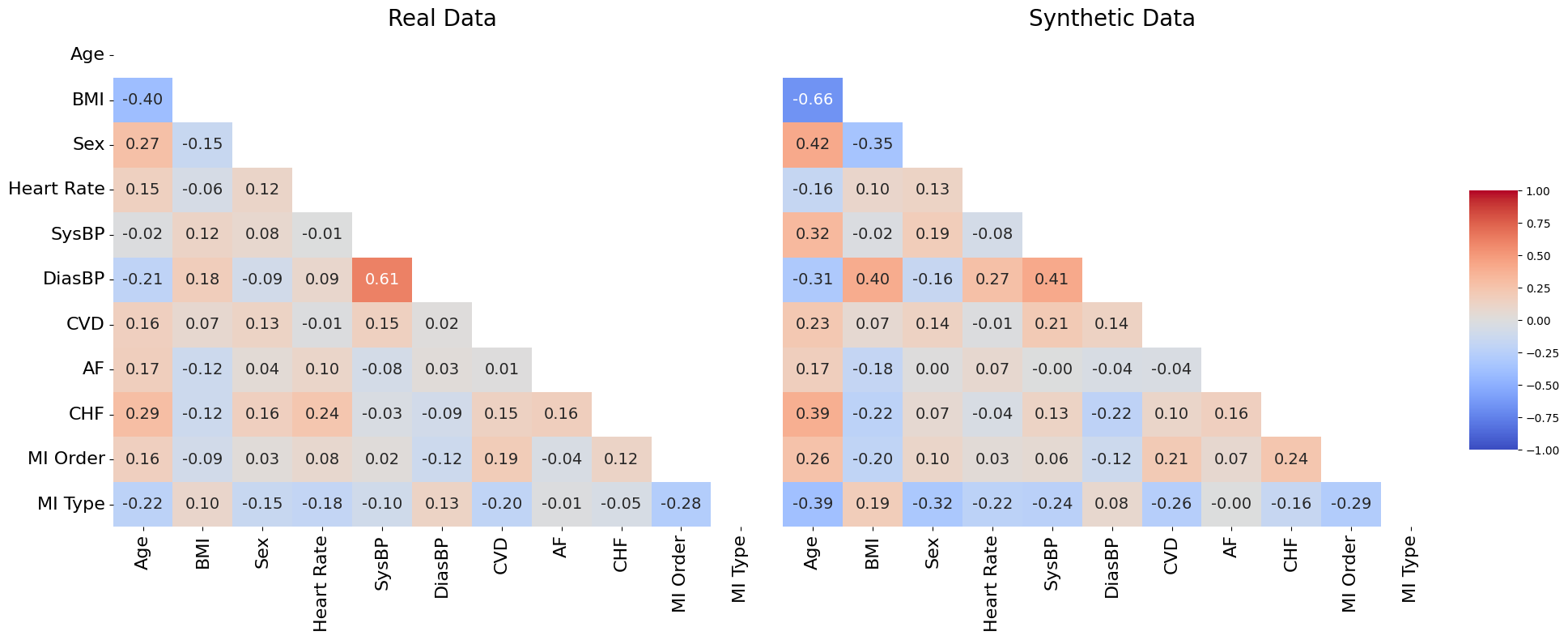}
    \caption{A side-by-side comparison of correlation matrices for the WHAS500 dataset.}
    \label{fig:Corr_WHAS500}
\end{figure}

\begin{figure}[h]
    \centering
    \includegraphics[width=0.9\textwidth]{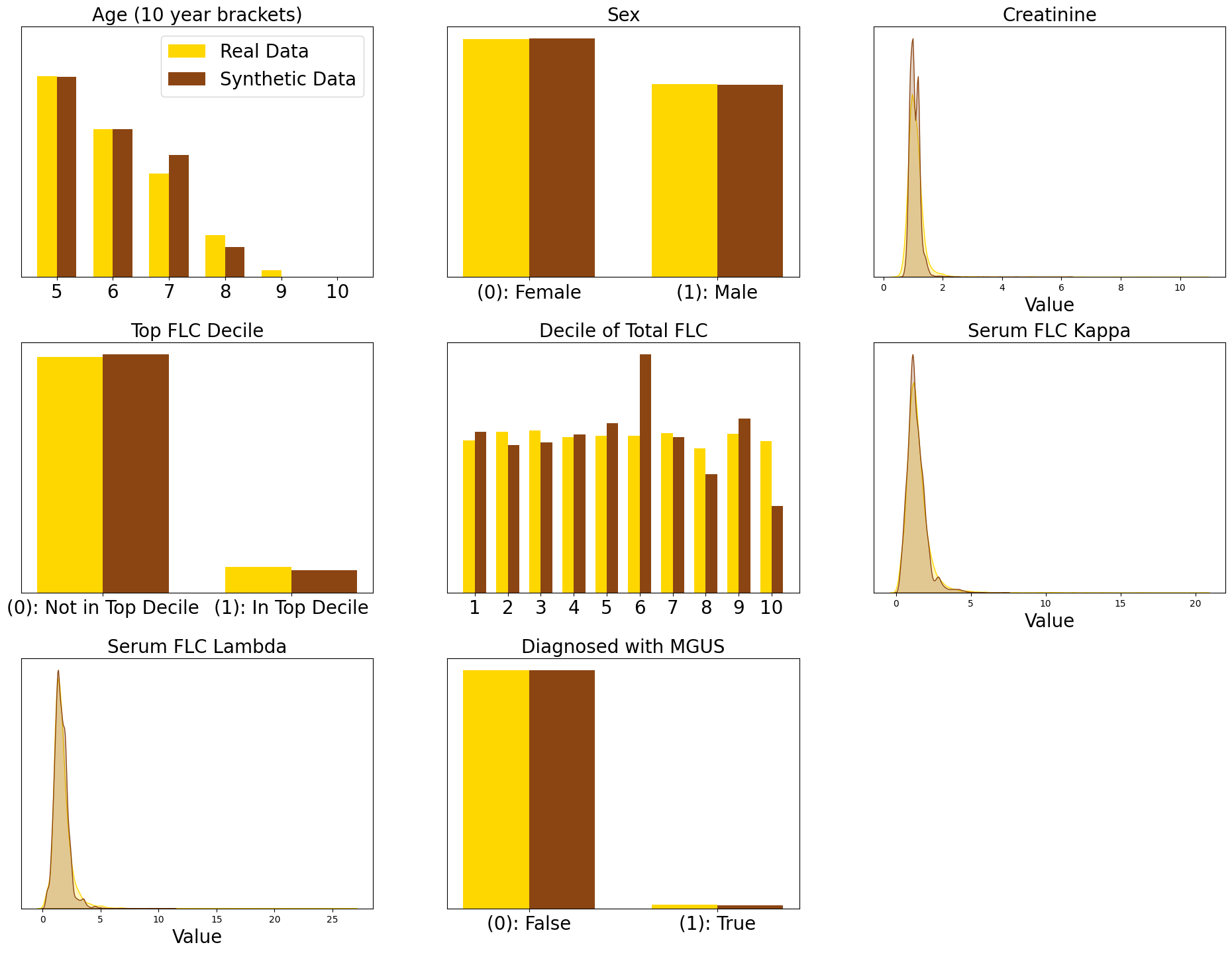}
    \caption{Comparing FLChain dataset variables, with real data in gold and synthetic data in brown.}\label{fig:Comp_FLChain}
\end{figure}

\begin{figure}[h]
    \centering
    \includegraphics[width=0.9\textwidth]{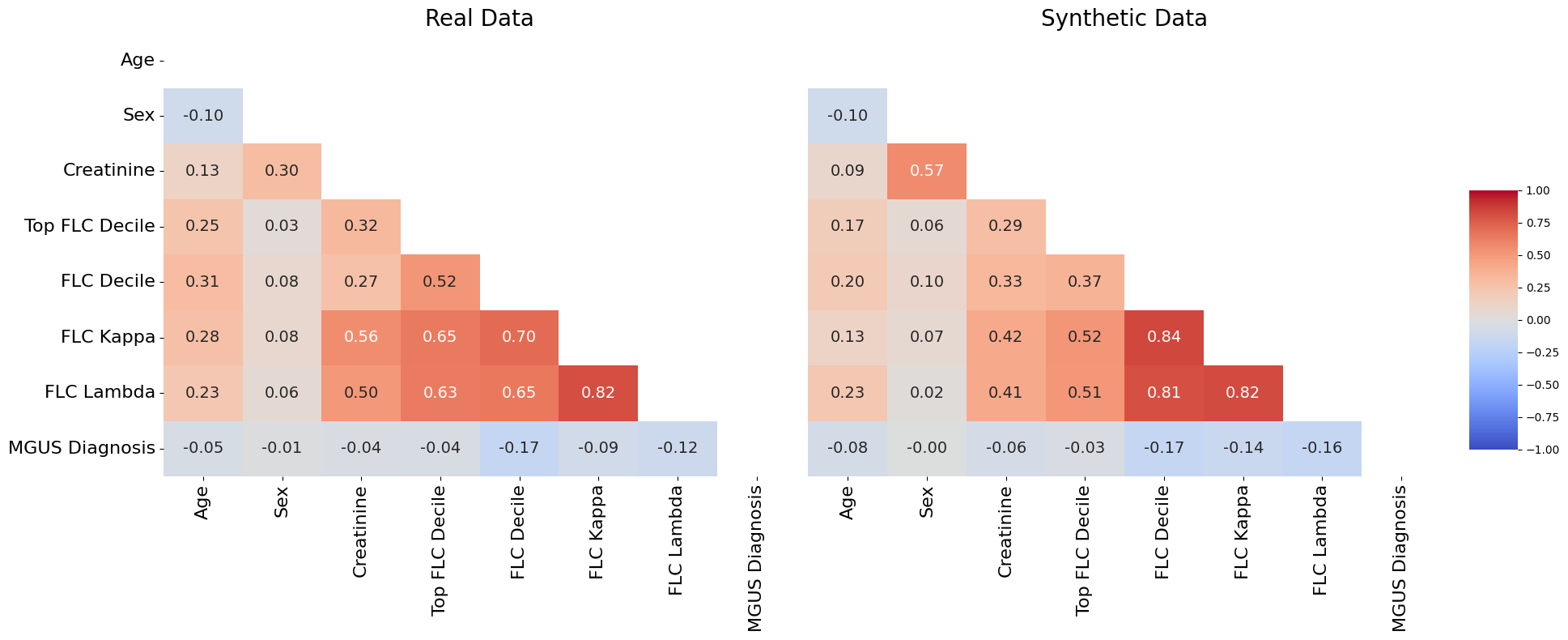}
    \caption{A side-by-side comparison of correlation matrices for the FLChain dataset.}
    \label{fig:Corr_FLChain}
\end{figure}

Similar to the results observed in the synthetic GBSG2 dataset, the distributions and correlations in the synthetic ACTG320, WHAS500, and FLChain datasets are also highly realistic. Figures \ref{fig:Comp_ACTG320}, \ref{fig:Comp_WHAS500}, and \ref{fig:Comp_FLChain} display the distributions for ACTG320, WHAS500, and FLChain, respectively; and Figures \ref{fig:Corr_ACTG320}, \ref{fig:Corr_WHAS500}, and \ref{fig:Corr_FLChain} present the corresponding correlations. These results demonstrate that CK4Gen is scalable across different medical conditions and can generate datasets with both balanced and imbalanced binary variables, as well as numeric variables with extreme long-tail distributions.

Note that KDEs use Gaussian kernels, which can sometimes display strictly positive variables with negative values as an artifact of the method~\cite{crossvalidated_kde_negative_2020}. For accurate statistics, readers should refer to Tables \ref{Tab:GBSG2} -- \ref{Tab:FLChain}. We would like to elaborate on some of the visualisation choices. Although Functional Impairment in the ACTG320 dataset, and Age and Decile of Total FLC in the FLChain dataset, are numeric variables, we visualised them as side-by-side histograms due to their ordinal nature. 

\newpage

\newpage

\newpage

\newpage

\newpage

\newpage
\subsection{Data Augmentation Assessment}

The objective of this phase is to verify that CK4Gen can also function effectively as a data augmentation technique. We establish baseline results using CoxPH models trained on the ground truth data with a 5x2 cross-validation setup~\cite{dietterich1998approximate}. For each validation fold, we augment the CoxPH model with additional data from the synthetic datasets. Performance evaluation will be based on risk prediction accuracy using Harrell's C-index~\cite{harrell1982evaluating} and risk prediction precision using the calibration slope~\cite{spiegelhalter1986probabilistic}.

We further compare CK4Gen's performance in data augmentation against nine alternative techniques: three well-known oversampling methods (synthetic minority oversampling technique [SMOTE]~\cite{chawla2002smote}, and the SMOTE-based extensions of ADASYN~\cite{he2008adasyn}, SVMSMOTE~\cite{nguyen2011borderline}), three undersampling methods (random undersampling, Tomek links~\cite{tomek1976two}, edited neighborhood cleaning rules [NCR]~\cite{laurikkala2001improving, wilson1972asymptotic}), ensemble CoxPH models, and two influential generative models (variational autoencoders [VAEs]~\cite{kingma2014auto} and Wasserstein generative adversarial networks [WGANs]~\cite{kuo2022health, arjovsky2017wasserstein, gulrajani2017improved}).

Detailed descriptions of all techniques, and their setups, can be found in Supplementary Section G.

\subsubsection{Discrimination Analysis Using Harrell's C-index}

\begin{table}[h]
\small
\centering
\begin{tabular}{|p{3cm}|l|l|l|l|}
\hline
\textbf{Techniques} & \textbf{GBSG2} & \textbf{ACTG320} & \textbf{WHAS500} & \textbf{FLChain} \\ 
\hline
\hline
\textbf{No Synthetic Data} & 0.6873 (0.0148) & 
0.6967 (0.0256) & 
0.7642 (0.0222) & 
0.7719 (0.0458) \\ 
\hline
\multicolumn{5}{l}{\textit{\textbf{Oversampling}}} \\ 
\hline
\textbf{SMOTE} & \cellcolor{blue!25}
0.6973 (0.0147) & \cellcolor{red!25}
0.6518 (0.0312) & \cellcolor{blue!25}
0.7651 (0.0217) & \cellcolor{red!25}
0.7680 (0.0618) \\
\hline
\textbf{ADASYN} & \cellcolor{blue!25}
0.6974 (0.0144) & \cellcolor{red!25}
0.6520 (0.0307) & \cellcolor{blue!25}
0.7650 (0.0213) & \cellcolor{red!25}
0.7539 (0.0690) \\ 
\hline
\textbf{SVMSMOTE} & \cellcolor{blue!25}
0.6976 (0.0142) & \cellcolor{red!25}
0.6514 (0.0305) & \cellcolor{blue!25}
0.7656 (0.0212) & \cellcolor{red!25}
0.7537 (0.0692) \\ 
\hline
\multicolumn{5}{l}{\textit{\textbf{Undersampling}}} \\ 
\hline
\textbf{Random Undersample} & \cellcolor{red!25}
0.6866 (0.0171) & \cellcolor{red!25}
0.6659 (0.0367) & \cellcolor{red!25}
0.7627 (0.0232) & \cellcolor{blue!25}
\textbf{0.7881 (0.0057)} \\ 
\hline
\textbf{Tomek Links} & \cellcolor{blue!25}
0.6880 (0.0157) & \cellcolor{red!25}
0.6960 (0.0263) & \cellcolor{red!25}
0.7634 (0.0221) & \cellcolor{red!25}
0.6392 (0.1946) \\ 
\hline
\textbf{NCR} & \cellcolor{red!25}
0.6866 (0.0131) & \cellcolor{red!25}
0.6915 (0.0257) & \cellcolor{red!25}
0.7593 (0.0229) & \cellcolor{blue!25}
0.7875 (0.0057) \\ 
\hline
\multicolumn{5}{l}{\textit{\textbf{Advanced Methods}}} \\ 
\hline
\textbf{Ensemble} & \cellcolor{red!25}
0.6869 (0.0180) & \cellcolor{blue!25}
0.6971 (0.0263) & \cellcolor{blue!25}
0.7656 (0.0232) & \cellcolor{red!25}
0.7015 (0.1414) \\ 
\hline
\textbf{VAE} & \cellcolor{blue!25}
0.6899 (0.0162) & \cellcolor{blue!25}
0.7094 (0.0294) & \cellcolor{blue!25}
0.7692 (0.0151) & \cellcolor{blue!25}
0.7860 (0.0059) \\ 
\hline
\textbf{WGAN} & \cellcolor{red!25}
0.6699 (0.0117) & \cellcolor{blue!25}
0.7038 (0.0331) & \cellcolor{blue!25}
0.7651 (0.0187) & \cellcolor{red!25}
0.7441 (0.1212) \\ 
\hline
\multicolumn{5}{l}{\textit{\textbf{Ours}}} \\ 
\hline
\textbf{CK4Gen} & \cellcolor{blue!25}
\textbf{0.7009 (0.0151)} & \cellcolor{blue!25}
\textbf{0.7156 (0.0253)} & \cellcolor{blue!25}
\textbf{0.7710 (0.0195)} & \cellcolor{blue!25}
0.7870 (0.0057) \\ \hline
\end{tabular}
\caption{A comparison of the data augmentation techniques using Harrell's C-index (the higher the better). The bolded values refer to the best scores achieved within each dataset. The cells are color-coded to compare each technique's performance against the baseline (No Synthetic Data): blue denotes better performances while red indicates worse performances.}
\label{tab:C_score_comparison}
\end{table}

Harrell’s C-index is a core metric in survival analysis, analogous to accuracy in classification but tailored to rank individuals by their event risk. It assesses the proportion of correctly ranked pairs, where a shorter survival time corresponds to a higher predicted risk. If the model ranks two individuals correctly, the pair is concordant; if not, it is discordant. The C-index ranges from 0 (poor discrimination) to 1 (perfect discrimination), with 0.5 indicating random chance. For further details, refer to Supplementary Section H for more details on the C-index. The C-index scores for all datasets across various techniques are summarised in Table \ref{tab:C_score_comparison}.

Across all datasets, CK4Gen's synthetic datasets consistently outperformed alternative methods. CK4Gen enabled the downstream CoxPH model to achieve the highest C-index scores on GBSG2, ACTG320, and WHAS500, and comparable results to the best-performing method, random undersampling, on FLChain.

Oversampling methods improved the performance of the CoxPH model on datasets such as GBSG2 and WHAS500 but underperformed on ACTG320 and FLChain. Synthetic data generated by WGAN did not consistently boost CoxPH model accuracy. Among the tested approaches, VAE was the most reliable, consistently increasing C-index scores across all datasets, demonstrating its robustness. In contrast, undersampling techniques, which eliminate data near decision boundaries to create a more generalised model, resulted in slightly lower, but not drastically different, C-index scores. Ensemble methods of CoxPH did not consistently outperform single CoxPH models in these tests.

\newpage
\subsubsection{Calibration Analysis Using Calibration Slopes}\label{Sec:Calibration}

\begin{figure}[h]
    \centering
    \begin{minipage}[b]{0.45\textwidth}
        \centering
        \includegraphics[width=\textwidth]{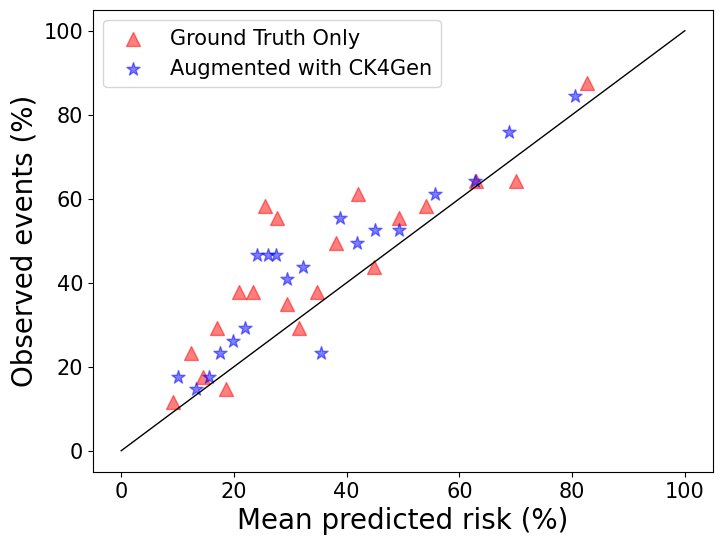}
        \subcaption{GBSG2}
    \end{minipage}
    \hfill
    \begin{minipage}[b]{0.45\textwidth}
        \centering
        \includegraphics[width=\textwidth]{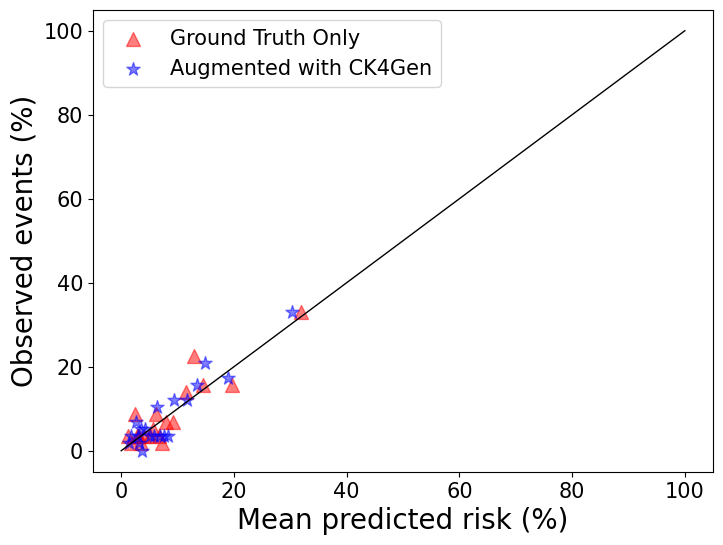}
        \subcaption{ACTG320}
    \end{minipage}

    \vspace{0.4cm} 
    \begin{minipage}[b]{0.45\textwidth}
        \centering
        \includegraphics[width=\textwidth]{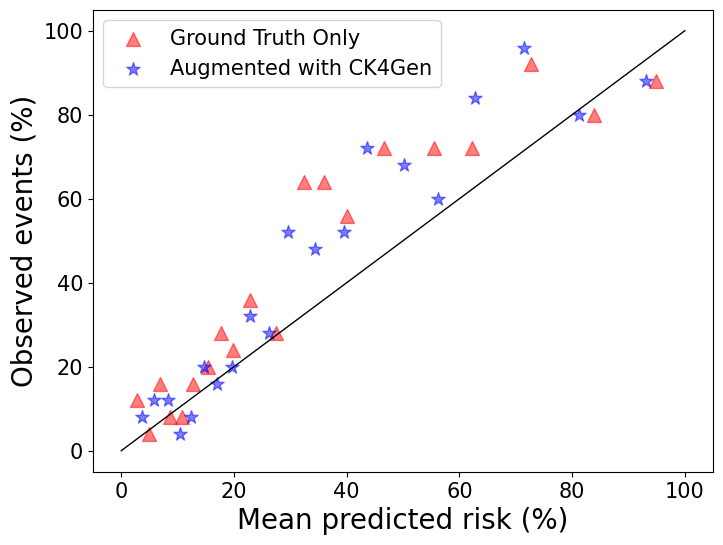}
        \subcaption{WHAS500}
    \end{minipage}
    \hfill
    \begin{minipage}[b]{0.45\textwidth}
        \centering
        \includegraphics[width=\textwidth]{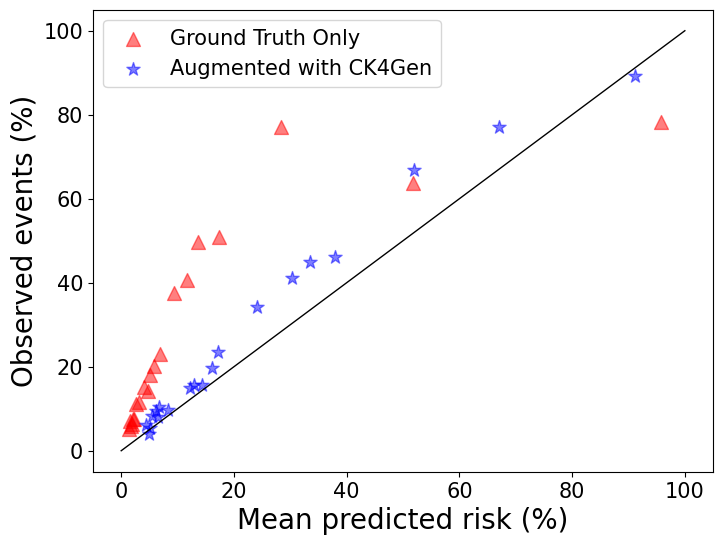}
        \subcaption{FLChain}
    \end{minipage}

    \caption{Calibration curves for different datasets showing the performance of CoxPH models trained on ground truth data only (red triangles) versus models augmented with CK4Gen synthetic data (blue stars). The specific time for each dataset corresponds to the median follow-up time. CK4Gen consistently improves calibration, bringing the curves closer to the ideal diagonal line.}
    \label{fig:CalibrationCurves}
\end{figure}

Calibration in survival analysis evaluates how well predicted event probabilities match observed outcomes. It measures how precisely predicted risks reflect actual event probabilities, minimising misclassifications. This can be visualised using a calibration curve, which plots predicted risk against observed risk at a specific time point. Ideally, the curve follows the 45-degree line, indicating perfect calibration. For more details including a formal formulation, refer to Supplementary Section H.

Figure \ref{fig:CalibrationCurves} shows calibration across datasets at the median follow-up time, contrasting CoxPH models trained on ground truth data alone with those augmented by CK4Gen synthetic data. For all datasets, the CK4Gen-augmented models consistently show better alignment with the 45-degree line, indicating improved calibration.

We quantify calibration deviation using $\mathscr{D}_{\Delta1}$, the distance of the slope from 1, where lower values indicate better calibration. This metric allows for consistent comparisons across models, datasets, and time points. Since disease progression alters risk factors over time, evaluating $\mathscr{D}_{\Delta1}$ at multiple follow-up periods is essential. We assess $\mathscr{D}_{\Delta1}$ at the 25\textsuperscript{th}, 50\textsuperscript{th}, and 75\textsuperscript{th} percentiles of follow-up, ensuring reliable risk predictions throughout the course of patient care. We exclude undersampling and ensemble methods from further analysis, as they had performed poorly in discrimination tests.

Tables \ref{tab:Calibration_Slope_comparison_25_percentile} -- \ref{tab:Calibration_Slope_comparison_75_percentile} demonstrate that CK4Gen improves calibration over longer follow-up periods. Although its early-stage performance at the 25th percentile is suboptimal, leading to inaccuracies, it becomes increasingly effective with extended follow-ups. In contrast, generative ML models and traditional oversampling techniques show instability, performing inconsistently across different time points. This inconsistency poses challenges in clinical applications, where reliable predictions across time are essential. Improved discrimination hence does not guarantee better calibration.

\newpage
\begin{table}[h]
\small
\centering
\begin{tabular}{|p{3cm}|l|l|l|l|}
\hline
\textbf{Techniques} & \textbf{GBSG2} & \textbf{ACTG320} & \textbf{WHAS500} & \textbf{FLChain} \\ 
\hline
\hline
\textbf{No Synthetic Data} & 0.2064 (1.2064) & 
0.0942 (1.0942) & 
0.1949 (1.1949) & 
\textbf{0.0132 (0.9868)} \\ 
\hline
\multicolumn{5}{l}{\textit{\textbf{Oversampling}}} \\ 
\hline
\textbf{SMOTE} & \cellcolor{blue!25}
\textbf{0.0043 (1.0043)} & \cellcolor{red!25}
0.1888 (0.8112) & \cellcolor{red!25}
0.3459 (1.3459) & \cellcolor{red!25}
0.5113 (1.5113) \\ 
\hline
\textbf{ADASYN} & \cellcolor{blue!25}
0.0598 (1.0598) & \cellcolor{red!25}
0.1709 (0.8291) & \cellcolor{red!25}
0.4054 (1.4054) & \cellcolor{red!25}
0.5142 (1.5142) \\ 
\hline
\textbf{SVMSMOTE} & \cellcolor{blue!25}
0.0262 (1.0262) & \cellcolor{blue!25}
\textbf{0.0704 (0.9296)} & \cellcolor{red!25}
0.3805 (1.3805) & \cellcolor{red!25}
0.4990 (1.4990) \\ 
\hline
\multicolumn{5}{l}{\textit{\textbf{Advanced Methods}}} \\ 
\hline
\textbf{VAE} & \cellcolor{red!25}
1.2123 (2.2123) & \cellcolor{red!25}
0.3465 (1.3465) & \cellcolor{blue!25}
\textbf{0.1882 (1.1882)} & \cellcolor{red!25}
0.3197 (0.6803) \\ 
\hline
\textbf{WGAN} & \cellcolor{blue!25}
0.1453 (0.8547) & \cellcolor{red!25}
0.3287 (1.3287) & \cellcolor{red!25}
0.2023 (1.2023) & \cellcolor{red!25}
0.7933 (1.7933) \\ 
\hline
\multicolumn{5}{l}{\textit{\textbf{Ours}}} \\ 
\hline
\textbf{CK4Gen} & \cellcolor{red!25}
0.3949 (1.3949) & \cellcolor{red!25}
0.2071 (1.2071) & \cellcolor{red!25}
0.3211 (1.3211) & \cellcolor{red!25}
0.2328 (1.2328) \\ 
\hline
\end{tabular}
\caption{Calibration comparison across different techniques for the 25\textsuperscript{th} percentile of duration. Values represent $\mathscr{D}_{\Delta1}$, with the corresponding calibration slope in parentheses. Cells are color-coded relative to the baseline (No Synthetic Data): blue indicates better performance, while red indicates worse performance. Bolded values highlight the best-performing techniques within each dataset. CK4Gen showed limitations in prediction precision for the 25\textsuperscript{th} percentile of duration.
\label{tab:Calibration_Slope_comparison_25_percentile}}
\end{table}

\begin{table}[h]
\small
\centering
\begin{tabular}{|p{3cm}|l|l|l|l|}
\hline
\textbf{Techniques} & \textbf{GBSG2} & \textbf{ACTG320} & \textbf{WHAS500} & \textbf{FLChain} \\ 
\hline
\hline
\textbf{No Synthetic Data} & 0.1738 (0.8262) & 
0.1223 (0.8777) & 
0.0517 (1.0517) & 
0.1286 (0.8714) \\ 
\hline
\multicolumn{5}{l}{\textit{\textbf{Oversampling}}} \\ 
\hline
\textbf{SMOTE} & \cellcolor{red!25}
0.2801 (0.7199) & \cellcolor{red!25}
0.3539 (0.6461) & \cellcolor{red!25}
0.1696 (1.1696) & \cellcolor{red!25}
0.1663 (1.1663) \\ 
\hline
\textbf{ADASYN} & \cellcolor{red!25}
0.2457 (0.7543) & \cellcolor{red!25}
0.3399 (0.6601) & \cellcolor{red!25}
0.2115 (1.2115) & \cellcolor{red!25}
0.1779 (1.1779) \\ 
\hline
\textbf{SVMSMOTE} & \cellcolor{red!25}
0.2674 (0.7326) & \cellcolor{red!25}
0.2631 (0.7369) & \cellcolor{red!25}
0.1940 (1.1940) & \cellcolor{red!25}
0.1627 (1.1627) \\ 
\hline
\multicolumn{5}{l}{\textit{\textbf{Advanced Methods}}} \\ 
\hline
\textbf{VAE} & \cellcolor{red!25}
0.4256 (1.4256) & \cellcolor{blue!25}
0.0745 (1.0745) & \cellcolor{blue!25}
\textbf{0.0507 (1.0507)} & \cellcolor{red!25}
0.4881 (0.5119) \\ 
\hline
\textbf{WGAN} & \cellcolor{red!25}
0.3705 (0.6295) & \cellcolor{blue!25}
0.0588 (1.0588) & \cellcolor{red!25}
0.0638 (1.0638) & \cellcolor{red!25}
0.3561 (1.3561) \\ 
\hline
\multicolumn{5}{l}{\textit{\textbf{Ours}}} \\ 
\hline
\textbf{CK4Gen} & \cellcolor{blue!25}
\textbf{0.0603 (0.9397)} & \cellcolor{blue!25}
\textbf{0.0352 (0.9648)} & \cellcolor{red!25}
0.1427 (1.1427) & \cellcolor{blue!25}
\textbf{0.0008 (0.9992)} \\ 
\hline
\end{tabular}
\caption{Calibration comparison across different techniques for the 50\textsuperscript{th} percentile of duration. CK4Gen showed mixed results, but generally performs better than its competitors.}
\label{tab:Calibration_Slope_comparison_50_percentile}
\end{table}

\begin{table}[h]
\small
\centering
\begin{tabular}{|p{3cm}|l|l|l|l|}
\hline
\textbf{Techniques} & \textbf{GBSG2} & \textbf{ACTG320} & \textbf{WHAS500} & \textbf{FLChain} \\ 
\hline
\hline
\textbf{No Synthetic Data} & 0.2667 (0.7333) & 
0.1760 (0.8240) & 
0.0306 (0.9694) & 
0.1469 (0.8531) \\ 
\hline
\multicolumn{5}{l}{\textit{\textbf{Oversampling}}} \\ 
\hline
\textbf{SMOTE} & \cellcolor{red!25}
0.3441 (0.6559) & \cellcolor{red!25}
0.3955 (0.6045) & \cellcolor{red!25}
0.0680 (1.0680) & \cellcolor{blue!25}
0.0953 (1.0953) \\ 
\hline
\textbf{ADASYN} & \cellcolor{red!25}
0.3139 (0.6861) & \cellcolor{red!25}
0.3825 (0.6175) & \cellcolor{red!25}
0.0985 (1.0985) & \cellcolor{blue!25}
0.1122 (1.1122) \\ 
\hline
\textbf{SVMSMOTE} & \cellcolor{red!25}
0.3340 (0.6660) & \cellcolor{red!25}
0.3115 (0.6885) & \cellcolor{red!25}
0.0850 (1.0850) & \cellcolor{blue!25}
0.0946 (1.0946) \\ 
\hline
\multicolumn{5}{l}{\textit{\textbf{Advanced Methods}}} \\ 
\hline
\textbf{VAE} & \cellcolor{blue!25}
0.2109 (1.2109) & \cellcolor{blue!25}
\textbf{0.0070 (1.0070)} & \cellcolor{blue!25}
0.0207 (0.9793) & \cellcolor{red!25}
0.5169 (0.4831) \\ 
\hline
\textbf{WGAN} & \cellcolor{red!25}
0.4310 (0.5690) & \cellcolor{blue!25}
0.0087 (0.9913) & \cellcolor{blue!25}
\textbf{0.0176 (0.9824)} & \cellcolor{red!25}
0.2774 (1.2774) \\ 
\hline
\multicolumn{5}{l}{\textit{\textbf{Ours}}} \\ 
\hline
\textbf{CK4Gen} & \cellcolor{blue!25}
\textbf{0.1732 (0.8268)} & \cellcolor{blue!25}
0.0955 (0.9045) & \cellcolor{blue!25}
0.0370 (1.0370) & \cellcolor{blue!25}
\textbf{0.0380 (0.9620)} \\ 
\hline
\end{tabular}
\caption{Calibration comparison across different techniques for the 75\textsuperscript{th} percentile of duration. CK4Gen showed good results, always outperforming the baseline.}
\label{tab:Calibration_Slope_comparison_75_percentile}
\end{table}

The uneven performance across models stems from the skewed distribution of event durations in the datasets. CK4Gen preserves these biases by replicating the original distributions, as seen in FLChain, where the median follow-up is 4,302 days, with an interquartile range from 2,852 to 4,773 days. Consequently, CK4Gen generates more synthetic data for longer durations, enhancing later-stage calibration. However, it struggles to capture sufficient variability at early stages. The synthetic data should be applied with caution for augmentation, especially when early-stage predictions are critical.

\newpage
\subsection{Utility Verification}
In this final phase, we assess the utility of CK4Gen’s synthetic datasets by focusing on two critical metrics in survival analysis: Kaplan-Meier survival curves (KM curves)~\cite{kaplan1958nonparametric} and Hazard Ratios (HRs) (see Equation \eqref{Eq:CoxPH}). These metrics are widely used to evaluate temporal survival patterns and the relationships between variables and event risks.

KM curves are essential tools for estimating survival probabilities over time. We present KM curves for both real and synthetic data, focusing on three critical features: slope, variability, and final survival rate. The slope reflects the event rate, with steeper slopes indicating higher event frequencies. Variability captures fluctuations in event occurrences, essential for understanding dynamic risk changes. The final survival rate represents the proportion of subjects surviving at the end of the observation period, crucial for long-term risk projections in chronic disease management. Together, these features assess whether CK4Gen reliably replicates temporal survival patterns for research use.

HRs from CoxPH models offer insight into event risks associated with predictors. We present HR plots for real and synthetic data, focusing on two key aspects: point estimates and confidence interval (CI) spreads. Point estimates indicate the strength of associations, while CI spreads reflect uncertainty. Assessing whether the CI includes 1 is critical -- CIs including 1 suggest no significant effect, while those entirely above or below 1 indicate increased or protective risks. Alignment of HR estimates and CIs between real and synthetic data confirms CK4Gen’s ability to replicate both associations and uncertainties, ensuring reliability for healthcare research.

\subsubsection{Assessing the Utility of CK4Gen}

\begin{figure}[h]
    \centering
    \begin{subfigure}[b]{0.8\textwidth}
        \centering
        \includegraphics[width=\textwidth]{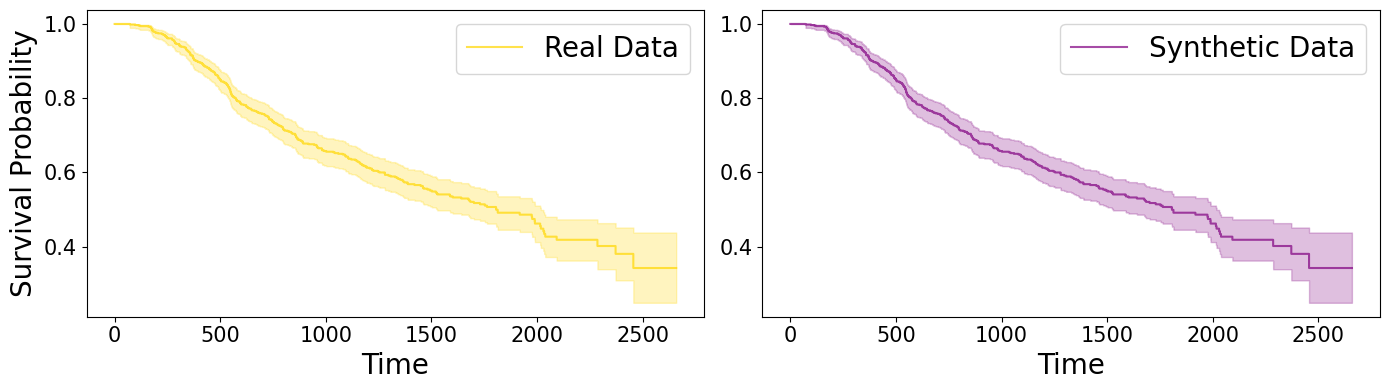}
        \caption{KM curves comparison with CK4Gen}
        \label{fig:KM_GBSG2}
    \end{subfigure}
    
    \vspace{0.5cm} 

    \begin{subfigure}[b]{0.8\textwidth}
        \centering
        \includegraphics[width=\textwidth]{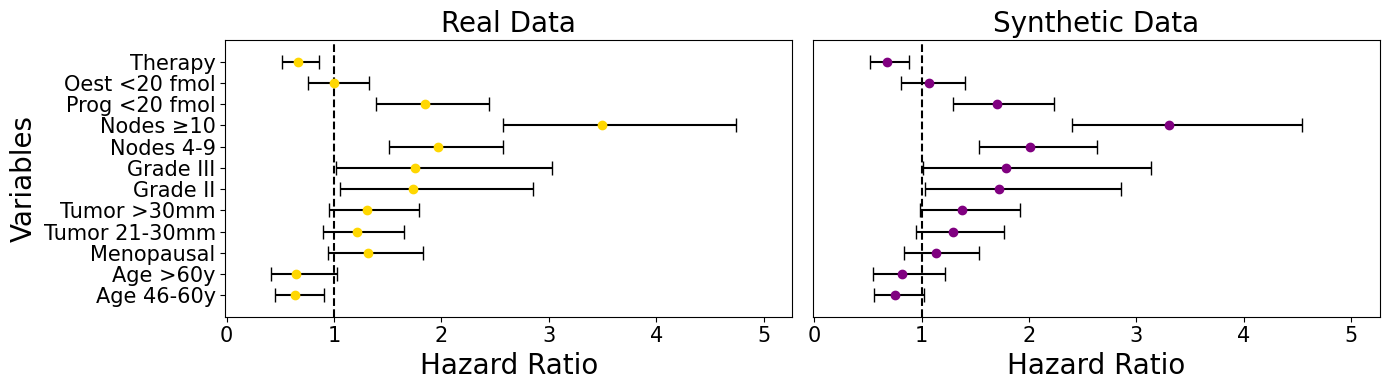}
        \caption{HR consistency comparison with CK4Gen}
        \label{fig:HR_GBSG2}
    \end{subfigure}
    
    \caption{Comparison of KM curves and HR consistencies between real and synthetic data using the GBSG2 dataset. The KM curves evaluate the temporal survival probabilities, while the HRs assess the consistency in risk estimates across key variables.}
    \label{fig:GBSG2_comparison}
\end{figure}

The utility verification for the GBSG2 dataset is illustrated in Figure \ref{fig:GBSG2_comparison}. In subplot (a), the KM curves for real and synthetic data show close alignment, with the gradients, variability, and ending survival probabilities accurately captured by the synthetic data. This demonstrates CK4Gen’s ability to replicate the temporal survival patterns of the real dataset. In subplot (b), the side-by-side comparison of HRs further confirms CK4Gen’s reliability. The point estimates of HRs are closely aligned, indicating that the strength of associations between predictors and survival outcomes is well-preserved. The similar CIs shows that CK4Gen accurately reflects the uncertainty present in real-world data. Most HRs maintain their position relative to the baseline hazard of 1, with only the synthetic variable for age between 46-60 slightly exceeding 1 in its upper CI, indicating a marginal deviation from the real data. Overall, the differences are minimal.

\begin{figure}[h]
    \centering
    \begin{subfigure}[b]{0.8\textwidth}
        \centering
        \includegraphics[width=\textwidth]{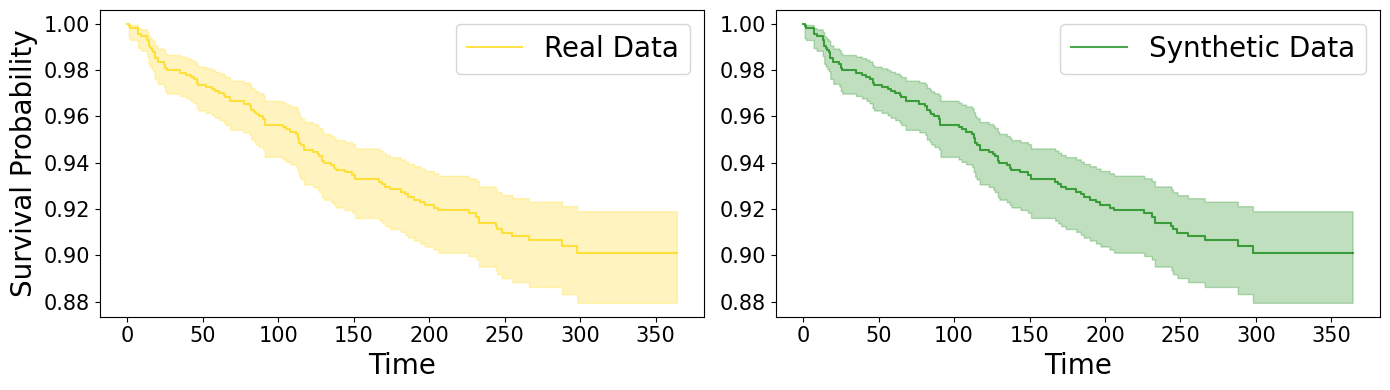}
        \caption{KM curves comparison with CK4Gen}
    \end{subfigure}
    
    \vspace{0.5cm} 

    \begin{subfigure}[b]{0.8\textwidth}
        \centering
        \includegraphics[width=\textwidth]{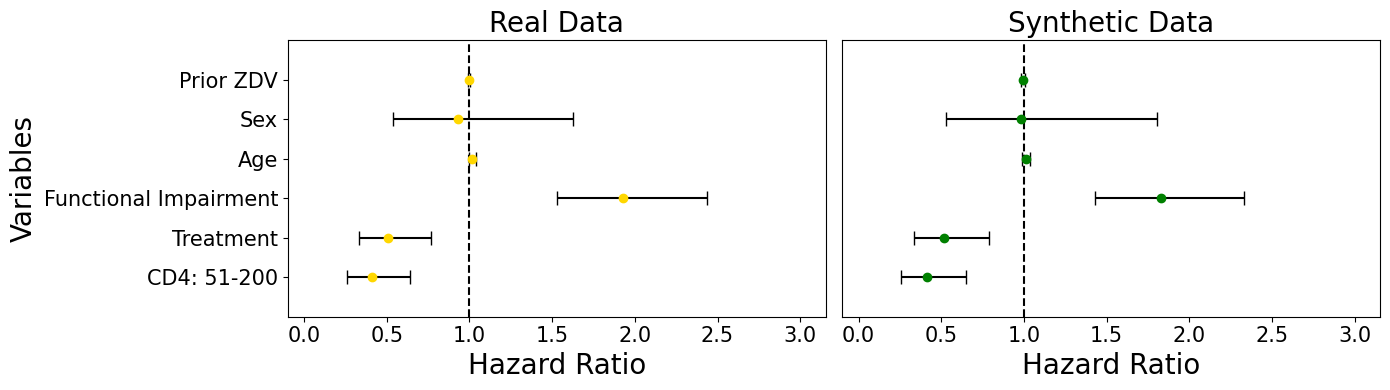}
        \caption{HR consistency comparison with CK4Gen}
    \end{subfigure}
    
    \caption{Comparison of KM curves and HR consistencies between real and synthetic data using the ACTG320 dataset.}
    \label{fig:actg320_comparison}
\end{figure}

\begin{figure}[h]
    \centering
    \begin{subfigure}[b]{0.8\textwidth}
        \centering
        \includegraphics[width=\textwidth]{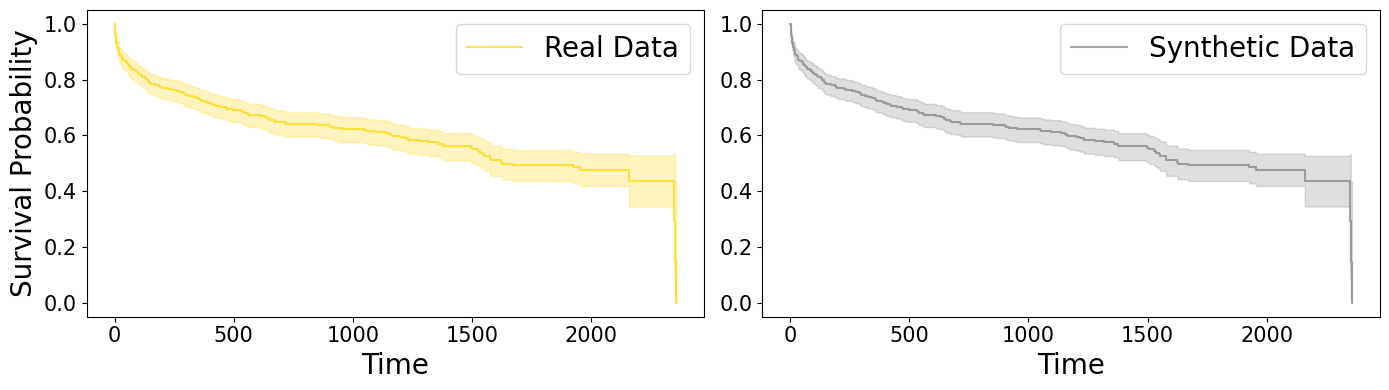}
        \caption{KM curves comparison with CK4Gen}
    \end{subfigure}
    
    \vspace{0.5cm} 

    \begin{subfigure}[b]{0.8\textwidth}
        \centering
        \includegraphics[width=\textwidth]{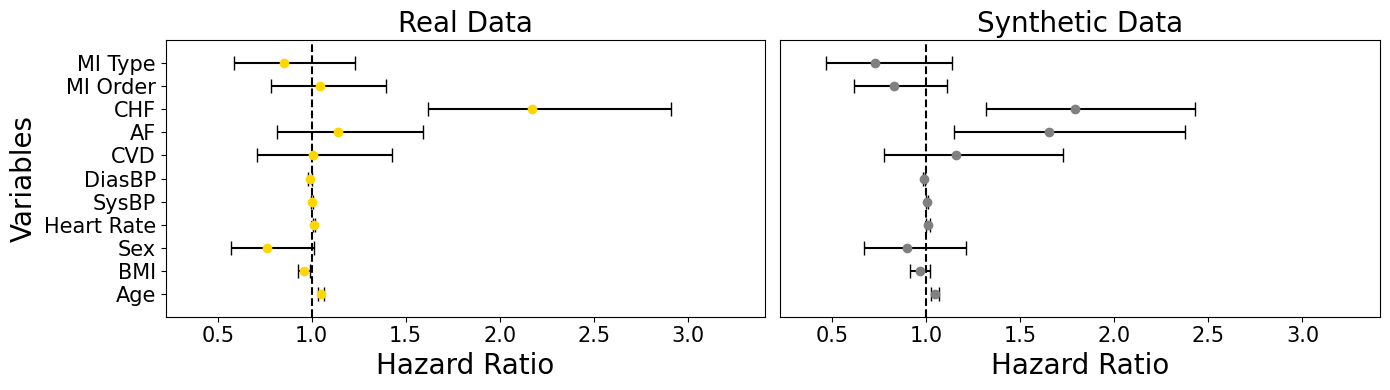}
        \caption{HR consistency comparison with CK4Gen}
    \end{subfigure}
    
    \caption{Comparison of KM curves and HR consistencies between real and synthetic data using the WHAS500 dataset.}
    \label{fig:whas500_comparison}
\end{figure}

\begin{figure}[h]
    \centering
    \begin{subfigure}[b]{0.8\textwidth}
        \centering
        \includegraphics[width=\textwidth]{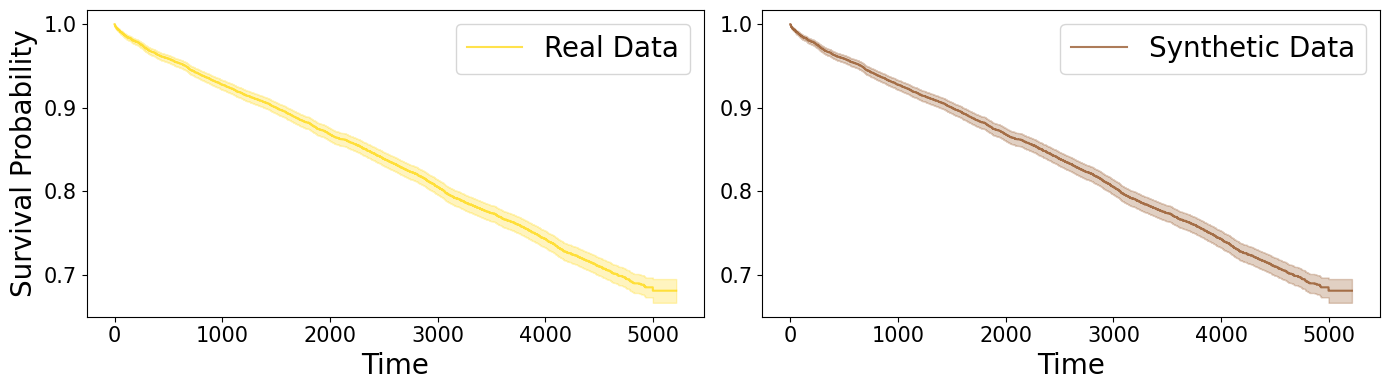}
        \caption{KM curves comparison with CK4Gen}
    \end{subfigure}
    
    \vspace{0.5cm} 

    \begin{subfigure}[b]{0.8\textwidth}
        \centering
        \includegraphics[width=\textwidth]{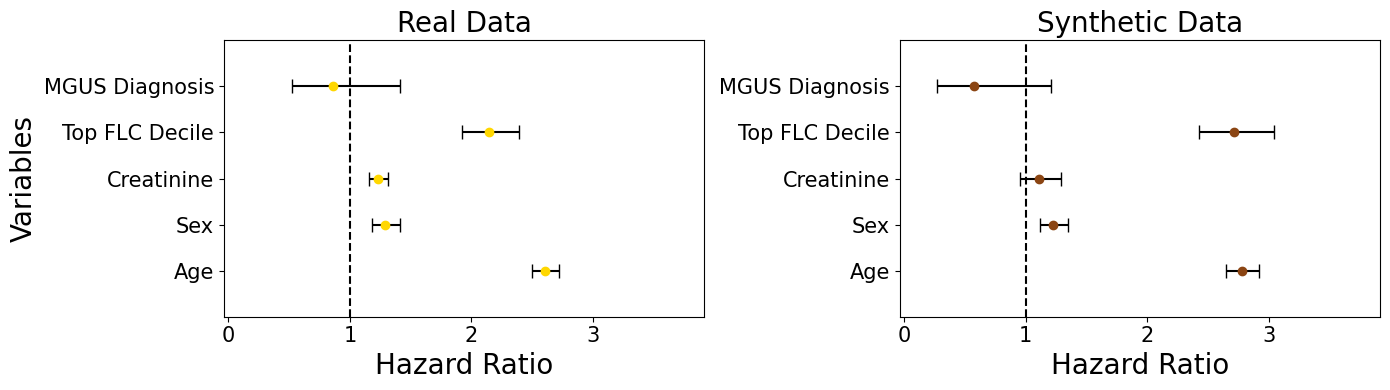}
        \caption{HR consistency comparison with CK4Gen}
    \end{subfigure}
    
    \caption{Comparison of KM curves and HR consistencies between real and synthetic data using the FLChain dataset.}
    \label{fig:flchain_comparison}
\end{figure}

The utility verification for ACTG320, WHAS500, and FLChain datasets is presented in Figures \ref{fig:actg320_comparison} -- \ref{fig:flchain_comparison}. To ensure convergence in the CoxPH model, we excluded variables with the highest collinearity in FLChain. Across all datasets, KM curves demonstrate close alignment between real and synthetic data, with high consistency in HRs. Synthetic datasets replicate HR point estimates accurately, with variability in CIs largely preserved. However, minor discrepancies exist: in WHAS500, Atrial Fibrillation (AF) has a CI including 1 in real data but not in synthetic, and in FLChain, Creatinine shows a similar inconsistency. These rare inconsistencies do not compromise the overall utility and realism of the synthetic datasets, which remain effective for survival and risk analysis.

\newpage

\newpage

\newpage
\subsubsection{Assessing the Utility and Limitations of SMOTE and VAE}

\begin{figure}[h]
    \centering
    \begin{subfigure}[b]{0.8\textwidth}
        \centering
        \includegraphics[width=\textwidth]{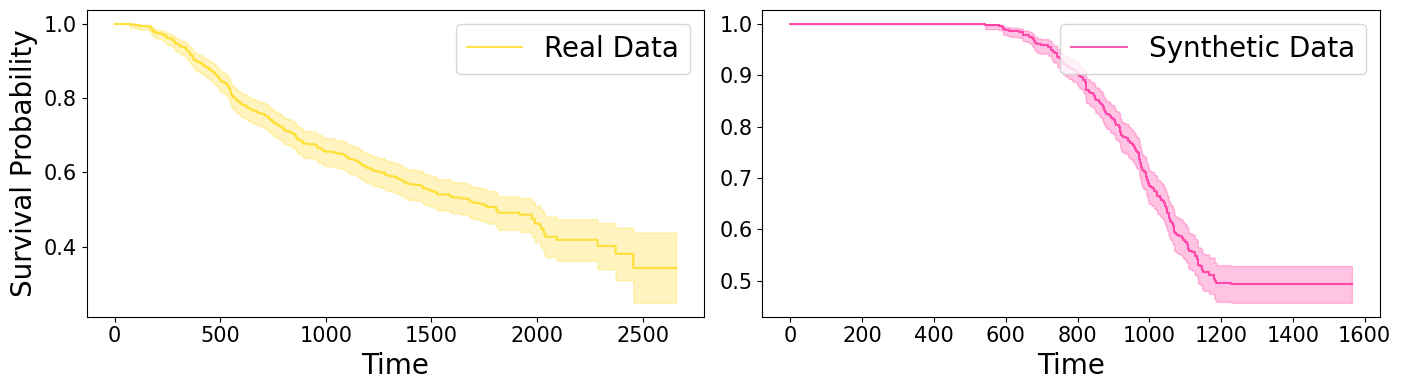}
        \caption{KM curves comparison with SMOTE}
    \end{subfigure}
    
    \vspace{0.5cm} 

    \begin{subfigure}[b]{0.8\textwidth}
        \centering
        \includegraphics[width=\textwidth]{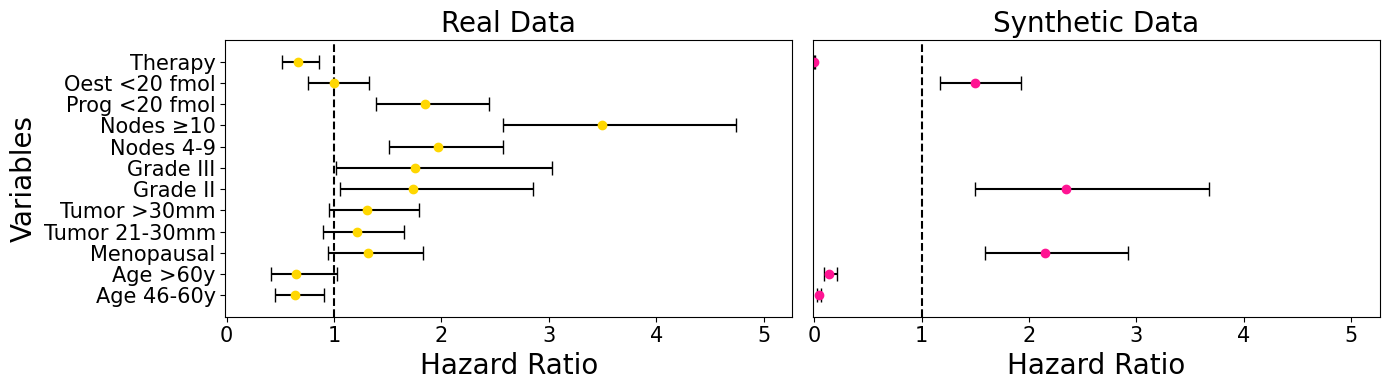}
        \caption{HR consistency comparison with SMOTE}
    \end{subfigure}
    
    \caption{Comparison of KM curves and HR consistencies between real and synthetic data from SMOTE using the GBSG2 dataset.}
    \label{fig:GBSG2_comparison(SMOTE)}
\end{figure}

\begin{figure}[h]
    \centering
    \begin{subfigure}[b]{0.8\textwidth}
        \centering
        \includegraphics[width=\textwidth]{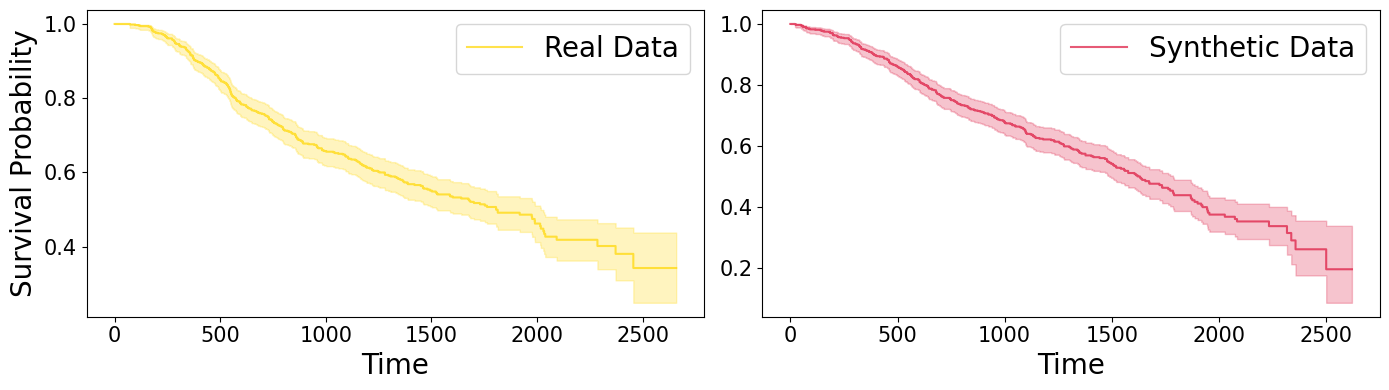}
        \caption{KM curves comparison with VAE}
    \end{subfigure}
    
    \vspace{0.5cm} 

    \begin{subfigure}[b]{0.8\textwidth}
        \centering
        \includegraphics[width=\textwidth]{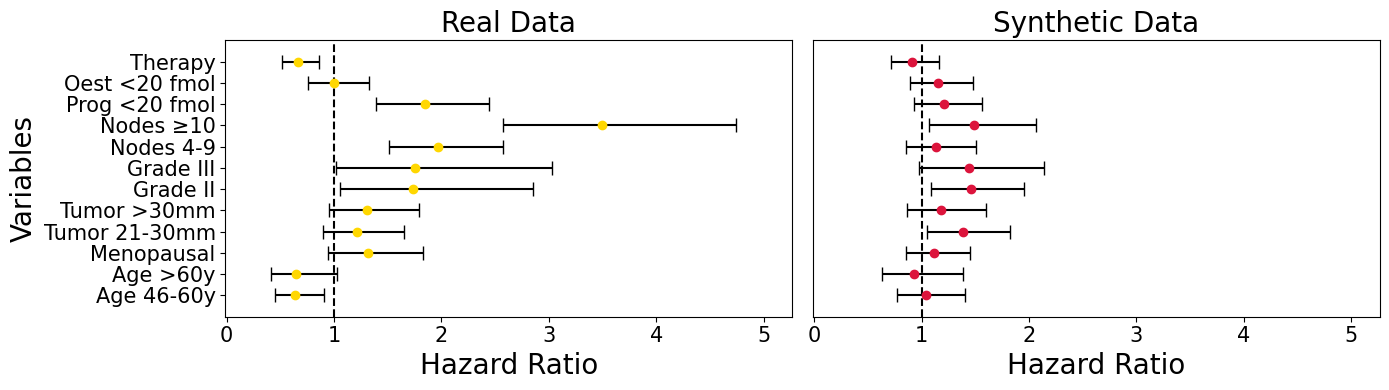}
        \caption{HR consistency comparison with VAE}
    \end{subfigure}
    
    \caption{Comparison of KM curves and HR consistencies between real and synthetic data from VAE using the GBSG2 dataset.}
    \label{fig:GBSG2_comparison(VAE)}
\end{figure}

We found that neither SMOTE nor VAE generated synthetic survival datasets with sufficient utility to replace ground truth data for clinical research. To explore this further, we generated GBSG2 synthetic datasets using both methods and compared their KM curves and HR consistencies in Figures \ref{fig:GBSG2_comparison(SMOTE)} and \ref{fig:GBSG2_comparison(VAE)}. SMOTE struggles with survival data, as evident from notable deviations in its KM curves, particularly in long-term survival probabilities. The steep decline in survival rates highlights SMOTE’s difficulty in capturing key temporal dynamics. HR consistency analysis further reveals significant discrepancies in point estimates and CIs, indicating SMOTE’s limitations in modelling complex risk factors. In extreme cases, HRs for some variables fall outside the plot frame, with the x-axis limits held consistent to emphasise these discrepancies.

While VAE demonstrates better alignment in KM curves, it remains inadequate for survival analysis. Despite improved realism, VAE-generated datasets show substantial gaps in replicating the relationships between risk factors and outcomes, with marked deviations in several HRs. These limitations are especially evident in long-term risk modelling, where accurate reproduction of real-world estimates remains critical. Although VAE shows potential, its current limitations restrict its ability to fully capture the nuanced dynamics required for reliable survival analysis.

For more comparative results across all datasets, refer to Supplementary Section I.

\section{Usage Notes}
Access to real clinical data is often restricted by privacy regulations, making synthetic data a valuable alternative for healthcare research. However, balancing realism, security, and utility remains challenging. To address this, we present CK4Gen (Cox Knowledge for Generation, see Section \ref{Sec:Methods}), a novel machine learning framework that generates high-utility synthetic datasets for survival analysis. CK4Gen is scalable across diverse clinical conditions with minimal modifications, offering a flexible solution for various applications.

A major contribution of this work is the creation of four high-utility synthetic datasets derived from well-established survival datasets (GBSG2, ACTG320, WHAS500, and FLChain, see Section \ref{Sec:DataRecord}). We advance validation practices by not only assessing statistical realism but also evaluating the datasets' impact in survival models through Kaplan-Meier survival curves and hazard ratio consistency (see Section \ref{Sec:TechValid}). This comprehensive validation demonstrates that CK4Gen preserves critical clinical insights; and because the datasets are benchmarked against prior research (see Supplementary Section B), they are ready for immediate use in educational settings~\cite{nicholas2024enriching}.

\underline{Broader Impact}\\
In real-world healthcare applications, high utility in synthetic data is crucial, particularly for survival analysis and treatment planning, where the stakes are high. While methods like SMOTE and VAE may generate data that appears realistic on the surface, focusing on individual distributions and correlations, they fail to capture the complex, multi-dimensional relationships between variables necessary for accurate predictions. As our study shows, these limitations can lead to significant deviations in long-term survival predictions and hazard ratio estimates (see Figures \ref{fig:GBSG2_comparison(SMOTE)} and \ref{fig:GBSG2_comparison(VAE)}), which are critical for clinical decision-making. Inaccurate models based on such synthetic data risk compromising patient safety by providing misleading insights, potentially resulting in harmful treatment recommendations. High-utility synthetic data, as generated by CK4Gen, ensures that these deeper interactions are preserved, providing a reliable foundation for developing models that can be trusted in healthcare practice and regulatory environments.

\underline{Limitations and Future Work}\\
This paper emphasises utility and addresses realism comprehensively, while offering only a limited discussion on security. Since the datasets used are publicly available, immediate privacy concerns are reduced. However, security remains essential, particularly given the potential risks associated with CK4Gen’s autoencoder architecture (see Figure \ref{fig:CK4GenOvervoew}). The risk of reverse-engineering arises if sufficient pseudo-identifiers~\cite{el2020evaluating}, such as chronic conditions, age, or gender, are available, making re-identification feasible. Future work will aim to enhance the security of synthetic data without compromising its utility, ensuring privacy while maintaining CK4Gen’s practicality. The framework already offers significant value for generating synthetic datasets that comply with disclosure controls for public release and for augmenting datasets used within secure environments. Balancing data utility with robust privacy measures remains a priority for further research to align CK4Gen with ethical data practices and regulatory standards.

Dempster, Laird, and Rubin~\cite{dempster1977maximum} noted that \textit{“[...] augmentation process allows for the extraction of complete data from incomplete or missing observations”}. Data augmentation fills gaps within datasets to enhance robustness, especially under scarcity or imbalance, while data synthesis creates entirely new datasets, replicating the statistical properties of the original data. CK4Gen performs well when sufficient real data is available but struggles with sparse datasets (see Section \ref{Sec:Calibration}). For instance, CK4Gen generated realistic data for longer-duration patients in the FLChain dataset (see Table \ref{tab:Calibration_Slope_comparison_75_percentile}), yet faced difficulties in improving calibration for shorter durations due to limited real data variability (see Table \ref{tab:Calibration_Slope_comparison_25_percentile}). This distinction highlights that while synthesis mirrors the original data’s structure, augmentation addresses missingness. Future work may explore conditional data generation~\cite{micheletti2023generative} to improve calibration and enhance diversity within minority patient groups.

CK4Gen’s performance on large datasets, such as state-wide health records, remains to be tested. The smallest dataset we used was WHAS500 with 500 patients, and the largest was FLChain with 7,874 patients. In contrast, cardiovascular risk equations developing using survival analysis in the UK (QRISK4~\cite{hippisley2024development}), the US (PREVENT~\cite{khan2024development}), and New Zealand (PREDICT~\cite{barbieri2022predicting}) were developed using over 9 million, 3 million, and 2 million patient records, respectively. While CK4Gen is likely well-suited for smaller randomised control trials, future research should assess its performance on larger datasets.

\bibliographystyle{IEEEtran}
\bibliography{iclr2023_conference}

\newpage
\appendix
\section*{Supplementary Material}

\hrule  
\vspace{0.05cm}
\hrule
\vspace{0.3cm}  
\textbf{Nicholas I-Hsien Kuo}, \textbf{Blanca Gallego}, \textbf{Louisa Jorm}\\
Centre for Big Data Research in Health, the University of New South Wales, Sydney, Australia\\
\textcolor{white}{*}\\
Corresponding author: Nicholas I-Hsien Kuo (\texttt{n.kuo@unsw.edu.au})

The following supplementary material provides additional details and supporting information for the paper ``CK4Gen: A Knowledge Distillation Framework for Generating High-Utility Synthetic Survival Datasets in Healthcare''. In the main text, we propose a novel framework for generating synthetic survival analysis datasets with a specific focusing on retaining hgih utility. Below, we present additional details and extra experimental outcomes. 

\vspace{0.3cm}
\hrule  
\vspace{0.05cm}
\hrule
\vspace{0.3cm}  

\section{Reproducibility}
\begin{table}[h]
\scriptsize
\centering
\begin{tabular}{|l|l|l|}
\hline
\textbf{Dependency} & \textbf{Details} & \textbf{Category} \\ \hline
\hline
CUDA Version & 12.1 & Hardware \\ \hline
GPU Name & NVIDIA A100 & \\ \hline
\hline
Python Version~\cite{vanRossum1995} & 3.10.12 & General Computation \\ \hline
\hline
Pandas~\cite{mckinney-proc-scipy-2010} & 2.0.3 & Data Science \\ \hline
NumPy~\cite{harris2020array} & 1.25.2 & \\ \hline
scikit-learn~\cite{sklearn_api} & 1.2.2 & \\ \hline
imbalanced-learn~\cite{JMLR:v18:16-365} & 0.9.1 & \\ \hline
\hline
Lifelines~\cite{Davidson-Pilon2019} & 0.28.0 & Survival Analysis \\ \hline
scikit-survival~\cite{polsterl2020scikit} & 0.23.0 & \\ \hline
\hline
PyTorch~\cite{paszke2019pytorch} & 2.3.0+cu121 & Deep Learning \\ \hline
TorchTuples~\cite{torchtuples2024} & 0.2.2 & \\ \hline
\hline
MatPlotLib~\cite{Hunter2007} & 3.7.1 & Visualisation \\ \hline
\end{tabular}
\caption{Hardware and Software Dependencies}
\end{table}

This appendix provides a summary of the hardware and software dependencies utilised in our analysis. Reporting these details is crucial for ensuring the reproducibility of our results. The hardware environment includes an NVIDIA A100 GPU with CUDA version 12.2, running on a Linux platform; and the Python version is 3.10.12. The software dependencies involve several key libraries essential for data manipulation, statistical analysis, machine learning, and deep learning.

We implemented a seeding function within our computational framework to ensure reproducibility. The function is designed to set seeds for various random number generators across multiple libraries and environments. This includes the Python random module, NumPy, and PyTorch, both for CPU and GPU operations. By setting the \texttt{PYTHONHASHSEED} environment variable, we ensure consistency in hash-based operations. Additionally, we enforce deterministic behaviour in PyTorch by setting \texttt{torch.backends.cudnn.deterministic = True} and disable cuDNN benchmarking to prevent the selection of different convolution algorithms during training.

\newpage
\section{Validation and Details of Ground Truth Datasets}\label{App:ReferenceToAllSup}

This appendix provides detailed information on the original datasets used in this study and the validation processes conducted to ensure the reliability of the generated synthetic datasets. The datasets include the German breast cancer study group 2 (GBSG2)~\cite{schumacher1994randomized}, the AIDS clinical trials group study 320 (ACTG320)~\cite{hammer1997controlled}, the Worcester heart attack study 500 (WHAS500)~\cite{goldberg1988incidence}, and the free light chain (FLChain) dataset~\cite{dispenzieri2012use}. 

For each dataset, we present a comprehensive review of the original study, the key clinical variables involved, and the methods used to replicate the survival analysis results from the original papers. This validation process ensures that the synthetic datasets generated by CK4Gen retain the statistical properties and clinical relevance necessary for meaningful research applications. The sections below detail these processes for each dataset.

\subsection{GBSG2}\label{App:MoreDetailsGbsg2}

The dataset from the original study by Schumacher \textit{et al.}~\cite{schumacher1994randomized} was used to develop prognostic models for node-positive breast cancer patients. In the subsequent analysis by Sauerbrei \textit{et al.}~\cite{sauerbrei1999modelling}, the numeric variables were transformed into categorical variables to facilitate analysis and improve interpretability. Age was divided into three categories (\(\leq 45\) years, 45-60 years, \(>60\) years), tumour size into three categories (\(\leq 20\) mm, 21-30 mm, \(>30\) mm), and the number of positive lymph nodes into three categories (\(\leq 3\), 4-9, \(\geq 10\) nodes). Menopausal status was binary (pre-menopausal, post-menopausal), while tumour grade, progesterone receptor level, and oestrogen receptor level were similarly categorised to reflect their ordinal relationship with survival.

Our research is based on a variation of the GBSG2 dataset available from the \texttt{lifelines} package, transformed according to the descriptions in Sauerbrei \textit{et al.}, which we treat as the ground truth dataset. To ensure that the synthetic dataset, synthesised from this ground truth, is valuable for clinical research and medical education, it is crucial that the modelling outcomes based on our ground truth dataset replicate the results in Sauerbrei \textit{et al.}.

Following the description of Sauerbrei \textit{et al.}, we employed a CoxPH model to derive three types of HR-related statistics: the variable coefficients (\(\beta\)), the HR (\(\exp(\beta)\)), and the 95\% CI of the HR. The results from Sauerbrei \textit{et al.} are from their Table 1 (located in Section 3.1 on page 79), specifically the coefficients in the “Full” model of the “Multivariable models”. In our model, we directly incorporated hormonal therapy information, while Sauerbrei \& Royston adjusted their estimates for hormonal treatment without explicitly listing it in the table. We found that our HR-related statistics closely matched those reported in their study. This not only replicates their findings but also suggests that hormonal treatment is not a significant confounder for the common variables in our datasets. We included hormonal treatment in our data synthesis process as it is likely to be investigated in future applications. Refer to results in Table \ref{Tab:AppGBSG2}.

There are some differences between the data available from \texttt{lifelines} and the original study by Schumacher \textit{et al.}. For instance, Schumacher \textit{et al.} had 473 patients, whereas the \texttt{lifelines} dataset includes 686 patients. However, as reflected in the table, almost all predictor variables are consistent. The main discrepancy is the coefficient for age greater than 60 years, where Sauerbrei \textit{et al.} reported -0.78 and ours is -0.43.

\newpage
\begin{table}[h]
\small
\centering
\begin{tabular}{|c|p{2.75cm}|p{2.5cm}|p{2.5cm}|}
\hline
\textbf{Variable} & \textbf{Coefficient\newline (Sauerbrei \textit{et al.)}} & \textbf{Coefficient\newline (Ours)} & \textbf{HR (95\% CI)\newline (Ours)} \\
\hline
\hline
\textbf{Age} & & & \\
\(\leq 45\) years & Baseline & Baseline & Baseline \\
46-60 years & -0.40 & -0.45 & 0.64 (0.45, 0.91) \\
\(>60\) years & -0.78 & -0.43 & 0.65 (0.41, 1.03) \\
\hline
\hline
\textbf{Menopausal State} & & & \\
Pre-menopausal & Baseline & Baseline & Baseline \\
Post-menopausal & 0.27 & 0.27 & 1.31 (0.94, 1.83) \\
\hline
\hline
\textbf{Tumour Size} & & & \\
\(\leq 20\) mm & Baseline & Baseline & Baseline \\
21-30 mm & 0.21 & 0.20 & 1.22 (0.90, 1.65) \\
\(>30\) mm & 0.27 & 0.27 & 1.31 (0.95, 1.79) \\
\hline
\hline
\textbf{Tumour Grade} & & & \\
I & Baseline & Baseline & Baseline \\
II & 0.55 & 0.55 & 1.74 (1.06, 2.85) \\
III & 0.56 & 0.56 & 1.75 (1.02, 3.02) \\
\hline
\hline
\textbf{Number of Positive Nodes} & & & \\
\(\leq 3\) & Baseline & Baseline & Baseline \\
4-9 & 0.68 & 0.68 & 1.97 (1.51, 2.58) \\
\(\geq 10\) & 1.26 & 1.25 & 3.50 (2.57, 4.74) \\
\hline
\hline
\textbf{Progesterone Receptor Level} & & & \\
\(\geq 20\) fmol/mg & Baseline & Baseline & Baseline \\
\(<20\) fmol/mg & 0.61 & 0.61 & 1.85 (1.39, 2.45) \\
\hline
\hline
\textbf{Oestrogen Receptor Level} & & & \\
\(\geq 20\) fmol/mg & Baseline & Baseline & Baseline \\
\(<20\) fmol/mg & 0.01 & 0.00 & 1.00 (0.77, 1.33) \\
\hline
\hline
\textbf{Hormonal Therapy} & & & \\
False & Baseline & Baseline & Baseline \\
True & & -0.41 & 0.67 (0.52, 0.86) \\
\hline
\end{tabular}
\caption{\label{Tab:AppGBSG2}Confirming results using the ground truth GBSG2 dataset.}
\end{table}

\newpage
\subsection{ACTG320}\label{App:MoreDetailsACTG320}
The dataset from the original study by Hammer \textit{et al.}~\cite{hammer1997controlled} was used to compare the efficacy and safety of adding a protease inhibitor (indinavir) to a two-drug regimen in HIV-1 infected patients with no more than 200 CD4 cells per cubic millimetre who had prior zidovudine therapy. An event was defined as the development of an AIDS-defining condition or death. Although their dataset included extensive patient information, the authors focused on treatment groups and CD4 cell count strata as the primary variables because these were the most clinically relevant factors for assessing the efficacy of the new treatment regimen. In Cole \& Stuart’s study~\cite{cole2010generalizing}, the authors generalised the findings from Hammer \textit{et al.} to the 2006 US HIV-infected population. Since Cole \& Stuart aimed to apply their results to a broader population, they included racial background in their modelling process.

Our research is based on a variation of the AIDS dataset available from the \texttt{sksurv} package, which we treat as the ground truth dataset. To ensure that the synthetic dataset, synthesised from this ground truth, is valuable for clinical research and medical education, it is crucial that the modelling outcomes replicate the results in Hammer \textit{et al.}. We fit univariate models validating the HR-related statistics to Hammer \textit{et al.}'s study. The results from Hammer \textit{et al.} are from their Table 2 (located in Section “Progression of Disease” on page 4). The variable included was the treatment indicator, with the baseline as the two-drug regimen and the three-drug regimen as the alternative. Refer to results in Table \ref{Tab:AppACTG320-1}.

There are notable differences between the data available from \texttt{sksurv} and the original study by Hammer \textit{et al.}. While Hammer \textit{et al.} included 1,156 observations, our dataset has 1,151. The primary outcome in both studies was the time to the development of an AIDS-defining event or death, but Hammer \textit{et al.} also included the additional set of outcomes relating to time to death-only events. The main discrepancy is the HR when stratified by CD4 count (51-200), where Hammer \textit{et al.} reported 0.51 (non-significant) and ours is 0.38 (significant). However, the HR results over the entire cohort are similar (0.50 vs 0.50, both significant), as are those for CD4 $\le$ 50 (0.49 vs 0.58, both significant).

Given that the \texttt{sksurv} dataset includes more variables than both Hammer \textit{et al.} and Cole \& Stuart, we included additional variables in our data generation process. This adds complexity and makes the dataset more flexible for different analytical purposes. We ultimately included six features: age, sex, CD4 stratum, treatment indicator, functional impairment, and months of prior ZDV use. Racial background was not included because the ACTG320 dataset has more detailed racial categories than those used by Cole \& Stuart. Selecting specific racial categories according to the setup in Cole \& Stuart’s study would require discarding some data. As Hammer \textit{et al.}'s original study did not include racial background, we also decided to exclude it. However, our CK4Gen model for creating synthetic data is openly available for those who wish to extend beyond our study. In Table \ref{Tab:AppACTG320-2}, we show the results of the HR of a CoxPH modelled with all these selected variables.

Of note, the Karnofsky Performance Scale (KPS)~\cite{karnofsky1948use} was initially recorded as a categorical variable indicating the level of a patient's functioning, ranging from 100 (normal) to 70 (cares for self). For modelling, this scale was transformed into a numeric severity variable, termed "Functional impairment". KPS values were mapped to a severity scale where higher values indicate greater impairment: KPS 100 was mapped to severity 0 (no impairment), KPS 90 to severity 1 (minor impairment), KPS 80 to severity 2 (moderate impairment), and KPS 70 to severity 3 (severe impairment). This new variable allows for a clearer understanding of how the patient's functional status impacts the outcomes of interest.

\newpage
\begin{table}[h]
\small
\centering
\begin{tabular}{|p{4cm}|p{4cm}|p{4cm}|}
\hline
\textbf{Variable} & \textbf{HR (95\% CI)\newline (Hammer \textit{et al.})} & \textbf{HR (95\% CI)\newline (Ours)} \\
\hline
\hline
\textbf{All patients} & & \\
Controlled group & Baseline & Baseline \\
Treatment group & 0.50 (0.33, 0.76) & 0.50 (0.33, 0.77) \\
\hline
\hline
\textbf{\(\leq 50\) cells/mm\(^3\) CD4} & & \\
Controlled group & Baseline & Baseline \\
Treatment group & 0.49 (0.30, 0.82) & 0.58 (0.36, 0.95) \\
\hline
\hline
\textbf{51 – 200 cells/mm\(^3\) CD4} & & \\
Controlled group & Baseline & Baseline \\
Treatment group & 0.51 (0.24, 1.10) & 0.38 (0.16, 0.88) \\
\hline
\end{tabular}
\caption{\label{Tab:AppACTG320-1}Confirming results using the ground truth ACTG320 dataset.}
\end{table}

\begin{table}[h]
\small
\centering
\begin{tabular}{|p{4cm}|p{6cm}|}
\hline
\textbf{Variable} & \textbf{HR (95\% CI) (Ours)} \\
\hline
\hline
\textbf{Age} & 1.02 (1.00, 1.04) \\
\hline
\hline
\textbf{Sex} & \\
Female & Baseline \\
Male & 0.93 (0.53, 1.62) \\
\hline
\hline
\textbf{CD4 Cell Count} & \\
\(\leq 50\) cells/mm\(^3\) & Baseline \\
51 – 200 cells/mm\(^3\) & 0.41 (0.26, 0.64) \\
\hline
\hline
\textbf{Treatment Indicator} & \\
Control group & Baseline \\
Treatment group & 0.51 (0.33, 0.77) \\
\hline
\hline
\textbf{Functional Impairment} & 1.93 (1.53, 2.44) \\
\hline
\hline
\textbf{Months of Prior ZDV Use} & 1.00 (0.99, 1.01) \\
\hline
\end{tabular}
\caption{\label{Tab:AppACTG320-2}The additional information here shows how the HR would look like in the ground truth ACTG320 dataset, if the CoxPH model were to be modelled using more variables as its inputs.}
\end{table}

\newpage
\subsection{WHAS500}\label{App:MoreDetailsWHAS500}
The Worcester Heart Attack Study (WHAS) introduced in Goldberg \textit{et al.}~\cite{goldberg1988incidence} is a longitudinal, population-based study initiated to explore the incidence and survival rates following hospital admission for acute myocardial infarction (AMI). The study spans over 13 one-year periods from 1975 to 2001 and includes data from over 11,000 admissions. The primary objective is to delineate factors associated with AMI incidence and survival, offering insights into heart attack management and outcomes over time.

The WHAS500 dataset was derived for more focused analytical purposes. It includes 500 subjects and retains a core set of variables from the original WHAS. This dataset is utilised extensively for demonstrating and discussing various aspects of modelling time-to-event data in survival analysis, as outlined by Hosmer \textit{et al.}~\cite{hosmer2008applied}. Our research is based on a variation available from the \texttt{sksurv} package, treated as the ground truth dataset. Ensuring that the synthetic dataset, synthesised from this ground truth, is valuable for clinical research and medical education is crucial. The modelling outcomes must replicate results from well-known sources. We follow instructions from UCLA’s Office of Advanced Research Computing (OARC), as detailed on their webpage. Interested readers should search for “Table 6.7 on page 198” on the OARC website of\\
\textcolor{pink}{https://stats.oarc.ucla.edu/sas/examples/asa2/applied-survival-analysis-by-hosmer-lemeshow-and-maychapter-6-assessment-of-model-adequacy/} . 

In the UCLA example, WHAS500 underwent several transformations to facilitate CoxPH analysis. These transformations included creating quadratic and cubic terms of BMI (bmifp1 and bmifp2) by dividing BMI by 10 and then squaring or cubing the result, and rescaling heart rate (hr10) and diastolic blood pressure (diasbp10) by dividing each by 10. An interaction term (ga) was created by multiplying gender and age, while the follow-up time (lenfol) was converted from days to years by dividing by 365.25. The resulting model excluded observations where diastolic blood pressure was 198 and included the predictors bmifp1, bmifp2, age, hr10, diasbp10, gender, chf, and ga. Our validation shows that the variables align with those from the UCLA OARC example. Refer to Table \ref{Tab:ComparisonUCLA} for the HRs of the variables from both the UCLA OARC example and our ground truth dataset.

This UCLA OARC model did not include all variables, and for our data synthesis research we increased the complexity by including additional variables such as age, BMI, gender, heart rate, systolic blood pressure, diastolic blood pressure, history of cardiovascular disease, atrial fibrillation, congestive heart complications, MI order, MI type, follow-up duration, and vital status at last follow-up. Unlike their work, we synthesised the data directly from the raw data without making specific transformations, such as not dividing the diastolic BP by 10. Table \ref{Tab:AppWHAS-2}, we show the results of the HR of a CoxPH modelled with all these selected variables.

\newpage
\begin{table}[h]
\small
\centering
\begin{tabular}{|p{3cm}|p{4cm}|p{4cm}|}
\hline
\textbf{Variable} & \textbf{HR (95\% CI) (UCLA OARC)} & \textbf{HR (95\% CI) (Ours)} \\
\hline
\hline
\textbf{bmifp1} & 0.504 (0.358, 0.709) & 0.50 (0.36, 0.71) \\
\hline
\textbf{bmifp2} & 1.156 (1.070, 1.249) & 1.16 (1.07, 1.25) \\
\hline
\hline
\textbf{AGE} & 1.061 (1.044, 1.079) & 1.06 (1.04, 1.08) \\
\hline
\textbf{hr10} & 1.129 (1.066, 1.197) & 1.13 (1.07, 1.20) \\
\hline
\hline
\textbf{diasbp10} & 0.885 (0.826, 0.948) & 0.88 (0.83, 0.95) \\
\hline
\hline
\textbf{GENDER} & 6.282 (0.961, 41.070) & 6.33 (0.97, 41.37) \\
\hline
\hline
\textbf{CHF} & 2.294 (1.720, 3.060) & 2.30 (1.72, 3.07) \\
\hline
\hline
\textbf{ga} & 0.973 (0.950, 0.996) & 0.97 (0.95, 1.00) \\
\hline
\end{tabular}
\caption{\label{Tab:ComparisonUCLA}Confirming results using the ground truth WHAS500 dataset.}
\end{table}

\begin{table}[h]
\small
\centering
\begin{tabular}{|p{5cm}|p{5cm}|}
\hline
\textbf{Variable} & \textbf{HR (95\% CI) (Ours)} \\
\hline
\hline
\textbf{Age} & 1.05 (1.04, 1.06) \\
\hline
\hline
\textbf{BMI} & 0.96 (0.93, 0.99) \\
\hline
\hline
\textbf{Sex} & \\
Male & Baseline \\
Female & 0.76 (0.57, 1.01) \\
\hline
\hline
\textbf{Heart Rate} & 1.01 (1.00, 1.02) \\
\hline
\textbf{Systolic Blood Pressure (SysBP)} & 1.00 (0.99, 1.01) \\
\hline
\textbf{Diastolic Blood Pressure (DiasBP)} & 0.99 (0.98, 1.00) \\
\hline
\hline
\textbf{History of\newline Cardiovascular Disease (CVD)} & \\
False & Baseline \\
True & 1.01 (0.71, 1.43) \\
\hline
\textbf{History of\newline Atrial Fibrillation (AF)} & \\
False & Baseline \\
True & 1.14 (0.81, 1.59) \\
\hline
\textbf{History of\newline Congestive Heart Failure (CHF)} & \\
False & Baseline \\
True & 2.17 (1.62, 2.91) \\
\hline
\hline
\textbf{Myocardial Infarction Order\newline (MI Order)} & \\
First & Baseline \\
Recurrent & 1.04 (0.78, 1.40) \\
\hline
\textbf{Myocardial Infarction Type\newline (MI Type)} & \\
non Q-wave & Baseline \\
Q-wave & 0.85 (0.59, 1.23) \\
\hline
\end{tabular}
\caption{\label{Tab:AppWHAS-2}The additional information here shows how the HR would look like in the ground truth WHAS500 dataset, if the CoxPH model were to be modelled using more variables as its inputs.}
\end{table}

\newpage
\subsection{FLChain}\label{App:MoreDetailsFLChain}
The FLChain dataset, introduced by Dispenzieri \textit{et al.}~\cite{dispenzieri2012use}, originates from a study assessing the prognostic value of nonclonal serum immunoglobulin free light chains (FLC) in predicting overall survival in the general population. It encompasses data on 15,859 residents of Olmsted County, Minnesota, aged 50 years or older, collected between March 13, 1995, and November 21, 2003, with follow-up through June 30, 2009. The dataset includes serum FLC measurements (kappa and lambda), demographic details (age, sex), baseline serum creatinine levels, and follow-up status, including cause of death. Participants were tested for kappa and lambda FLCs and serum creatinine levels to explore the relationship between elevated nonclonal FLC levels and overall survival, with events defined as death from any cause. The extensive follow-up period, with a median duration of 12.7 years, facilitates a detailed analysis of mortality predictors in an older adult population.

Our research uses a variation available from the \texttt{sksurv} package, treated as the ground truth dataset. To ensure the synthetic dataset's value for research, modelling outcomes must replicate results from the original paper. We follow Dispenzieri \textit{et al.}'s 4-input multivariate CoxPH setup, as shown in Table 2 in their paper; including the top decile (top 10\%) of total FLC, creatinine, age (in 10-year brackets), and sex (female as baseline). The dataset from the \texttt{sksurv} package has missing information; hence after selecting the mentioned variables, we employed MICE (multiple imputation through chained equations) to impute the missing data before modelling. The results are presented in Table \ref{Tab:AppFLChain-1}. Despite the original study not mentioning missing data, our imputed dataset reflects similar HR-related statistics to their study.

To increase the complexity of data synthesis and the flexibility of the synthetic dataset for downstream applications, we include additional variables available in the FLChain study not part of Dispenzieri \textit{et al.} survival data analysis. Fitting a CoxPH model over all variables in the dataset setup, using the ground truth data, we observe the HR-related statistics in Table \ref{Tab:AppFLChain-2}. Note that some information, such as the top decile of total FLC and the decile of FLC values, may be redundant. Practitioners can decide how to utilise the dataset for downstream applications.

\newpage
\begin{table}[h]
\small
\centering
\begin{tabular}{|p{3cm}|p{4cm}|p{4cm}|}
\hline
\textbf{Variable} & \textbf{HR (95\% CI)\newline (Dispenzieri \textit{et al.})} & \textbf{HR (95\% CI)\newline (Ours)} \\
\hline
\hline
\textbf{Total FLC} & & \\
Not top decile & Baseline & Baseline \\
Top decile & 2.07 (1.91, 2.24) & 2.14 (1.92, 2.39) \\
\hline
\hline
\textbf{Creatinine} & 1.26 (1.20, 1.31) & 1.24 (1.16, 1.32) \\
\hline
\hline
\textbf{Age (10 year brackets)} & 2.85 (2.76, 2.95) & 2.61 (2.50, 2.72) \\
\hline
\hline
\textbf{Sex} & & \\
Female & Baseline & Baseline \\
Male & 1.34 (1.25, 1.43) & 1.29 (1.18, 1.41) \\
\hline
\end{tabular}
\caption{\label{Tab:AppFLChain-1}Confirming results using the ground truth FLChain dataset.}
\end{table}

\begin{table}[h]
\small
\centering
\begin{tabular}{|p{5cm}|p{5cm}|}
\hline
\textbf{Variable} & \textbf{HR (95\% CI) (Ours, Ground Truth Dataset)} \\
\hline
\hline
\textbf{Age (10 year brackets)} & 2.55 (2.44, 2.67) \\
\hline
\hline
\textbf{Sex} & \\
Female & Baseline \\
Male & 1.30 (1.19, 1.42) \\
\hline
\hline
\textbf{Creatinine} & 1.05 (0.95, 1.16) \\
\hline
\hline
\textbf{Top FLC Decile} & \\
Not in Top Decile & Baseline \\
In Top Decile & 1.39 (1.21, 1.60) \\
\hline
\textbf{Decile of Total FLC} & 1.05 (1.03, 1.07) \\
\hline
\textbf{Serum FLC Kappa} & 1.03 (0.96, 1.10) \\
\hline
\textbf{Serum FLC Lambda} & 1.13 (1.07, 1.20) \\
\hline
\hline
\textbf{Diagnosed with MGUS} & \\
False & Baseline \\
True & 1.14 (0.69, 1.88) \\
\hline
\end{tabular}
\caption{\label{Tab:AppFLChain-2}The additional information here shows how the HR would look like in the ground truth FLChain dataset, if the CoxPH model were to be modelled using more variables as its inputs.}
\end{table}

\newpage
\section{More Details on the DCM Encoder Implementation}\label{App:TrainingDetailsOfEncoder}

The following section includes the architecture and specific hyperparameters utilised during the training of the DCM encoder. Note, this paper is published with sample codes provided.

\subsection{DCM Model Architecture}

\begin{table}[h!]
    \small
    \centering
    \begin{tabular}{lcll}
        \hline
        \textbf{Layer Name} & \textbf{Output Size} & \textbf{Layer Configuration} \\
        \hline
        \hline
        \textbf{input} & N $\times$ input\_dim & Input features (\textit{e.g.,} Size, Therapy, etc.) \\
        \textbf{hidden1} & N $\times$ 64 & [Linear(input\_dim, 64), ReLU, BatchNorm1d(64)]\\
        \textbf{hidden2} & N $\times$ 32 & [Linear(64, 32), ReLU, BatchNorm1d(32)]\\
        \textbf{hidden3} & N $\times$ 16 & [Linear(32, 16), ReLU, BatchNorm1d(16)]\\
        \textbf{output\_layer} & N $\times$ num\_mixtures & Linear(16, num\_mixtures)\\
        \textbf{mixture\_weights} & N $\times$ num\_mixtures & Linear(16, num\_mixtures)\\
        \textbf{risk\_score} & N $\times$ 1 & (output\_layer * mixture\_weights).sum(dim=1, keepdim=True)\\
        \hline
    \end{tabular}
    \caption{The DCM encoder model architecture.}
\end{table}

In this architecture, \textbf{N} represents the batch size used during model training and inference, while \textbf{input\_dim} denotes the number of input features provided to the model, such as clinical and demographic data. The network begins by processing the input features through a series of three fully connected hidden layers, each with decreasing dimensionality (64, 32, and 16 neurons, respectively). Each hidden layer is equipped with a linear transformation followed by a ReLU activation function, which introduces non-linearity, crucial for capturing complex interactions between the input features. To further enhance the stability and convergence of the training process, batch normalisation is applied after the ReLU activation in each hidden layer.

The \textbf{output\_layer} is a linear layer that projects the final hidden state into a space defined by \textbf{num\_mixtures}, which corresponds to the number of mixture components or latent survival patterns the model is designed to capture. These logits are then combined with the mixture weights, which are generated by another linear layer followed by a softmax function. The softmax operation ensures that the mixture weights are non-negative and sum to one, effectively interpreting them as probabilities. The final risk score is computed as a weighted sum of these logits, representing the aggregated risk across the mixture components. This architecture is tailored to model the heterogeneous survival outcomes within a patient cohort by leveraging the mixture of experts approach, allowing for the identification and interpretation of distinct survival trajectories within the data.

\subsection{Hyperparameters for the DCM Encoder}

\begin{table}[h!]
    \small
    \centering
    \begin{tabular}{p{3cm}p{4cm}p{5cm}}
        \hline
        \textbf{Category} & \textbf{Hyperparameter} & \textbf{Value/Description} \\
        \hline
        \hline
        \textbf{Model Architecture\newline Hyperparameters} & 
        \texttt{input\_dim}\newline \texttt{hidden\_layers} & 
        Depends on \texttt{X.shape[1]}\newline
        [64, 32, 16]\\
        & \texttt{num\_mixtures} & 5 \\
        \hline
        \hline
        \textbf{Training\newline Hyperparameters} &
        \texttt{criterion}\newline \texttt{optimizer}& 
        \texttt{nn.MSELoss()}\newline \texttt{optim.Adam(model.parameters(), \hspace*{18mm}lr=0.001)}\\
        & \texttt{num\_epochs} & 10,000\\
        & \texttt{penalizer} & 0.0001 ridge regularisation for CoxPH \\
        \hline
        \hline
        \textbf{Batch Processing\newline Hyperparameters} & \texttt{batch size} & 
        Since the datasets are relatively small, the entire dataset is processed at once in the training loop, hence no specific batch size. \\
        \hline
    \end{tabular}
    \caption{Hyperparameters for the DCM encoder.}
\end{table}

We provide a summary of the hyperparameters employed in the development and training of the DCM encoder. Each hyperparameter is categorised to distinguish between those affecting model architecture, training dynamics, and batch processing.

\newpage
\section{More Details on the SynthNet Decoder Implementation}\label{App:TrainingDetailsOfDecoder}

This section provides a detailed overview of the architecture and hyperparameters used in the training of the SynthNet decoder. Note, this paper is published with sample codes provided.

\subsection{SynthNet Model Architecture}

\begin{table}[h!]
    \small
    \centering
    \begin{tabular}{lcll}
        \hline
        \textbf{Layer Name} & \textbf{Output Size} & \textbf{Layer Configuration} \\
        \hline
        \hline
        \textbf{input} & N $\times$ hidden\_dim & Latent features from the DCM encoder \\
        \textbf{synth\_layer1} & N $\times$ 128 & [Linear(hidden\_dim, 128), ReLU] \\
        \textbf{synth\_layer2} & N $\times$ 128 & [Linear(128, 128), ReLU] \\
        \textbf{output\_layer} & N $\times$ input\_dim & [Linear(128, input\_dim), Sigmoid] \\
        \hline
    \end{tabular}
    \caption{The SynthNet decoder model architecture.}
\end{table}

In this architecture, \textbf{N} denotes the batch size used during model training and inference. The \textbf{hidden\_dim} represents the size of the latent feature vector provided by the DCM encoder, while \textbf{input\_dim} corresponds to the number of original features in the dataset (again, not including Event and Durtation). The model begins by processing the latent features through two fully connected hidden layers, each with 128 neurons, where ReLU activation functions introduce non-linearity to capture complex patterns. The final output layer projects the data back to the original input dimension, with a Sigmoid activation ensuring that the reconstructed data remains within the [0, 1] range.

\subsection{Hyperparameters for the SynthNet Decoder}

\begin{table}[h!]
    \small
    \centering
    \begin{tabular}{p{3cm}p{4cm}p{5cm}}
        \hline
        \textbf{Category} & \textbf{Hyperparameter} & \textbf{Value/Description} \\
        \hline
        \hline
        \textbf{Model Architecture\newline Hyperparameters} & 
        \texttt{input\_dim}\newline \texttt{hidden\_layers} & 
        Depends on \texttt{X.shape[1]}\newline
        [128, 128]\\
        \hline
        \hline
        \textbf{Training\newline Hyperparameters} &
        \texttt{criterion}\newline \texttt{optimizer}& 
        \texttt{nn.MSELoss()}\newline \texttt{optim.Adam(model.parameters(),\newline \hspace*{18mm}lr=0.001)}\\
        & \texttt{num\_epochs} & 20,000\\
        \hline
        \hline
        \textbf{Batch Processing\newline Hyperparameters} & \texttt{batch size} & 
        Since the datasets are relatively small, the entire dataset is processed at once during training, hence no specific batch size. \\
        \hline
    \end{tabular}
    \caption{Hyperparameters for the SynthNet decoder.}
\end{table}

This table summarises the hyperparameters used in developing and training the SynthNet decoder. The hyperparameters are categorised based on their impact on the model architecture, training process, and batch processing dynamics.

\newpage
\section{Preprocessing and Postprocessing: Using WHAS500 as an Example}\label{App:Processing}

This section details the preprocessing and postprocessing procedures found in our CK4Gen framework using WHAS500 as an example. Specifically, we focus on the rationale highlighting the distinct approaches required for the SynthNet decoder as compared to the DCM encoder.

\subsection{Preprocessing for SynthNet Decoder $\mathbf{x} \xrightarrow{\text{preprocess}} \mathbf{\tau}$}

The SynthNet decoder requires specific preprocessing steps to transform the WHAS500 dataset’s variables into a suitable format for training. Unlike the DCM encoder, which directly uses raw input features identical to those of a traditional CoxPH model, the SynthNet decoder involves additional transformations to align the data with the neural network's operational requirements.

\colorbox{cyan!25}{\{}Numerical variables such as Age, BMI, Heart Rate, Systolic Blood Pressure, and Diastolic Blood Pressure are first subjected to a Box-Cox transformation to stabilise variance and approximate a normal distribution. Post-transformation, these variables are standardised to ensure they have a mean of zero and a standard deviation of one. This standardisation is crucial for ensuring that the model’s parameters are not disproportionately influenced by variables on different scales.

Binary variables, including Sex, CVD, AF, CHF, MI Order, and MI Type, are mapped to [0, 1]. This normalisation is necessary because the SynthNet decoder outputs continuous values, and this step ensures these outputs can later be converted back into discrete binary values during postprocessing.\colorbox{cyan!25}{\}}

Following the descriptions given in the past two paragraphs (\textit{i.e.,} the material described in \colorbox{cyan!25}{\{} \colorbox{cyan!25}{\}}), all reconstruction targets now have values lying in [0, 1]; making it much simpler to be reconstructed using the \texttt{Softmax} function in SynthNet.

\begin{table}[h!]
    \small
    \centering
    \begin{tabular}{p{3cm}p{4cm}p{5cm}}
        \hline
        \textbf{Variable Type} & \textbf{Transformation Applied} & \textbf{Example Variables} \\
        \hline
        \hline
        \textbf{Numeric Variables} & Box-Cox Transformation,\newline Standardisation & Age, BMI, Heart Rate, SysBP, DiasBP \\
        \textbf{Binary Variables} & Map to [0, 1] & Sex, CVD, AF, CHF, MI Order, MI Type \\
        \textbf{Duration/Event} & Nothing required,\newline not even included. & \\
        \hline
    \end{tabular}
    \caption{Preprocessing steps for SynthNet Decoder applied to the WHAS500 dataset.}
\end{table}

\subsection{Postprocessing of SynthNet Outputs $\mathbf{\hat{\tau}}_{\text{synth}} \xrightarrow{\text{postprocess}} \mathbf{\hat{x}}_{\text{synth}}$}

Following the generation of synthetic data by the SynthNet decoder, postprocessing is necessary to revert the transformed variables to their original scale and format. The numerical variables undergo rescaling and inverse Box-Cox transformations to match the original data’s distribution and range. Binary variables are thresholded at 0.5 to convert the continuous outputs from the SynthNet decoder back into discrete binary values.

The Duration and Event columns, which are crucial for survival analysis, are not synthesised by the model. Instead, these columns are copied directly from the original dataset into the synthetic dataset. This method is feasible because the SynthNet decoder reconstructs each patient’s data in a bijective (1 real data-1 synthetic data) manner, ensuring that the temporal and event data remain consistent with the original dataset. This preservation is essential for maintaining the integrity of survival analysis.

\begin{table}[h!]
    \small
    \centering
    \begin{tabular}{p{3cm}p{4cm}p{5cm}}
        \hline
        \textbf{Variable Type} & \textbf{Postprocessing Steps} & \textbf{Example Variables} \\
        \hline
        \hline
        \textbf{Numeric Variables} & Rescaling,\newline
        Inverse Box-Cox & Age, BMI, Heart Rate, SysBP, DiasBP \\
        \textbf{Binary Variables} & Thresholding at 0.5 & Sex, CVD, AF, CHF, MI Order, MI Type \\
        \textbf{Duration/Event} & Copy directly from original data & Duration, Event \\
        \hline
    \end{tabular}
    \caption{Postprocessing steps for SynthNet Decoder applied to the WHAS500 dataset.}
\end{table}

\newpage
\section{CK4Gen Framework and Novel Data Generation}\label{App:NovelCK4Gen}

\begin{table}[h]
\centering
\begin{tabular}{|p{3cm}|p{3cm}|p{3cm}|p{3cm}|}
\hline
\textbf{Aspect} & \textbf{CK4Gen's Encoding-Decoding} & \textbf{CK4Gen's\newline Novel Data\newline Generation} & \textbf{Traditional\newline VAE\newline Encoding-Decoding} \\ 
\hline
\hline
\textbf{Purpose} & 
Accurate\newline reconstruction. & 
Generating novel data\newline by perturbing latent\newline representations. & Generating novel data\newline through latent space\newline sampling. \\ 
\hline
\textbf{Procedure} & 
\(\mathbf{h} = F(\mathbf{\tau}; \theta_F)\) \newline 
\(\mathbf{\hat{\tau}}_{\text{synth}} = G(\mathbf{h}; \theta_G)\) & 
\(\mathbf{h}' = \mathbf{h} + \delta\) \newline 
\(\mathbf{\hat{\tau}}_{\text{synth}}' = G(\mathbf{h}'; \theta_G)\) & 
\(\mathbf{z} \sim \mathcal{N}(\mu(\mathbf{x}), \sigma^2(\mathbf{x}))\) \newline 
\(\mathbf{\hat{\tau}}''_{\text{synth}} = G(\mathbf{z})\) \\ \hline
\textbf{Novelty of\newline Synthetic Data} & Limited novelty. & Increased novelty. & High novelty. \\ \hline
\textbf{Handling of\newline Event and Duration} & Directly copied. & Imputed using MICE or similar methods. & Part of the generation procedure. \\ \hline
\textbf{Risk of\newline Unrealistic Data} & Low risk. & Moderate risk. & High risk. \\ \hline
\textbf{Utility for\newline Downstream Tasks} & High utility. & Variable utility. & Variable utility. \\ \hline
\end{tabular}
\caption{A comparison of CK4Gen to VAE.}
\label{tab:CK4Gen_VAE_Comparison}
\end{table}

This appendix provides a detailed exploration of the CK4Gen framework, including the potential for generating novel data through (complete) latent space perturbations and a comparison with traditional VAE approaches.

\subsection{CK4Gen Data Generation Process: Mathematical Formulation}

\paragraph{Preprocessing}
The original patient data \( \mathbf{x} \) undergoes preprocessing, transforming it into standardised data \( \mathbf{\tau} \) via:
\begin{align}
\mathbf{\tau} = P(\mathbf{x})
\end{align}
where \( P \) includes Box-Cox transformations, standardisation, and binary mapping.

\paragraph{DCM Encoder \( F \)}
The preprocessed data \( \mathbf{\tau} \) is then encoded by the DCM encoder \( F \), producing a latent representation \( \mathbf{h} \) that encapsulates essential survival-related features:
\begin{align}
\mathbf{h} = F(\mathbf{\tau}; \theta_F)
\end{align}
where \( \theta_F \) denotes the parameters of the encoder.

\paragraph{SynthNet Decoder \( G \)}
The latent representation \( \mathbf{h} \) is passed to the SynthNet decoder \( G \), which reconstructs the preprocessed data:
\begin{align}
\mathbf{\hat{\tau}}_{\text{synth}} = G(\mathbf{h}; \theta_G).
\end{align}
The information bottleneck in \( G \) results in \( \mathbf{\hat{\tau}}_{\text{synth}} \) being a novel yet statistically consistent representation, as it cannot capture every detail of \( \mathbf{\tau} \).

\paragraph{Postprocessing}
The reconstructed data \( \mathbf{\hat{\tau}}_{\text{synth}} \) is postprocessed to match the original data format:
\begin{align}
\mathbf{\hat{x}}_{\text{synth}} = P^{-1}(\mathbf{\hat{\tau}}_{\text{synth}})
\end{align}
ensuring it aligns with the original dataset's scale and structure.

\paragraph{Event and Duration Incorporation}
To maintain the integrity of survival outcomes, Event \( \mathbf{E} \) and Duration \( \mathbf{T} \) are directly copied from the original dataset:
\begin{align}
\mathbf{\hat{x}}_{\text{synth}} = [\mathbf{\hat{x}}_{\text{synth}}, \mathbf{E}, \mathbf{T}].
\end{align}

\paragraph{Overall Process}
The full CK4Gen process is summarised as:
\begin{align}
\mathbf{\hat{x}}_{\text{synth}} = P^{-1}\left(G\left(F\left(P(\mathbf{x}); \theta_F\right); \theta_G\right)\right),
\end{align}
with the final synthetic dataset expressed as:
\begin{align}
\mathbf{\hat{x}}_{\text{synth}} = \left[P^{-1}\left(G\left(F\left(P(\mathbf{x}); \theta_F\right); \theta_G\right)\right), \mathbf{E}, \mathbf{T}\right].
\end{align}

\newpage
\subsection{Impact of Latent Space Perturbation in the SynthNet Decoder \( G \)}

If the SynthNet decoder \( G \) receives a perturbed latent representation \( \mathbf{h}' = \mathbf{h} + \delta \) instead of the original \( \mathbf{h} \), the resulting synthetic data \( \mathbf{\hat{\tau}}_{\text{synth}}' \) may deviate from the expected output. The reconstruction is given by:
\begin{align}
\mathbf{\hat{\tau}}_{\text{synth}}' = G(\mathbf{h}'; \theta_G) = G(\mathbf{h} + \delta; \theta_G).
\end{align}

This perturbation introduces variation, hence leading to novel data points. The difference from the original reconstruction can be expressed as:
\begin{align}
\Delta \mathbf{\hat{\tau}}_{\text{synth}} = G(\mathbf{h} + \delta; \theta_G) - G(\mathbf{h}; \theta_G).
\end{align}

Such perturbations can create diversity but also risk producing unrealistic data if \( \mathbf{h}' \) lies outside the learned distribution. 

In contrast, VAE samples latent variables \( \mathbf{z} \) from a Gaussian distribution:
\begin{align}
q(\mathbf{z}|\mathbf{x}) = \mathcal{N}(\mu(\mathbf{x}), \sigma^2(\mathbf{x})).
\end{align}
The reconstruction is then:
\begin{align}
\mathbf{\hat{x}} = G(\mathbf{z}).
\end{align}
While this allows for the generation of diverse data, it can blend distinct patient profiles, risking misrepresentation of survival outcomes.

For perturbed data, directly copying Event \( \mathbf{E} \) and Duration \( \mathbf{T} \) is not appropriate. Instead, multiple imputation by chained equations (MICE) can be used to infer these values:
\begin{align}
(\mathbf{\hat{E}}', \mathbf{\hat{T}}') = \text{MICE}(\mathbf{\hat{\tau}}_{\text{synth}}').
\end{align}
This approach ensures that the synthetic dataset maintains internal coherence and utility for survival analysis.

\newpage
\section{Descriptions on Data Augmentation Techniques}\label{App:AugRelated}

This appendix details the data augmentation techniques used in our study, encompassing traditional oversampling and undersampling methods alongside advanced generative ML models. Except for the generative ML models, all techniques are implemented using the imblearn library~\cite{JMLR:v18:16-365}.

\subsection{Cross-Validation and Synthetic Data Augmentation}
\begin{algorithm}[h]
\caption{Cross-Validation and Risk Prediction with Data Augmentation Using CK4Gen}
\label{alg:cross_val_ck4gen}
\begin{algorithmic}[1]
\Require Dataset with predictors \( X \), time \( T \), and event status \( E \)
\State Partition the dataset using 5-fold cross-validation with 2 swaps per fold

\Procedure{Data Partitioning}{}
    \For{each fold \(i = 1, \dots, 5\)}
        \State Stratify the dataset into training and testing sets
        \State Assign fold labels: \( FOLD_i = 0 \) for training, \( FOLD_i = 1 \) for testing
    \EndFor
\EndProcedure

\Procedure{Cross-Validation}{}
    \For{each fold and swap}
        \State \textbf{Train CoxPH Model}:
        \State Extract training data from the current fold
        \If{include synthetic data}
            \State Append synthetic data to the training data
        \EndIf
        \State \textbf{Generate Risk Predictions}:
        \State Extract the test data corresponding to the current fold (no synthetic data used)
        \State Compute the linear predictor (LPH) for the test data
    \EndFor
\EndProcedure
\end{algorithmic}
\end{algorithm}

We employed 5x2 cross-validation~\cite{dietterich1998approximate} to assess the impact of CK4Gen's synthetic data on the performance of CoxPH models across various datasets. The dataset was initially divided into five folds, each of which was split into two sets -- one for training and one for testing. For each fold, the roles of the training and testing sets were swapped in the second iteration. Stratified sampling was employed during partitioning to ensure that the distribution of event occurrences remained consistent across both training and testing sets.

Depending on the experimental condition, the CoxPH model was trained either on the original data alone or augmented with synthetic data generated by CK4Gen. Importantly, the synthetic data was used solely during the training phase, not during testing, to evaluate its effect on the model's ability to generalise. Following model training, linear predictor values (\textit{i.e.,} log partial hazards~\cite{Davidson-Pilon2019}) were computed for the testing set, enabling an assessment of the model’s performance.

\subsection{Synthetic Data Augmentation using Oversampling Techniques and\newline Alternative Generative ML Models}
Whether utilising CK4Gen, traditional oversampling techniques, or advanced generative models like VAEs or GANs, the implementation would largely remain consistent. The distinction lies in the generation and integration of synthetic data, which occurs within the same experimental structure.

\subsubsection{Synthetic Minority Over-sampling Technique (SMOTE)}

SMOTE~\cite{chawla2002smote} is a technique designed to address class imbalance through enhancing the representation of the minority class. The technique works by generating synthetic samples along the lines connecting existing minority class instances. 

This process involves two key steps: identifying \( k \)-nearest neighbours for each minority class instance and then generating synthetic samples. For each minority instance \( x_i \), SMOTE randomly selects one of its \( k \)-nearest neighbours \( x_j \), and creates a synthetic instance \( x_{\text{new}} \) through linear interpolation:
\begin{equation}
x_{\text{new}} = x_i + \lambda \times (x_j - x_i)
\end{equation}
where \( \lambda \) is a random scalar drawn from a uniform distribution between 0 and 1. This interpolation introduces variability by creating synthetic samples that fall within the convex hull of the minority class distribution, rather than replicating existing data.

However, SMOTE’s reliance on linear interpolation can result in synthetic instances that do not accurately reflect the true distribution of the minority class, particularly in feature spaces with overlapping regions or uneven distributions. In survival analysis, the situation is further complicated by the time-to-event component. Interpolating not only in feature space but also in time-to-event can generate synthetic patients with non-existent risk profiles, as the process might blend high-risk and low-risk individuals in ways that are unrealistic. These synthetic instances, though mathematically valid, do not correspond to plausible clinical scenarios and can distort the model’s understanding of risk stratification, leading to degraded performance metrics such as Harrell's C-index.

\subsubsection{ADASYN and SVMSMOTE}
ADASYN~\cite{he2008adasyn} and SVMSMOTE~\cite{nguyen2011borderline} build on SMOTE by introducing more sophisticated techniques for identifying minority instances to generate synthetic samples. ADASYN focuses on harder-to-classify instances, placing greater emphasis on minority samples near the decision boundary or surrounded by majority class instances, dynamically adjusting the number of synthetic samples generated based on the instance's classification difficulty. Similarly, SVMSMOTE uses support vector machines (SVMs)~\cite{cortes1995support} to prioritise minority instances close to the decision boundary for generating synthetic data, aiming to improve decision-making in challenging regions.

Despite these advanced instance-selection mechanisms, both ADASYN and SVMSMOTE retain SMOTE’s linear interpolation strategy. This means they replicate the same clinical pitfalls seen with SMOTE, especially in survival analysis, where interpolating between high-risk and low-risk patients produces unrealistic synthetic profiles that can distort the model's predictions.

\subsubsection{Variational Autoencoder}
Both CK4Gen and VAE~\cite{kingma2014auto} share similarities as deep learning frameworks for synthetic data generation. Each employs an encoder-decoder architecture with ReLU-activated multi-layer subnetworks to capture data complexity, relying on pre-processing transformations for stability, and inverse transformations during post-processing.

The key differences between these models stem from their design and focus. CK4Gen is tailored for survival analysis, utilising a DCM encoder to capture latent survival patterns and generate clinically interpretable risk scores. VAE, by contrast, employs a general-purpose latent space, making it more flexible but less specialised for clinical contexts. Additionally, CK4Gen directly retains the outcome variables, Event and Duration, from the original dataset during post-processing, whereas VAE treats all features, including outcomes, uniformly.

\underline{VAE Model Architecture}

\begin{table}[h!]
    \small
    \centering
    \begin{tabular}{lcll}
        \hline
        \textbf{Layer Name} & \textbf{Output Size} & \textbf{Layer Configuration} \\
        \hline
        \hline
        \textbf{input} & N $\times$ input\_dim & Input features (e.g., Age, BMI, etc.) \\
        \textbf{hidden1} & N $\times$ 128 & [Linear(input\_dim, 128), ReLU] \\
        \textbf{hidden2} & N $\times$ 64 & [Linear(128, 64), ReLU] \\
        \textbf{mu} & N $\times$ latent\_dim & Linear(64, latent\_dim) \\
        \textbf{logvar} & N $\times$ latent\_dim & Linear(64, latent\_dim) \\
        \textbf{latent\_space} & N $\times$ latent\_dim & Reparameterisation using $\mu$ and \textbf{logvar} \\
        \textbf{decode1} & N $\times$ 64 & [Linear(latent\_dim, 64), ReLU] \\
        \textbf{decode2} & N $\times$ 128 & [Linear(64, 128), ReLU] \\
        \textbf{output\_layer} & N $\times$ input\_dim & [Linear(128, input\_dim), Sigmoid] \\
        \hline
    \end{tabular}
    \caption{The VAE model architecture.}
\end{table}
In this architecture, \textbf{N} denotes the batch size during training, while \textbf{input\_dim} represents the number of features from the original dataset, including outcome variables. The VAE processes the input through two fully connected hidden layers of 128 and 64 neurons, each activated by ReLU.

The model then branches into two linear layers, $\mu$ and \textbf{logvar}, which define the mean and variance of the latent space. These components are combined through reparameterisation, enabling the model to sample from a Gaussian distribution and generate synthetic data.

The decoder mirrors the encoder, with two fully connected layers and a final output layer that reconstructs the latent variables into the original feature space. A Sigmoid activation ensures the output remains within the [0, 1] range, preserving the scale of the original data.

\underline{Hyperparameters for the VAE}

\begin{table}[h!]
    \small
    \centering
    \begin{tabular}{p{3cm}p{3cm}p{6cm}}
        \hline
        \textbf{Category} & \textbf{Hyperparameter} & \textbf{Value/Description} \\
        \hline
        \hline
        \textbf{Model Architecture\newline Hyperparameters} & 
        \texttt{input\_dim}\newline \texttt{hidden\_layers} & 
        Depends on \texttt{X.shape[1]}\newline
        [128, 64] \\
        & \texttt{latent\_dim} & 32 \\
        \hline
        \hline
        \textbf{Training\newline Hyperparameters} &
        \texttt{criterion}\newline \texttt{optimizer} & 
        \texttt{nn.MSELoss()}\newline \texttt{optim.Adam(model.parameters(), \hspace*{18mm}lr=0.001)} \\
        & \texttt{num\_epochs} & 1,000 \\
        & \texttt{batch size} & 32 \\
        \hline
        \hline
        \textbf{Loss Function\newline Components} & \texttt{loss\_function} &
        A combination of three components:
        \begin{equation*}
            \text{MSE} + 0.001 \times \text{KLD} + \text{Correlation Loss}
        \end{equation*}
        \textbf{Reconstruction Loss (MSE)}:\newline Calculated using \texttt{nn.functional.mse\_loss} between the original and reconstructed output.
        
        \textbf{KL Divergence (KLD)}:\newline A regularisation term to ensure that the learned latent space distribution approximates a Gaussian distribution. The KLD term is scaled by a factor of 0.001.
        
        \textbf{Correlation Loss}:\newline
        An optional regularisation term that better preserves feature relationships.
        
        \\
        \hline
        \hline
        \textbf{Post-Processing\newline Hyperparameters} & \texttt{inverse Box-Cox} & 
        Applied to output to return synthetic data to original scale \\
        \hline
    \end{tabular}
    \caption{Hyperparameters for the VAE.}
\end{table}

This table summarises the hyperparameters and loss function components used in the training of the VAE model. The architecture parameters determine the structure of the model, while the loss function balances reconstruction accuracy, latent space regularisation, and preservation of feature relationships.

\subsubsection{Generative Adversarial Networks}
In a similar fashion to CK4Gen and VAE, GANs~\cite{goodfellow2014generative} leverage deep learning architectures for synthetic data generation. They use multi-layer networks with ReLU activations, and like the other models, employ pre-processing transformations for stability, followed by inverse transformations during post-processing. GANs distinguish themselves through their adversarial framework, where a generator creates synthetic data and a critic (or discriminator) evaluates its realism. This dynamic competition forces the generator to iteratively improve, leading to highly realistic data generation.

\underline{GAN Model Architecture}

In this architecture, \textbf{N} refers to the batch size, while \textbf{latent\_dim} denotes the dimensionality of the latent space used to generate data. The generator consists of two fully connected hidden layers of 64 and 128 neurons, with ReLU activations. The output layer, activated by a Sigmoid function, projects the data back into the original feature space, ensuring the synthetic data remains within [0, 1].

The critic mirrors the generator in structure but operates in reverse, processing real or synthetic data to evaluate its authenticity. The final output is a single scalar value indicating the likelihood that the input data is real.

\newpage
\begin{table}[h!]
    \small
    \centering
    \begin{tabular}{lcll}
        \hline
        \textbf{Layer Name} & \textbf{Output Size} & \textbf{Layer Configuration} \\
        \hline
        \hline
        \textbf{input} & N $\times$ latent\_dim & Random latent vector (e.g., sampled from normal distribution) \\
        \textbf{hidden1 (Generator)} & N $\times$ 64 & [Linear(latent\_dim, 64), ReLU] \\
        \textbf{hidden2 (Generator)} & N $\times$ 128 & [Linear(64, 128), ReLU] \\
        \textbf{output\_layer (Generator)} & N $\times$ input\_dim & [Linear(128, input\_dim), Sigmoid] \\
        \textbf{hidden1 (Critic)} & N $\times$ 128 & [Linear(input\_dim, 128), ReLU] \\
        \textbf{hidden2 (Critic)} & N $\times$ 64 & [Linear(128, 64), ReLU] \\
        \textbf{output\_layer (Critic)} & N $\times$ 1 & [Linear(64, 1)] \\
        \hline
    \end{tabular}
    \caption{The GAN model architecture.}
\end{table}

\underline{Hyperparameters for the GAN}

\begin{table}[h!]
    \small
    \centering
    \begin{tabular}{p{3cm}p{3cm}p{6cm}}
        \hline
        \textbf{Category} & \textbf{Hyperparameter} & \textbf{Value/Description} \\
        \hline
        \hline
        \textbf{Model Architecture\newline Hyperparameters} & 
        \texttt{latent\_dim}\newline \texttt{hidden\_layers} & 
        32 (latent space)\newline
        [128, 64] (for both generator and critic) \\
        \hline
        \hline
        \textbf{Training\newline Hyperparameters} &
        \texttt{optimizer\_G}\newline
        \textcolor{white}{.}\newline\texttt{optimizer\_C} & 
        \texttt{optim.Adam(generator.parameters(), \hspace*{18mm}lr=1e-4)}\newline \texttt{optim.Adam(critic.parameters(), \hspace*{18mm}lr=1e-4)} \\
        & \texttt{num\_epochs} & 500 \\
        & \texttt{n\_critic} & 5 (critic iterations per generator iteration) \\
        & \texttt{batch size} & 32 \\
        & \texttt{clip\_value} & 0.01 (for weight clipping) \\
        \hline
        \hline
        \textbf{Loss Function\newline Components} & \texttt{loss\_function} &
        \textbf{Critic Loss}:
        \begin{equation*}
            \text{Mean}(fake\_score) - \text{Mean}(real\_score)
        \end{equation*}
        
        \textbf{Generator Loss}:
        \begin{equation*}
            -\text{Mean}(fake\_score)
        \end{equation*}

        \textbf{Correlation Loss}:\newline
        An optional regularisation term that better preserves feature relationships.
        \\
        \hline
        \hline
        \textbf{Post-Processing\newline Hyperparameters} & \texttt{inverse Box-Cox} & 
        Applied to output to return synthetic data to original scales \\
        \hline
    \end{tabular}
    \caption{Hyperparameters for the GAN.}
\end{table}

This table summarises the hyperparameters used in training the GAN. The loss functions guide the adversarial training, while the correlation loss ensures the generated data maintains the relationships present in the original features.

\newpage
\subsubsection{Preprocessing for CK4Gen, VAE, and GANs: A Focus on Differences}

When applying synthetic data generation techniques to a mixed-type dataset like \textbf{ACTG320}, which contains numeric and binary variables, different preprocessing steps are needed depending on the model being used. This section focuses on the differences in preprocessing between \textbf{CK4Gen} and the combined approach for \textbf{VAE} and \textbf{GANs}.

\underline{CK4Gen}\\
CK4Gen (or more specifically the DCM-encoder in the CK4Gen) relies on \textbf{CoxPH} as a teacher model for \textit{knowledge distillation} during its training process. As a result, it processes input data in the same structure as CoxPH, avoiding the more extensive transformations required by VAE and GANs.

\begin{enumerate}
    \item \textbf{No Scaling of Numeric Variables}:\\
    CK4Gen does not require scaling or transformation of numeric data. It directly processes the original numeric variables, such as \textbf{Age}, \textbf{Functional Impairment}, and \textbf{Duration}, preserving their raw values to ensure consistency with the CoxPH input structure.

    \item \textbf{No One-Hot Encoding for Binary Variables}:\\
    In contrast to VAEs and GANs, CK4Gen does not apply one-hot encoding to binary variables like \textbf{Sex} and \textbf{CD4 Cell Count}. These binary variables are input in their original format (e.g., \textit{Female = 0}, \textit{Male = 1}), as the model inherits CoxPH’s ability to handle such variables without transformation.

    \item \textbf{Direct Processing of Mixed-Type Data}:\\
    CK4Gen handles both numeric and binary variables directly in their native format, eliminating the need for concatenation or complex transformations. This is crucial for retaining the clinical relevance of the dataset and ensuring the survival analysis framework remains intact.
\end{enumerate}

\underline{VAEs and GANs}\\
The preprocessing for VAEs and GANs is designed to prepare the mixed-type input data in a format suitable for their deep learning frameworks~\cite{kuo2022health}. Although the two models differ in architecture, they share similar preprocessing steps.

\begin{enumerate}
    \item \textbf{Scaling of Numeric Variables}:\\
    Numeric variables such as \textbf{Age}, \textbf{Functional Impairment}, \textbf{Months of Prior ZDV Use}, and \textbf{Duration} are scaled to the \textbf{[0, 1]} range using methods like \textit{min-max scaling} or \textit{Box-Cox transformation}. This stabilises model training and ensures consistency across features of different scales.

    \item \textbf{One-Hot Encoding of Binary Variables}:\\
    Binary variables like \textbf{Sex}, \textbf{CD4 Cell Count}, and \textbf{Event} are transformed into \textit{one-hot encoded vectors}. For example, \textbf{Sex} is represented as \textit{[1, 0]} for \textit{Female} and \textit{[0, 1]} for \textit{Male}. This standardises the binary data format, allowing the neural networks to process them efficiently.

    \item \textbf{Concatenation of Transformed Data}:\\
    After scaling numeric data and one-hot encoding binary variables, the transformed features are concatenated into a single input vector. This ensures all variables are represented in a unified format, allowing the VAE and GAN models to handle the entire dataset as a single input for training.
\end{enumerate}

In conclusion, \textbf{CK4Gen}’s reliance on CoxPH as a teacher model eliminates the need for complex transformations, while \textbf{VAEs} and \textbf{GANs} require additional preprocessing steps such as scaling, one-hot encoding, and concatenation to standardise the dataset for deep learning frameworks.

\newpage
\subsection{Cross-Validation and Risk Prediction with Undersampling Techniques}

\subsubsection{Experimental Setup}
\begin{algorithm}[h]
\caption{Cross-Validation and Risk Prediction with Undersampling Techniques}
\label{alg:cross_val_undersampling}
\begin{algorithmic}[1]
\Require Dataset with predictors \( X \), time \( T \), and event status \( E \)
\State Partition the dataset using repeated cross-validation with \( N \) folds and multiple repeats.

\Procedure{Data Partitioning}{}
    \For{each fold \( i = 1, \dots, N \)}
        \State Stratify the dataset into training and testing sets
        \State Assign fold labels: \( FOLD_i = 0 \) for training, \( FOLD_i = 1 \) for testing
    \EndFor
\EndProcedure

\Procedure{Cross-Validation with Undersampling}{}
    \For{each fold and repeat}
        \State \textbf{Apply Undersampling Technique}:
        \State Extract training data from the current fold
        \State Balance the training data using an undersampling technique
        \State \textbf{Train CoxPH Model}:
        \State Fit a Cox Proportional Hazards model on the undersampled training data
        \State \textbf{Generate Risk Predictions}:
        \State Extract the test data corresponding to the current fold
        \State Predict risk scores for the test data using the trained CoxPH model
    \EndFor
\EndProcedure
\end{algorithmic}
\end{algorithm}

Undersampling techniques differ from oversampling methods primarily in the strategy employed to address class imbalance. While oversampling introduces synthetic data to enhance minority class representation, undersampling reduces the majority class size to achieve balance. This approach is crucial for maintaining the integrity of real data during the training process, unlike oversampling, which can introduce artificial instances. 

One feature of undersampling is the exclusive reliance on real data, as the model is not exposed to potentially inaccurate synthetic examples. Conversely, oversampling aims to improve generalisation by augmenting the minority class, although it risks overrepresenting synthetic data that might not capture the true variability of the minority class. 

\subsubsection{Undersampling Techniques}
\underline{Random Undersampling}

Random undersampling addresses class imbalance by reducing the representation of the majority class in the training data. This is accomplished by randomly selecting and removing instances from the majority class until its size matches that of the minority class. While it is commonly used in many fields, it introduces the risk of information loss. By discarding potentially informative instances from the majority class, the model may become less capable of generalising to unseen data, particularly in complex feature spaces.

In survival analysis, removing instances from the majority class may also reduce the variability in time-to-event data, which can distort the model’s understanding of risk stratification. For example, in scenarios where the majority class includes patients with a broad range of survival times, undersampling could lead to a dataset that lacks sufficient representation of long-term survivors or high-risk individuals, potentially leading to degraded performance in terms of metrics like Harrell’s C-index.

\underline{Tomek Links and Neighbourhood Cleaning Rules}

Tomek links~\cite{tomek1976two} and Neighbourhood Cleaning Rules (NCR)~\cite{laurikkala2001improving} build upon the principles of random undersampling by introducing more sophisticated mechanisms to identify and remove problematic instances, thus reducing class overlap and enhancing the quality of the training data.

Tomek links are pairs of instances from different classes that are the nearest neighbours to each other, where the presence of such a pair indicates potential class overlap. If two instances from opposite classes are extremely close, one of them likely represents noise or an ambiguous region that contributes to class confusion. In practice, the majority class instance in a Tomek link is removed to increase the separability between classes.

NCR take this concept a step further by considering not only pairwise relationships between instances but also the local neighbourhood around each instance. NCR identifies noisy or borderline instances from the majority class that are likely to lead to misclassification. These are typically instances surrounded by minority class examples, where the majority class instance lies close to the decision boundary.

\subsection{Cross-Validation and Risk Prediction with Bootstrap Ensemble}

\begin{algorithm}[h]
\caption{Cross-Validation and Risk Prediction with Bootstrap Ensemble}
\label{alg:cross_val_bootstrap}
\begin{algorithmic}[1]
\Require Dataset with predictors \( X \), time \( T \), and event status \( E \)
\Require Number of estimators \( n_{\text{estimators}} \), number of folds \( N \), and repeats for cross-validation

\Procedure{Data Partitioning}{}
    \For{each fold \( i = 1, \dots, N \)}
        \State Stratify the dataset into training and testing sets
        \State Assign fold labels: \( FOLD_i = 0 \) for training, \( FOLD_i = 1 \) for testing
    \EndFor
\EndProcedure

\Procedure{Cross-Validation with Bootstrap Ensemble}{}
    \For{each fold and repeat}
        \State \textbf{Initialise Test Prediction List}:
        \State Initialise a list to store predictions from multiple CoxPH models
        \For{each bootstrap iteration \( j = 1, \dots, n_{\text{estimators}} \)}
            \State \textbf{Generate Bootstrap Sample}:
            \State Create a bootstrap sample from the training data
            \State \textbf{Train CoxPH Model}:
            \State Fit a Cox Proportional Hazards model on the bootstrap sample
            \State \textbf{Predict Risk on Test Data}:
            \State Predict risk scores for the test data using the trained CoxPH model
            \State Append predictions to the Test Prediction List
        \EndFor
        \State \textbf{Combine Predictions}:
        \State Compute the average risk score across all bootstrapped models for the test data
    \EndFor
\EndProcedure
\end{algorithmic}
\end{algorithm}

The bootstrap ensemble approach differs from oversampling methods that rely on synthetic data augmentation. While oversampling techniques introduce artificially generated samples to boost minority class representation, the bootstrap method operates solely on real data. By repeatedly sampling from the original dataset with replacement, it produces diverse training subsets, without creating any synthetic instances. This makes it particularly suitable for scenarios where maintaining the integrity of real data is essential.

Another key distinction lies in how predictions are aggregated. In bootstrap ensembles, multiple models are trained on different subsets of the original data, and their predictions are combined through averaging. In addition, the absence of synthetic data in the bootstrap ensemble method avoids potential issues of synthetic instances distorting the true underlying patterns in the data. This can be especially relevant in survival analysis, where precise time-to-event relationships are crucial for accurate risk prediction.

\newpage
\section{Evaluation Metrics for Survival Models}\label{App:TechValid_Metrics}

This appendix briefly explains the two primary evaluation metrics used in survival analysis: \textit{Harrell’s C-index} and the \textit{Calibration Slope}, comparing them to familiar machine learning metrics such as accuracy and Type 1/Type 2 errors. Interested readers should refer to \cite{harrell1982evaluating} and \cite{spiegelhalter1986probabilistic} for more information.

\subsection{Harrell’s C-index: Discrimination as Accuracy}

In machine learning, \textit{accuracy} measures the proportion of correct predictions. Harrell’s C-index evaluates \textit{discrimination} in survival models by assessing how effectively the model ranks individuals' risk of experiencing an event, accounting for the challenge of censored data, where the event may not have occurred during the study period.

The C-index focuses on the \textit{concordance} between predicted risk scores and actual survival outcomes. For a pair of individuals, if one person survives longer than the other, the model should assign a higher risk score to the person with the shorter survival time.

Mathematically, the \textit{concordance} is derived as
\begin{equation}
\text{Concordance} = \frac{1}{|\text{Comparable Pairs}|} \sum_{i,j} 1 \left( \hat{r}_i > \hat{r}_j \right)
\end{equation}
for any pair of individuals \( (i, j) \), where \( i \) has a shorter survival time than \( j \), the pair is \textit{concordant} if the predicted risk for \( i \) is greater than that for \( j \); \( \hat{r}_i \) and \( \hat{r}_j \) are the predicted risks, and \( 1(\cdot) \) is an indicator function that equals 1 if the condition holds and 0 otherwise. Risk is related to the CoxPH hazard ratio computation previously shown in Equation \eqref{Eq:CoxPH}, using \( \hat{r}_i = \mathbf{X}_i^\top \beta \).

\subsection{Calibration Slope: Calibration as Type 1 and Type 2 Errors}

The \textit{calibration slope} is derived from a regression of observed events on predicted risks, which is assessed by plotting predicted risk scores against actual outcomes. Ideally, this relationship follows a 45-degree line, where the predicted probabilities align perfectly with observed outcomes. The slope of this line is known as the calibration slope.

Observed risk \( O_q \) is calculated for each risk quantile \( q \) as the proportion of individuals within that quantile who experienced the event. Mathematically, for a quantile containing \( n_q \) individuals, the observed risk is computed as:
\begin{equation}
O_q = \frac{\sum_{i \in q} 1_{\text{Event}, i}}{n_q}
\end{equation}

where \( 1_{\text{Event}, i} \) is 1 if individual \( i \) experienced the event, and 0 otherwise. To express this as a percentage, it is scaled by 100:
\begin{equation}
O_q^{\%} = 100 \times \frac{\sum_{i \in q} 1_{\text{Event}, i}}{n_q}
\end{equation}

This calculation is used to compare the model's predicted risk against the actual observed event rates, providing a direct measure of calibration.

A slope less than 1 indicates \textit{overestimation of risk} (analogous to more \textit{Type 1 errors}, false positives), where the model predicts events more frequently than they occur. Conversely, a slope greater than 1 indicates \textit{underestimation of risk} (analogous to more \textit{Type 2 errors}, false negatives), where the model predicts events less frequently than they occur.

\newpage
\section{Synthetic Data Comparisons with SMOTE and VAE\newline Across ACTG320, WHAS500, and FLChain}\label{App:SMOTEVAEExtra}

In this appendix, we present a comparative analysis of KM curves and HR consistencies for synthetic datasets generated using SMOTE and VAE across the ACTG320, WHAS500, and FLChain datasets. These results underscore the severe limitations of these widely adopted methods in replicating the intricate survival dynamics and risk factor relationships present in real-world clinical data.

Across all datasets, the KM curves generated using SMOTE demonstrate significant deviations from the real data, particularly in long-term survival probabilities. SMOTE consistently leads to a premature decline in survival rates, which distorts the survival trends seen in the real data. For instance, in the ACTG320 dataset, SMOTE completely fails to preserve the real data's temporal dynamics, resulting in accelerated declines that grossly misrepresent patient outcomes. This pattern is similarly evident in the WHAS500 and FLChain datasets, where the SMOTE-generated KM curves display sharp, unrealistic drops in survival probabilities, particularly over extended follow-up periods. These findings reveal a critical flaw in SMOTE's capacity to handle time-to-event data, making it unsuitable for survival analysis where temporal accuracy is important. Although SMOTE is a popular choice for oversampling, it fundamentally lacks the ability to capture the complex survival patterns necessary for reliable clinical insights.

While VAE demonstrates marginally better performance in preserving KM curves than SMOTE, it remains inadequate for clinical applications. In datasets such as WHAS500 and FLChain, VAE-generated KM curves exhibit closer alignment with real data compared to SMOTE, particularly in the earlier phases of follow-up. However, as time progresses, significant divergences emerge, particularly in mid-to-late survival probabilities. VAE manages to avoid the steep declines characteristic of SMOTE, yet it still fails to capture the nuanced temporal dynamics required for high-utility survival data. These issues further highlight that while VAE can provide a closer approximation than SMOTE, it is still far from generating clinically valid synthetic datasets capable of replacing real-world data in survival analysis.

The HR consistency analysis further reveals the inadequacies of SMOTE and VAE in replicating clinical data. SMOTE consistently produces HRs with point estimates and CIs that deviate significantly from real data, compromising its ability to accurately model risk factors. In the WHAS500 dataset, for example, variables like heart rate and blood pressure show oddly placed point estimates in the synthetic data, failing to capture the true impact of the variables on the risk. The FLChain dataset reveals similar issues, where critical clinical variables such as MGUS diagnosis and creatinine levels show substantial misalignments in their HRs. These deviations render SMOTE-generated synthetic data unusable for clinical risk modelling. VAE, while slightly better than SMOTE, still fails to meet the requirements for high-utility survival data. 

In conclusion, despite the extensive application of SMOTE and VAE in research, our analysis clearly demonstrates that both methods fall far short of producing synthetic datasets with the utility required for real world clinical research. SMOTE’s extreme deviations and VAE’s moderate yet still significant shortcomings highlight the necessity for more sophisticated approaches like CK4Gen, which we propose as a superior alternative capable of achieving the complex multi-variable interactions and temporal accuracy required for reliable clinical decision-making. Our effort here aims to provide the research community with a detailed understanding of these limitations and the potential risks associated with naive application of SMOTE and VAE in clinical settings.

\newpage
\begin{figure}[h]
    \centering
    \begin{subfigure}[b]{0.8\textwidth}
        \centering
        \includegraphics[width=\textwidth]{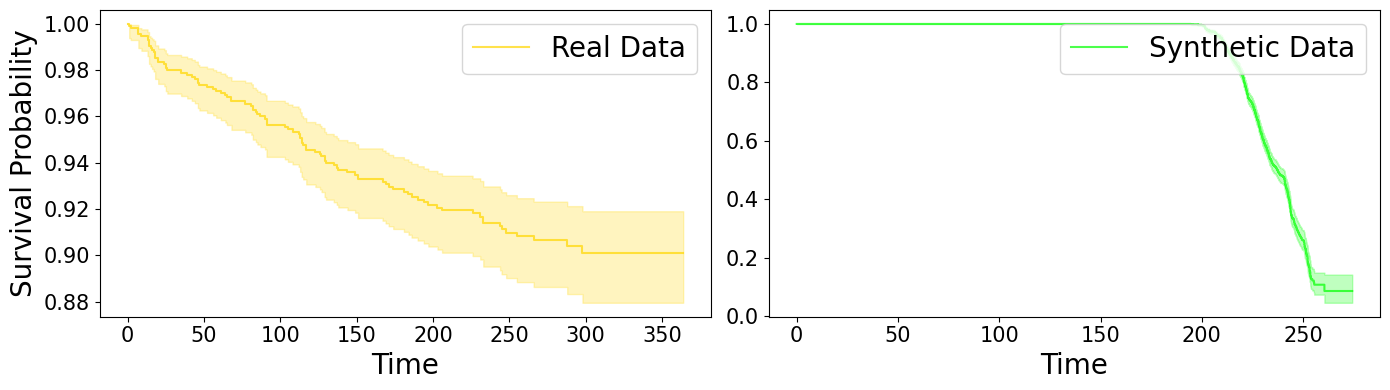}
        \caption{KM curves comparison with SMOTE}
    \end{subfigure}
    
    \vspace{0.5cm} 

    \begin{subfigure}[b]{0.8\textwidth}
        \centering
        \includegraphics[width=\textwidth]{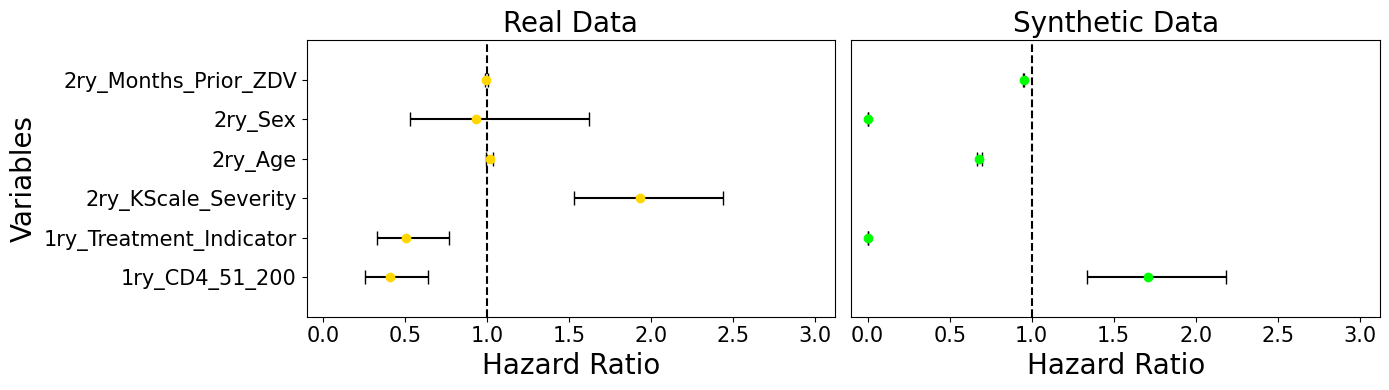}
        \caption{HR consistency comparison with SMOTE}
    \end{subfigure}
    
    \caption{Comparison of KM curves and HR consistencies between real and synthetic data from SMOTE using the ACTG320 dataset.}
\end{figure}

\begin{figure}[h]
    \centering
    \begin{subfigure}[b]{0.8\textwidth}
        \centering
        \includegraphics[width=\textwidth]{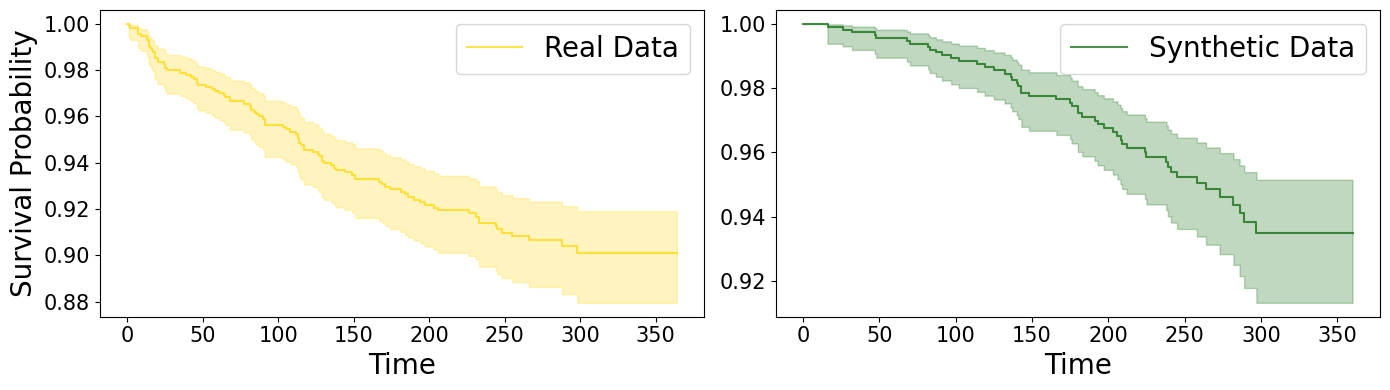}
        \caption{KM curves comparison with VAE}
    \end{subfigure}
    
    \vspace{0.5cm} 

    \begin{subfigure}[b]{0.8\textwidth}
        \centering
        \includegraphics[width=\textwidth]{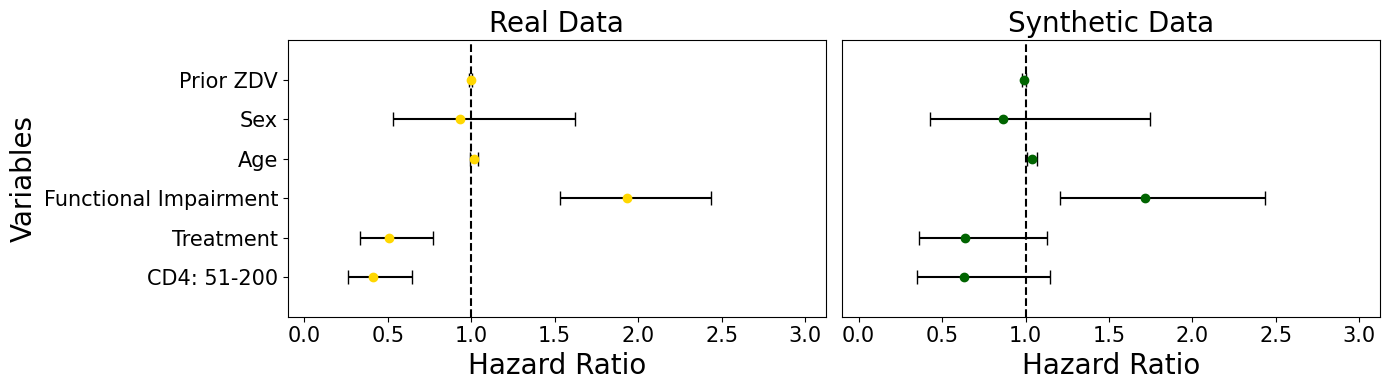}
        \caption{HR consistency comparison with VAE}
    \end{subfigure}
    
    \caption{Comparison of KM curves and HR consistencies between real and synthetic data from VAE using the ACTG320 dataset.}
\end{figure}

\newpage
\begin{figure}[h]
    \centering
    \begin{subfigure}[b]{0.8\textwidth}
        \centering
        \includegraphics[width=\textwidth]{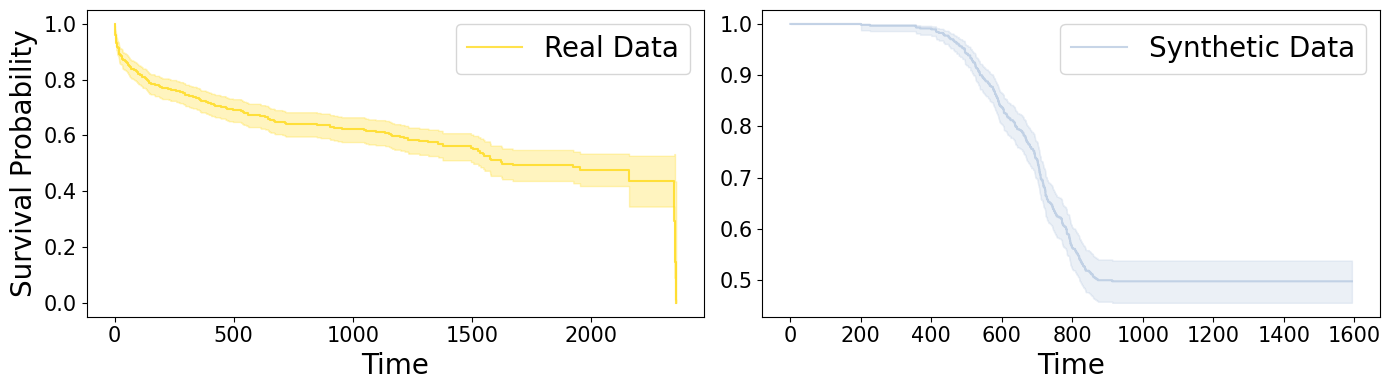}
        \caption{KM curves comparison with SMOTE}
    \end{subfigure}
    
    \vspace{0.5cm} 

    \begin{subfigure}[b]{0.8\textwidth}
        \centering
        \includegraphics[width=\textwidth]{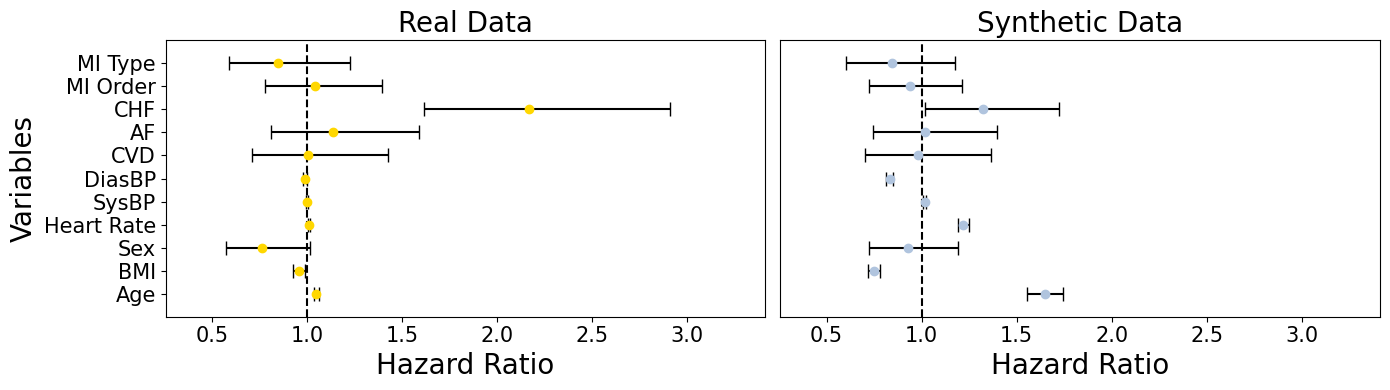}
        \caption{HR consistency comparison with SMOTE}
    \end{subfigure}
    
    \caption{Comparison of KM curves and HR consistencies between real and synthetic data from SMOTE using the WHAS500 dataset.}
\end{figure}

\begin{figure}[h]
    \centering
    \begin{subfigure}[b]{0.8\textwidth}
        \centering
        \includegraphics[width=\textwidth]{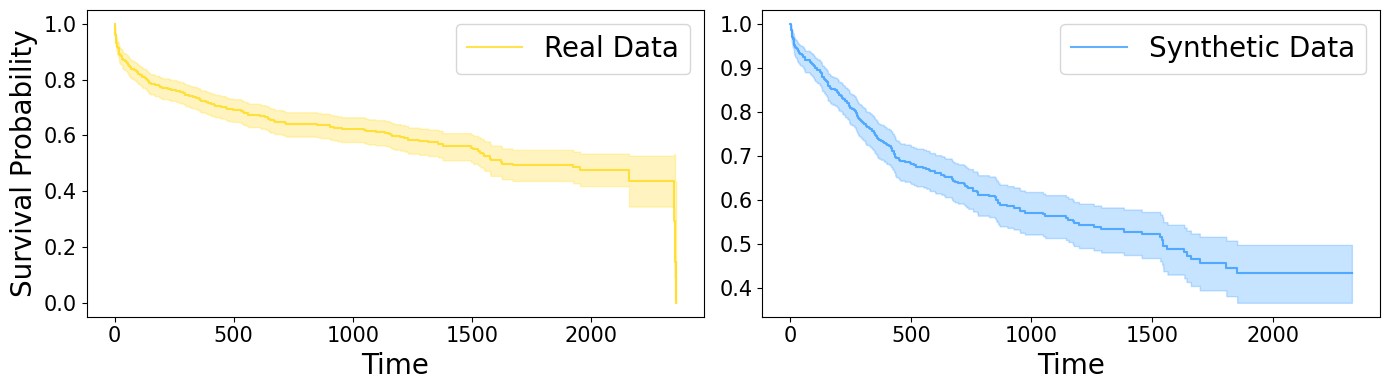}
        \caption{KM curves comparison with VAE}
    \end{subfigure}
    
    \vspace{0.5cm} 

    \begin{subfigure}[b]{0.8\textwidth}
        \centering
        \includegraphics[width=\textwidth]{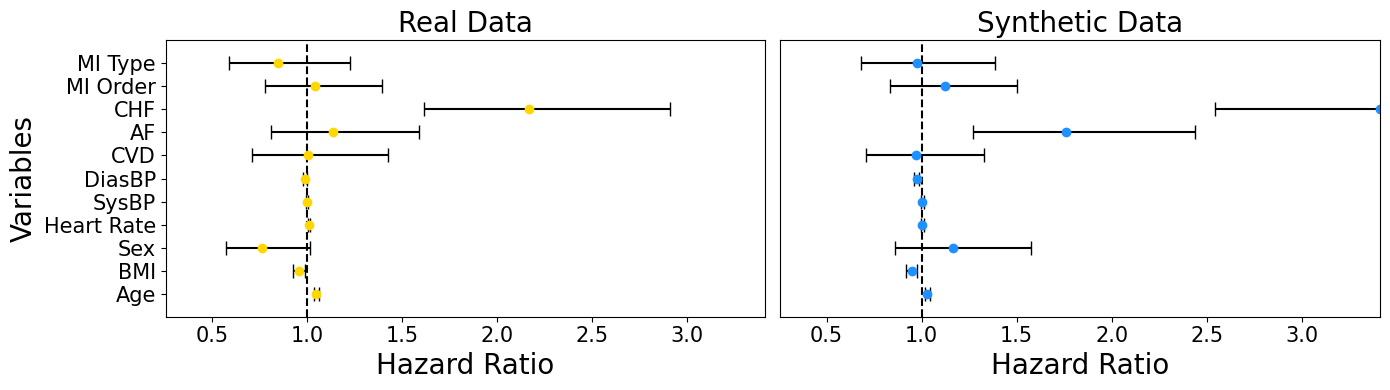}
        \caption{HR consistency comparison with VAE}
    \end{subfigure}
    
    \caption{Comparison of KM curves and HR consistencies between real and synthetic data from VAE using the WHAS500 dataset.}
\end{figure}

\newpage
\begin{figure}[h]
    \centering
    \begin{subfigure}[b]{0.8\textwidth}
        \centering
        \includegraphics[width=\textwidth]{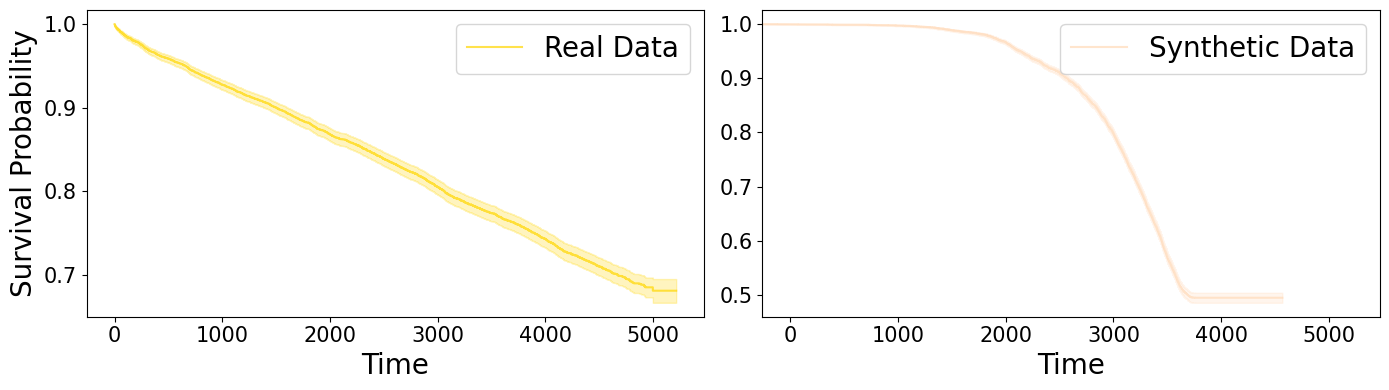}
        \caption{KM curves comparison with SMOTE}
    \end{subfigure}
    
    \vspace{0.5cm} 

    \begin{subfigure}[b]{0.8\textwidth}
        \centering
        \includegraphics[width=\textwidth]{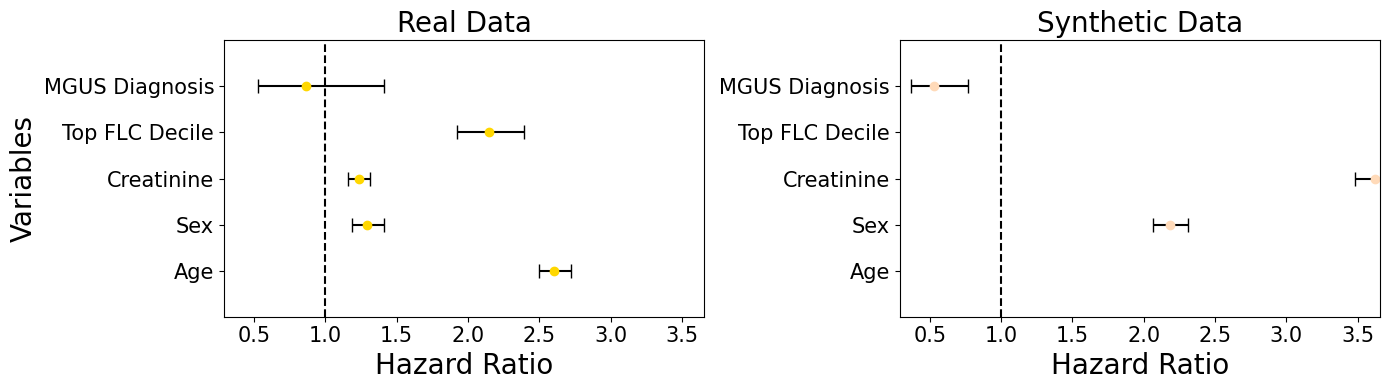}
        \caption{HR consistency comparison with SMOTE}
    \end{subfigure}
    
    \caption{Comparison of KM curves and HR consistencies between real and synthetic data from SMOTE using the FLChain dataset.}
\end{figure}

\begin{figure}[h]
    \centering
    \begin{subfigure}[b]{0.8\textwidth}
        \centering
        \includegraphics[width=\textwidth]{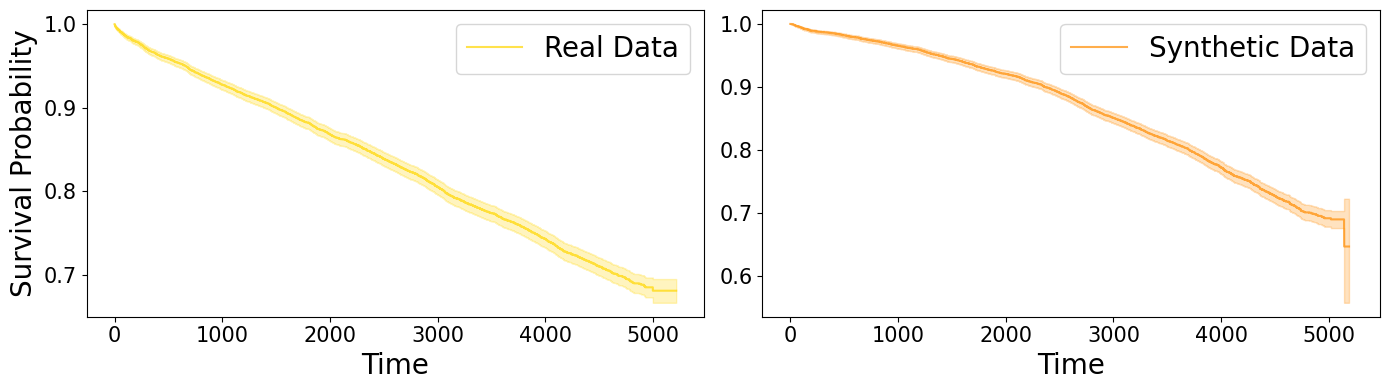}
        \caption{KM curves comparison with VAE}
    \end{subfigure}
    
    \vspace{0.5cm} 

    \begin{subfigure}[b]{0.8\textwidth}
        \centering
        \includegraphics[width=\textwidth]{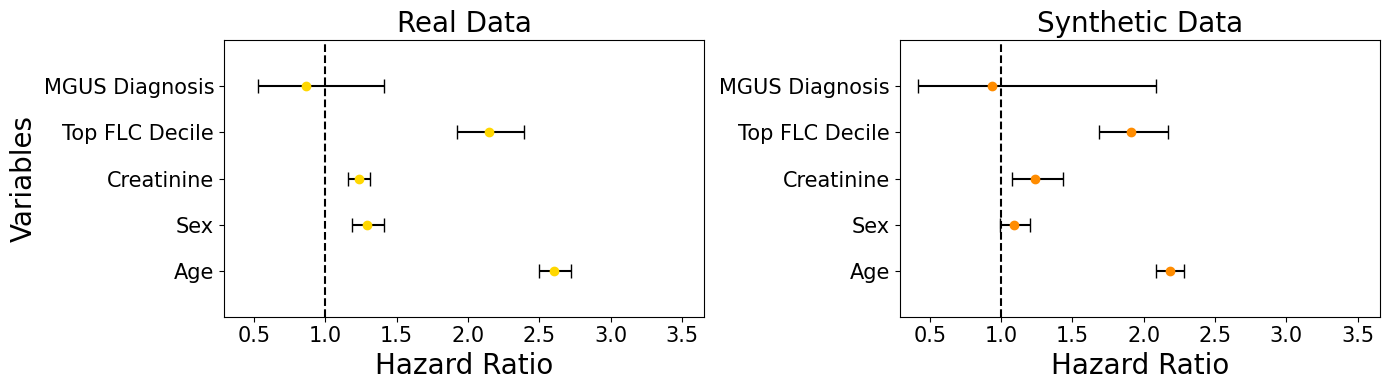}
        \caption{HR consistency comparison with VAE}
    \end{subfigure}
    
    \caption{Comparison of KM curves and HR consistencies between real and synthetic data from VAE using the FLChain dataset.}
\end{figure}

\end{document}